%% file: arxiv_main.tex
\documentclass{article}

\usepackage[preprint]{neurips_2026}
\usepackage[utf8]{inputenc}
\usepackage[T1]{fontenc}
\usepackage{hyperref}
\hypersetup{hidelinks}
\usepackage{url}
\usepackage{booktabs}
\usepackage{amsfonts}
\usepackage{microtype}
\usepackage{xcolor}
\usepackage{algorithm}
\usepackage{algorithmic}
\usepackage{graphicx}
\usepackage{enumerate}
\usepackage{multirow}
\usepackage{makecell}
\usepackage{enumitem}
\usepackage{subcaption}
\usepackage{amsmath,amssymb,mathtools}
\usepackage{wrapfig}
\usepackage{placeins}

\usepackage{float}

\usepackage[table]{xcolor}
\usepackage{tabularx}
\usepackage{adjustbox}
\newcommand{\rankchip}[2]{\begingroup\setlength{\fboxsep}{1.2pt}\textcolor{black}{\colorbox[HTML]{#1}{\scriptsize\,#2\,}}\endgroup}

\definecolor{linkcol}{RGB}{46,111,176}
\newcommand{\codeicon}{\raisebox{-0.22em}{\includegraphics[height=1.05em]{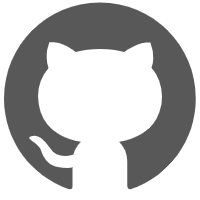}}}
\newcommand{\globeicon}{\raisebox{-0.22em}{\includegraphics[height=1.05em]{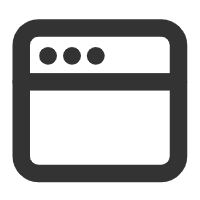}}}
\newcommand{\videoicon}{\raisebox{-0.22em}{\includegraphics[height=1.05em]{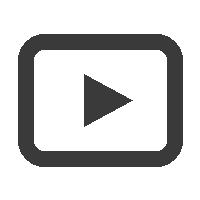}}}
\newcommand{\iconlink}[3]{\href{#1}{#2\,\,{\color{linkcol}\textbf{#3}}}}

\makeatletter
\renewcommand{\@noticestring}{Accepted at the 40th Conference on Neural Information Processing Systems (NeurIPS 2026).}
\makeatother

\title{State of Thought Enables Endogenous Reasoning}

\author{%
{Zhiren Gong$^{1,2}$ \quad Yikun Hou$^{1}$ \quad Zihao Zeng$^{1}$ \quad Ming Xiao$^{4}$} \\
{\bf Chau Yuen$^{3}$ \quad Wei Yang Bryan Lim$^{1}$}\\
{$^{1}$College of Computing and Data Science, Nanyang Technological University, Singapore}\\
{$^{2}$Interdisciplinary Graduate Programme, Nanyang Technological University, Singapore}\\
{$^{3}$School of Electrical and Electronic Engineering, Nanyang Technological University, Singapore}\\
{$^{4}$Information Science \& Engineering, KTH Royal Institute of Technology, Sweden}\\
{\texttt{zhiren001@e.ntu.edu.sg} \quad \texttt{zihao.zeng@ntu.edu.sg} \quad \texttt{bryan.limwy@ntu.edu.sg}}
}

\begin{document}

\maketitle

\vspace{-10mm}
\begin{center}\vspace{-2pt}
{%
\iconlink{https://github.com/GongZhiren/State-of-Thought}{\codeicon}{Code}\hspace{20pt}%
\iconlink{https://gongzhiren.github.io/SoT-website/}{\globeicon}{Project Page}\hspace{20pt}%
\iconlink{https://gongzhiren.github.io/SoT-website/tutorial.html}{\videoicon}{Tutorial}}
\end{center}
\vspace{2mm}

\begin{abstract}
Test-time compute has emerged as a major approach to improving the capabilities of Large Language Models (LLMs).
However, existing test-time reasoning paradigms rely heavily on externally imposed control, either through fixed reasoning programs or through costly expansion in constrained search spaces, limiting both generalization and efficiency.
We propose \textbf{State of Thought (SoT)}, a new reasoning paradigm that enables \textbf{endogenous reasoning} in LLMs, with the model's internal reasoning state governing how reasoning unfolds.
Concretely, SoT extracts a compact \emph{dynamics-geometric state} from the model's internal information transfer and uses a {582-parameter controller on frozen backbones to} \emph{selectively activate historical reasoning support} useful under the current reasoning state, framing reasoning as a state-conditioned process over evidence rather than an externally prescribed token chain.
{Across quantitative (\(1.34\times\)), general (\(1.62\times\)), symbolic-and-code (\(1.76\times\)), and long-context (\(2.51\times\)) reasoning on 3 LLMs and 16 datasets, SoT consistently improves mean-baseline accuracy while reducing generated tokens by \(62.6\%\) and end-to-end latency by \(44.6\%\). Across 2 VLM scales and 3 reasoning tasks, it improves mean accuracy by 3.8 points over reasoning baselines, with \(74.9\%\) fewer completion tokens and \(73.5\%\) lower latency than search-based methods. Under constrained access, SoT retains \(38.2\%\)/\(36.5\%\) mean accuracy gains in training-free/embedding-only settings, while trajectory-only judging reaches \(84.1\%\) agreement across 3 API models. Together, endogenous state-driven reasoning provides a generalizable and efficient alternative.}

\end{abstract}

\begin{center}
    \begin{minipage}{1.0\textwidth}
    \vspace{-2mm}
        \centering
        \includegraphics[width=0.86\textwidth]{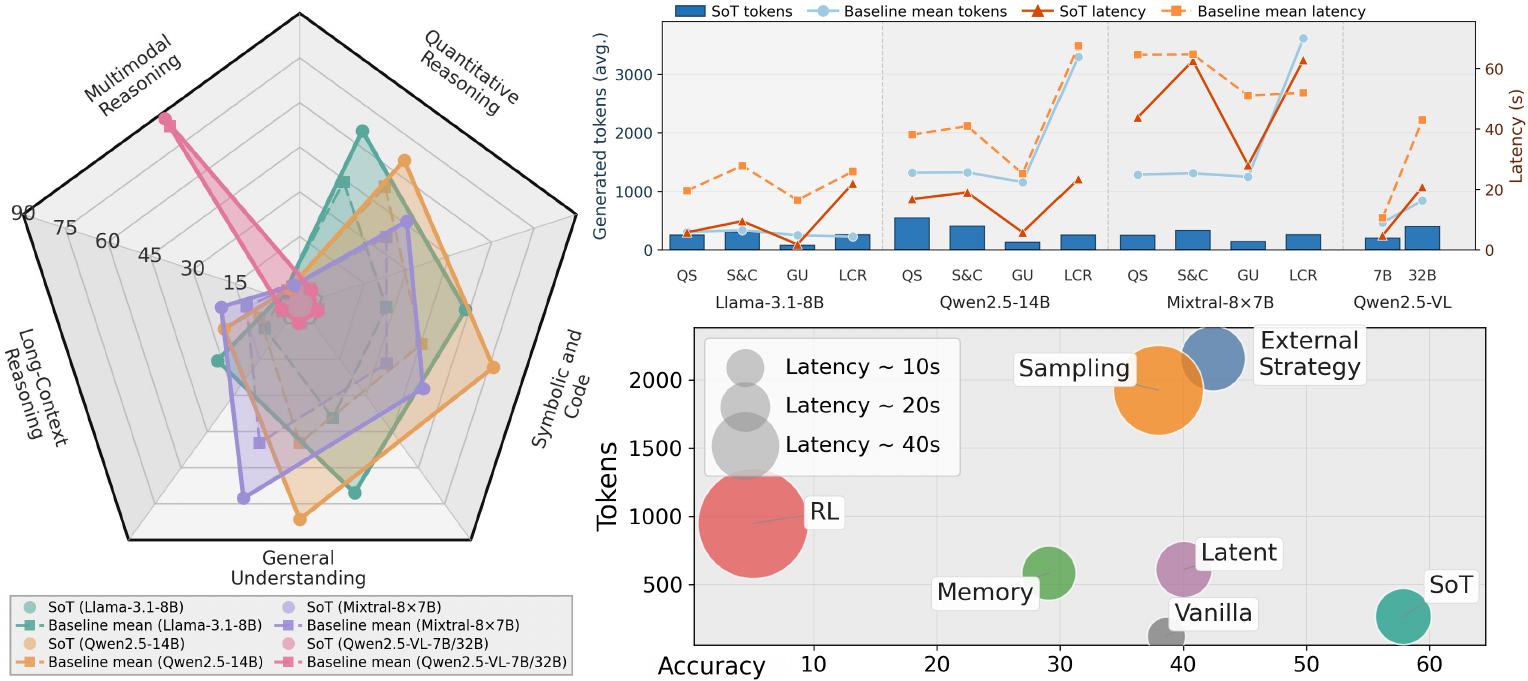}
        \vspace{-1mm}
        \captionof{figure}{
        SoT Performance Across Models, Tasks, and Efficiency Metrics.
        }
        \label{fig:intro_overview}
    \end{minipage}
\end{center}


\vspace{-1mm}
\input{chapters/1-intro}
\input{chapters/2-SoT}
\input{chapters/3-experiment}
\input{chapters/4-explore}
\input{chapters/6-conclusion}

\clearpage
\section*{Acknowledgements}
This research is supported by the Ministry of Education, Singapore, under its Academic Research Fund Tier 2 (Award MOE-T2EP20125-0005).

\bibliographystyle{plainnat}
\bibliography{reference}

\FloatBarrier
\clearpage
\appendix
\input{chapters/Appendix}


\end{document}

%% file: chapters/1-intro.tex
\vspace{-1mm}
\section{Introduction}
\enlargethispage{-2\baselineskip}
\vspace{-1mm}
\begin{figure*}[t]
    \centering
    \includegraphics[width=0.92\textwidth]{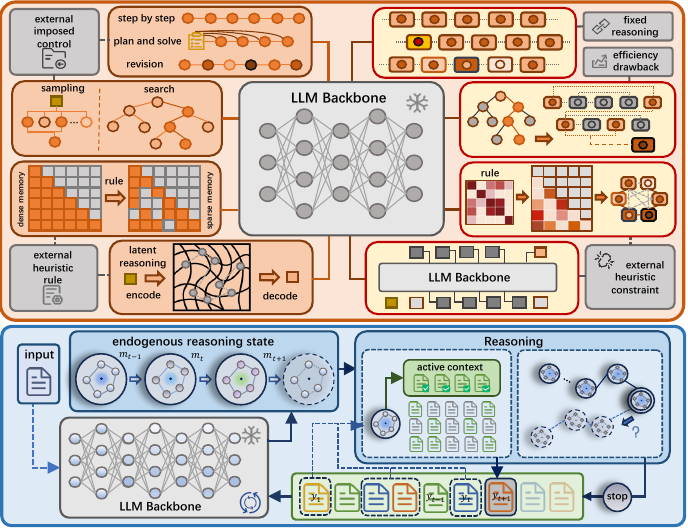}
    \vspace{-2mm}
    \caption{
    From External Control (Top) to State-of-Thought (Bottom).
    }
    \vspace{-2mm}
    \label{fig:intro_main}
\end{figure*}

Reasoning in Large Language Models (LLMs) is fundamentally the process of organizing historical evidence context for subsequent decisions
\citep{press2023measuring,trivedi2023interleaving,yao2023react}.
From this perspective, progress in test-time compute comes from better strategies for organizing the evidence context with additional computation, thereby improving reasoning capabilities \citep{brown2024large,snell2024scaling}.
One dominant line relies on externally structured reasoning programs, requiring the model to follow pre-specified reasoning formats such as step-by-step \citep{wei2022chain}, planning \citep{wang2023plan}, or revision \citep{madaan2023self}.
Another line allocates more test-time search, exploring additional samples \citep{brown2024large,wang2023selfconsistency}, branches \citep{yao2023tree}, or trajectories \citep{snell2024scaling} to accumulate broader candidate evidence for subsequent reasoning.
Despite their empirical success in adapting to particular reasoning strategies, both lines impose reasoning control from outside the model (i.e., through pre-specified formats or search policies) and ignore its evolving internal state as a signal for what evidence is actually needed at subsequent decisions, limiting generalization across heterogeneous reasoning structures {\citep{hassid2025dont,zhiren2026xdomainbench}} and efficiency under constrained inference \citep{zhou2023least}.

Prior work has shown that LLMs internals exhibit structured representation geometry \citep{zou2023repe}, trajectory-level regularities \citep{sun2026llm}, and confidence dynamics \citep{tikhonov2026confidence}.
However, these signals are typically used descriptively, such as monitoring and reasoning-trajectory interpretation {\citep{ballon2026probing,damirchi2026truth,feng2025monitoring,gong2026conditionalcoablationrecoveringselfrepair}}, or early stopping criteria \citep{hosseini2026early}, rather than to govern reasoning itself.
Together, these observations point to the significance of the model's \emph{endogenous state} (i.e., the internally evolving state induced by its own reasoning trajectory).
In parallel, reasoning also depends critically on whether decisions are steadily grounded in suitable and decision-relevant historical evidence rather than generic or noisy history \citep{creswell2022selection,press2023measuring,wang2024rat}. Existing approaches address this through context retrieval and compression \citep{jiang2024longllmlingua,li2023compressing,packer2023memgpt}, or latent-space reasoning \citep{hao2025training,wang2025system}, both of which bypass endogenous state as a vital signal for organizing the evidence genuinely needed for subsequent steps.
Appendix~\ref{app:related_work} provides further related discussion.

This motivates a reframing of reasoning itself: \emph{effective reasoning is actually a dynamic process of organizing the right evidence context for the next decision under the current endogenous state, rather than following an externally prescribed token chain.}
We introduce \textbf{State of Thought (SoT)}, a reasoning paradigm for \emph{endogenous reasoning}.
At each reasoning step, SoT reads out a compact dynamics-geometric state from the model's internal information transfer and uses {a lightweight 582-parameter controller} to selectively activate the historical evidence context most useful for the next decision.
The active reasoning context is therefore not generated by pre-specified reasoning programs or broader searches, but by the model's own evolving state.
In this way, SoT turns reasoning into a closed loop: endogenous state organizes active evidence, which conditions the next reasoning step, and the new step in turn updates the state.

We evaluate SoT at scale on {3} LLM backbones over 16 datasets and {2 VLM scales over 3 reasoning tasks}.
Across the 16-dataset LLM suite, SoT improves accuracy by \({89.8}\%\), \({58.6}\%\), and \({46.0}\%\) over {the mean }baselines on Llama-3.1-8B, Qwen2.5-14B, and Mixtral-8x7B, while reducing tokens/latency by \({19.7}\%\)/\({56.5}\%\), \({81.1}\%\)/\({62.2}\%\), and \({86.8}\%\)/\({15.1}\%\), respectively.
{Across Qwen2.5-VL-7B/32B and 3 reasoning tasks, SoT improves mean accuracy by 3.8 points over reasoning baselines, leads in 5 of 6 model--task settings, and uses \(74.9\%\) fewer completion tokens with \(73.5\%\) lower latency than search-based methods.}
As exploratory extensions under constrained access, a training-free controller and an embedding-only controller still achieve \({38.2\%}\) and \({36.5\%}\) improvement over baselines on accuracy across models, while a trajectory-level post-hoc judge reaches \(84.1\%\) agreement on three API-accessed models.
Together, these results suggest that changing the control variable of reasoning—from external token programs to endogenous state-conditioned evidence organization—offers a scalable and robust route to reasoning improvement.




%% file: chapters/2-SoT.tex
\section{State of Thought}
\label{sec:method}

\subsection{Problem Formulation}
\label{sec:problem_formulation}

We formulate test-time reasoning of a frozen autoregressive LLM \(f\) as a policy-governed multi-step trajectory generation started from a problem \(x\) presented by prompt \(p\) \citep{brown2024large}.
The model \(f\) dynamically generates the subsequent reasoning trajectory \(y_{1:T}=(y_1,\dots,y_T)\) leading to the final answer \(a\):
\begin{equation}
\mathbb{P}_{\pi}(a, y_{1:T}\mid p)
=
\prod_{t=1}^{T} \mathbb{P}_{\pi}(y_t \mid p,y_{<t})\;
\mathbb{P}_{\pi}(a \mid p,y_{1:T}),
\end{equation}
with the policy \(\pi\) determining how evidence from prior steps is organized for subsequent decisions, while different reasoning paradigms differ in prompt construction and policy \citep{snell2024scaling}.

State of Thought seeks such a policy that the model's current endogenous state (Sec.~\ref{sec:endogenous_state}) governs both evidence-context organization and reasoning continuation (Sec.~\ref{sec:state_driven_activation}), with no external reasoning instructions imposed on the prompt.
{This creates a direct bridge between \emph{continuous} internal evolution and \emph{discrete}, inspectable reasoning steps: the continuous state controls which explicit evidence is carried forward, without replacing language-space reasoning by an opaque latent chain.}

\begin{figure*}[!t]
    \centering
    \includegraphics[width=1.0\textwidth]{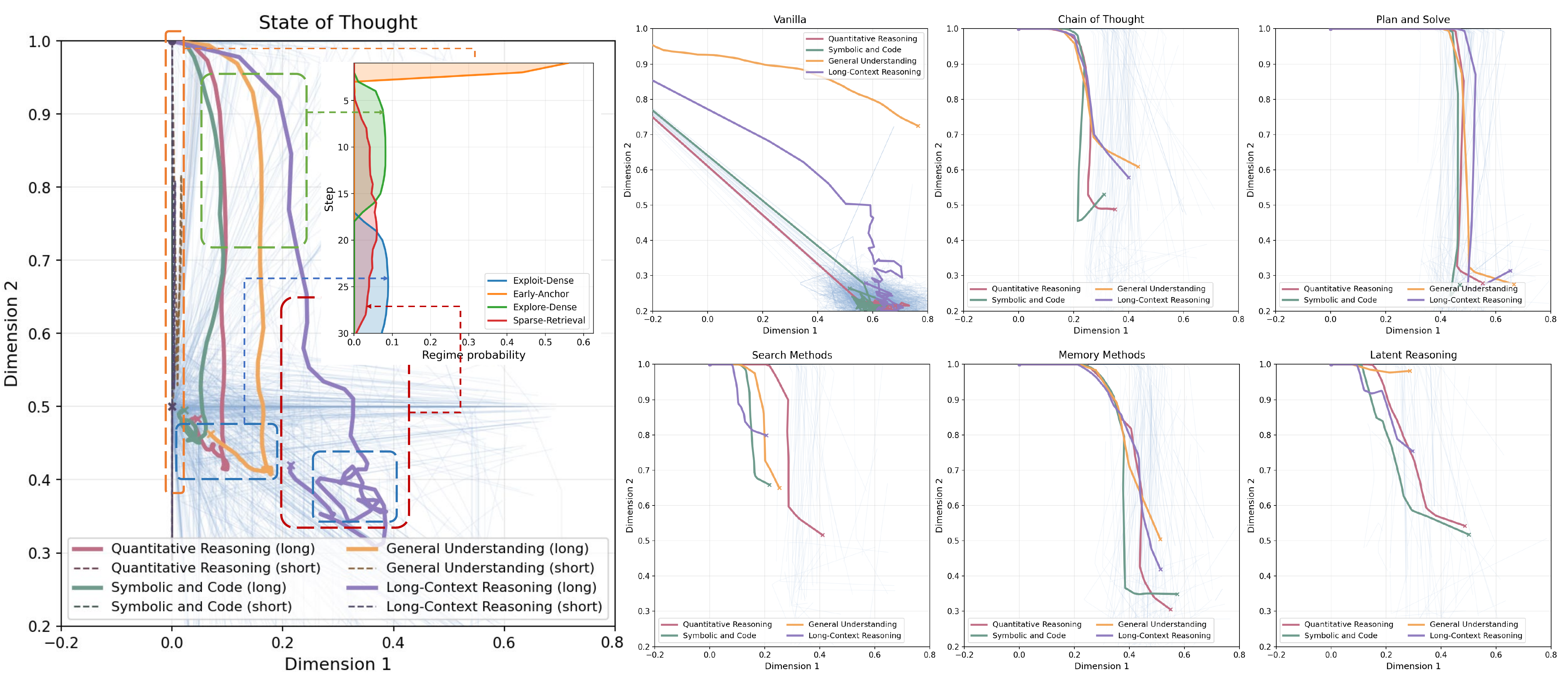}
    \caption{
    \textbf{Endogenous reasoning state trajectories across various reasoning paradigms} on Llama-3.1-8B with breakdown in Figure~\ref{fig:app_state_regimes} and reasoning state definition in Figure~\ref{fig:state_conditioned_activation}.
    Different externally induced reasoning strategies occupy distinguishable regions in the dynamics-geometric state space, while SoT supports a broader and more adaptive state distribution.
    }
    \vspace{-3mm}
    \label{fig:state_space_regimes}
\end{figure*}

\subsection{Endogenous Reasoning State}
\label{sec:endogenous_state}

Effective reasoning policy \(\pi\) should be governed not by external scripts but by a compact endogenous state that reflects the geometry and dynamics of ongoing reasoning \citep{park2025geometry,park2024linear,zou2023repe}. Accordingly, we read out a compact \emph{dynamics-geometric state} \(m_t \in \mathbb{R}^4\) from the model's internal information transfer of sentence-level reasoning unit \(y_t\), with {calculation details and design rationale} in Appendix~\ref{app:endogenous_state}:
\begin{equation}
m_t = \big[\delta_t,\; v_t,\; c_t,\; H_t\big],
\label{eq:sot_state}
\end{equation}
where \(\delta_t\) characterizes the internal organization distinguishing concentrated from diffuse local structure; \(v_t\) captures stepwise progress through magnitude; \(c_t\) captures directional consistency distinguishing stable continuation from deviation or redirection; and \(H_t\) measures the model's local predictive uncertainty. Together, these four quantities summarize the current reasoning endogenous control state and condition how the evidence context should be organized for subsequent decisions.

\paragraph{Interpretability of endogenous reasoning state.}
To verify that \(m_t\) can capture reasoning-relevant internal structure rather than arbitrary hidden variation, we aggregate trajectories from multiple external reasoning paradigms in the constructed state space in Figure~\ref{fig:state_space_regimes}.
SoT spans a broader and more adaptive state distribution, while others with externally structured reasoning programs occupy distinguishable regions, collapsing to fixed scripted patterns.
This indicates that SoT is less constrained by externally imposed reasoning formats and more aligned with the model's endogenous state evolution, enabling stronger generalization across heterogeneous reasoning structures.

\subsection{State-Conditioned Reasoning}
\label{sec:state_driven_activation}

SoT instantiates the reasoning policy \(\pi\) through evidence-organization operator \(\mathcal{S}\) and stopping operator \(\mathcal{T}\) as two state-conditioned operators with details in Appendix~\ref{app:sot-details}.

\emph{State-conditioned evidence selection.}
Given the current state \(m_t\) and the historical reasoning trajectory \(y_{<t}\), the evidence-organization operator \(\mathcal{S}\) selects the reasoning-related sparse subset \(\widetilde{y}_{<t}=\mathcal{S}(y_{<t};m_t)\subset y_{<t}\) as the evidence context for subsequent reasoning to generate the next step
\begin{equation}
\mathbb{P}_{\pi}(y_t \mid p,y_{<t}) = \mathbb{P}\!\left(y_t \mid p,\widetilde{y}_{<t}\right),
\label{eq:state_policy_select}
\end{equation}
In this way, SoT improves reasoning not by broadening search over more trajectories \citep{yao2023tree}, but by sharpening the evidence context carried into the next decision.

\begin{figure*}[!t]
    \centering
    \includegraphics[width=1.0\textwidth]{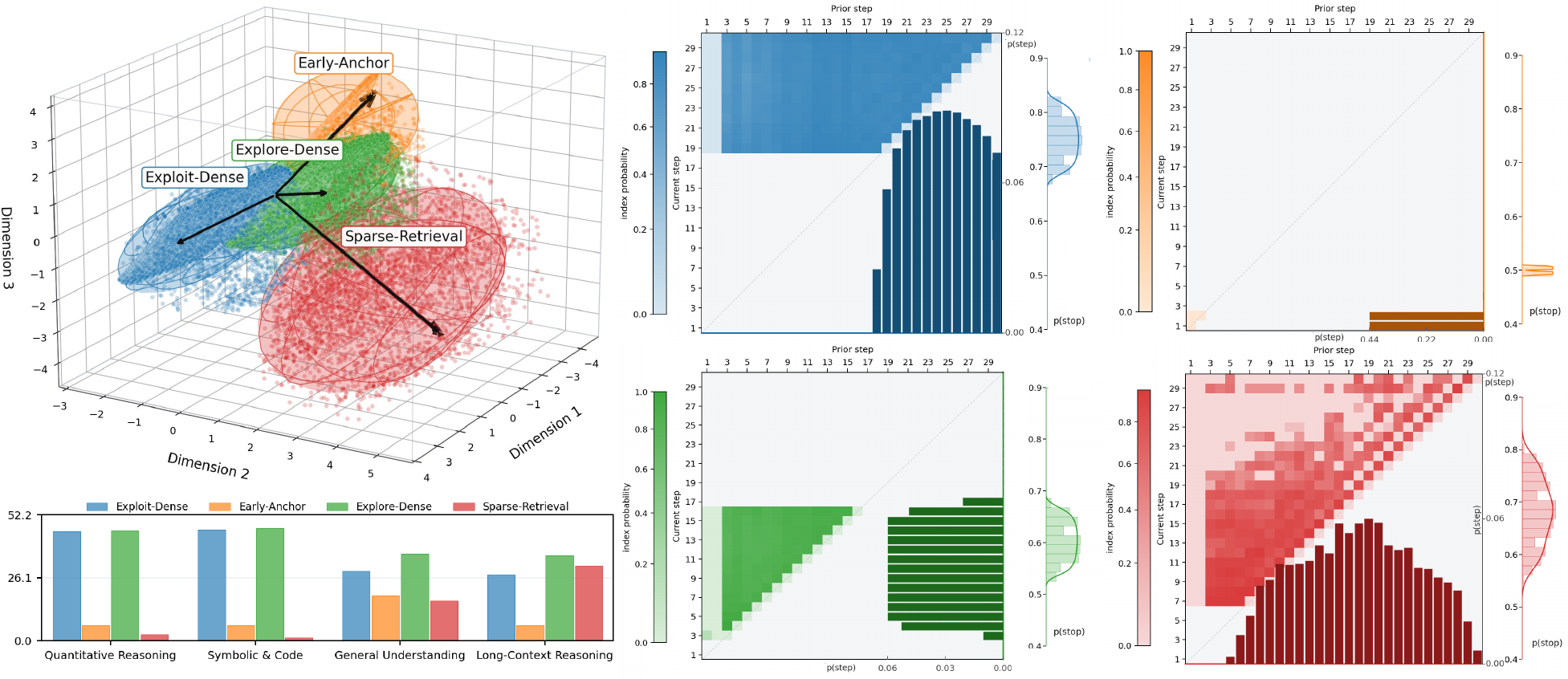}
    \caption{
    \textbf{State-conditioned reasoning in SoT} with breakdown in Figure~\ref{fig:app_state_clusters}.
    Different endogenous reasoning states induce different sparse activation patterns and different stop tendencies.
    }
    \vspace{-3mm}
    \label{fig:state_conditioned_activation}
\end{figure*}

\emph{State-conditioned stopping.}
SoT also produces a continuation decision \(z_t=\mathcal{T}(m_t)\in\{0,1\}\), where \(z_t=0\) continues reasoning and \(z_t=1\) terminates reasoning and triggers final answer generation as
\begin{equation}
\mathbb{P}_{\pi}(a \mid p,y_{1:t}) = \mathbb{P}(a \mid p,\widetilde{y}_{<t}, y_t).
\label{eq:state_policy_step}
\end{equation}
Stopping therefore emerges from the current reasoning endogenous state rather than from a fixed external schedule \citep{hosseini2026early} on reasoning depth.

Overall, SoT treats evidence sparsification and stopping as two outputs of the same endogenous reasoning policy: one determines what evidence context remains active, and the other determines when sufficient support has been accumulated for answer generation.

\paragraph{Interpretability of state-conditioned reasoning.}
We aggregate reasoning trajectories, cluster endogenous states, and analyze the induced evidence-selection and stopping patterns in each state regime in Figure~\ref{fig:state_conditioned_activation}.
We find that different state clusters consistently and stably correspond to distinct sparse activation profiles and stop probabilities, indicating that SoT organizes evidence context and continuation behavior in a state-dependent manner, supporting the core view that effective reasoning is governed by endogenous states rather than externally prescribed token chains.
{Cluster names summarize four-coordinate centroids and measured activation--stopping profiles, not task labels; Appendix~\ref{app:cluster_ops} details their interpretation.}

\paragraph{Lightweight training.}
SoT learns only the state-conditioned evidence-organization operator \(\mathcal{S}\) and stopping operator \(\mathcal{T}\), while the backbone remains frozen, with training details in Appendix~\ref{app:training_method}; Sec.~\ref{sec:training_free_sot} further removes this controller training.
Supervision is constructed offline from multi-step reasoning trajectories: each endogenous state is paired with (i) preference-style signals indicating which historical evidence remains useful for subsequent reasoning, in the spirit of direct preference learning \citep{hong2024orpo,rafailov2023direct}, and (ii) stop proxies indicating whether reasoning should continue, following findings that intermediate reasoning states can support learned early stopping \citep{liu2025answer}.
This design is also consistent with context-selection and compression methods that learn lightweight selection modules from offline trajectories or contextual supervision rather than retraining the backbone model \citep{jiang2024longllmlingua,li2023compressing}.
The resulting controller has only 582 parameters (a $16\!\times\!32$ selection projection with a scalar output, plus a linear $4\!\to\!1$ stopping head); across 5 problem-held-out refits, its selector AUC is $0.779\pm0.005$. Thus, SoT learns a compact control interface rather than updating or duplicating backbone reasoning capacity.

%% file: chapters/3-experiment.tex
\providecommand{\rankchip}[2]{\begingroup\setlength{\fboxsep}{1.2pt}\textcolor{black}{\colorbox[HTML]{#1}{\scriptsize\,#2\,}}\endgroup}
\begin{table*}[t]
    \centering
    \caption{\textbf{Main results on Llama-3.1-8B (accuracy, \%).}}
    \vspace{-2mm}
    \label{tab:main_results_llama}
    \footnotesize
    {\setlength{\tabcolsep}{1.85pt}%
     \setlength{\aboverulesep}{0pt}%
     \setlength{\belowrulesep}{0pt}%
     \setlength{\extrarowheight}{0pt}%
     \renewcommand{\arraystretch}{1.06}%
     \renewcommand{\tabularxcolumn}[1]{m{#1}}%
    \begin{tabularx}{\linewidth}{@{} >{\centering\arraybackslash}p{1.65cm} >{\centering\arraybackslash}m{1.28cm} !{\vrule width 0.4pt} >{\centering\arraybackslash}X >{\centering\arraybackslash}X >{\centering\arraybackslash}X >{\centering\arraybackslash}m{3.1em} >{\centering\arraybackslash}X >{\centering\arraybackslash}X >{\centering\arraybackslash}X >{\centering\arraybackslash}X >{\centering\arraybackslash}X >{\columncolor[HTML]{EAECEF}\centering\arraybackslash}m{3.1em} @{}}
    \toprule
    \multirow{2}{=}{\centering\textbf{Category}} & \multirow{2}{=}{\centering\textbf{Method}}
    & \multicolumn{4}{>{\cellcolor[HTML]{E3F2FD}}c!{\vrule width 1pt}}{\textbf{Quantitative Reasoning}}
    & \multicolumn{6}{>{\cellcolor[HTML]{E3F2FD}}c}{\textbf{Symbolic and Code}} \\
    &
    & {\small\textbf{GSM}} & {\small\textbf{MATH}} & {\small\textbf{DROP}} & \multicolumn{1}{>{\centering\arraybackslash}m{3.1em}!{\vrule width 1pt}}{\small\textbf{avg.}}
    & {\small\textbf{FOL}} & {\small\textbf{PW}} & {\small\textbf{BBH}} & {\small\textbf{HE}} & {\small\textbf{MBPP}} & \multicolumn{1}{>{\centering\arraybackslash}m{3.1em}}{\small\textbf{avg.}} \\
    \midrule
    Greedy & Vanilla & 79.0 & 58.2 & 6.2 & \multicolumn{1}{>{\centering\arraybackslash}m{3.1em}!{\vrule width 1pt}}{\cellcolor[HTML]{EAECEF}{53.9}} & 29.6 & 26.0 & 26.0 & 25.0 & \cellcolor[HTML]{FFD4D3}{55.2} & 33.0 \\
    \midrule
    \multirow{6}{=}{\centering Reasoning Paradigm}
    & CoT   & 80.8 & \cellcolor[HTML]{FCE4D5}{64.2} & 1.7 & \multicolumn{1}{>{\centering\arraybackslash}m{3.1em}!{\vrule width 1pt}}{\cellcolor[HTML]{EAECEF}{55.5}} & 4.9 & 24.5 & \cellcolor[HTML]{FCE4D5}{70.0} & 25.0 & \cellcolor[HTML]{FCE4D5}{51.6} & 31.2 \\
    & PS    & 71.2 & 40.0 & 1.4 & \multicolumn{1}{>{\centering\arraybackslash}m{3.1em}!{\vrule width 1pt}}{\cellcolor[HTML]{EAECEF}{43.4}} & 4.9 & 15.2 & \cellcolor[HTML]{FFFCE0}{55.0} & 10.0 & 22.0 & 17.6 \\
    & SR    & 75.8 & 44.8 & \cellcolor[HTML]{FFFCE0}{15.6} & \multicolumn{1}{>{\centering\arraybackslash}m{3.1em}!{\vrule width 1pt}}{\cellcolor[HTML]{EAECEF}{50.4}} & 32.5 & 31.0 & 46.0 & 25.0 & 36.4 & 32.9 \\
    & SC    & \cellcolor[HTML]{FFD4D3}{83.6} & 54.0 & 1.6 & \multicolumn{1}{>{\centering\arraybackslash}m{3.1em}!{\vrule width 1pt}}{\cellcolor[HTML]{EAECEF}{53.2}} & 16.8 & 28.5 & 52.0 & 10.0 & 35.2 & 27.3 \\
    & CB    & 48.0 & 36.5 & \cellcolor[HTML]{FCE4D5}{19.8} & \multicolumn{1}{>{\centering\arraybackslash}m{3.1em}!{\vrule width 1pt}}{\cellcolor[HTML]{EAECEF}{37.1}} & \cellcolor[HTML]{FFFCE0}{40.4} & \cellcolor[HTML]{FFFCE0}{35.8} & 34.0 & \cellcolor[HTML]{FFFCE0}{30.0} & \cellcolor[HTML]{FFFCE0}{49.2} & 38.6 \\
    & MCTS  & 55.2 & 33.5 & 13.2 & \multicolumn{1}{>{\centering\arraybackslash}m{3.1em}!{\vrule width 1pt}}{\cellcolor[HTML]{EAECEF}{37.5}} & 36.0 & 31.5 & 30.0 & \cellcolor[HTML]{FCE4D5}{35.0} & \cellcolor[HTML]{FFFCE0}{49.2} & 36.7 \\
    \midrule
    \multirow{3}{=}{\centering Memory-Oriented}
    & H2O    & 79.8 & 59.2 & 2.4 & \multicolumn{1}{>{\centering\arraybackslash}m{3.1em}!{\vrule width 1pt}}{\cellcolor[HTML]{EAECEF}{53.6}} & 6.4 & 27.8 & 48.0 & 10.0 & 42.0 & 26.3 \\
    & SNAP   & 80.4 & \cellcolor[HTML]{FFFCE0}{60.2} & 2.3 & \multicolumn{1}{>{\centering\arraybackslash}m{3.1em}!{\vrule width 1pt}}{\cellcolor[HTML]{EAECEF}{54.1}} & 8.4 & 26.2 & \cellcolor[HTML]{FFFCE0}{55.0} & 15.0 & 41.2 & 27.3 \\
    & STREAM & 70.4 & 44.2 & 1.3 & \multicolumn{1}{>{\centering\arraybackslash}m{3.1em}!{\vrule width 1pt}}{\cellcolor[HTML]{EAECEF}{44.4}} & 1.5 & 31.8 & 1.0 & 0.0 & 5.6 & 13.0 \\
    \midrule
    Latent & COCO & \cellcolor[HTML]{FFFCE0}{81.4} & 60.0 & 3.7 & \multicolumn{1}{>{\centering\arraybackslash}m{3.1em}!{\vrule width 1pt}}{\cellcolor[HTML]{EAECEF}{54.8}} & \cellcolor[HTML]{FCE4D5}{45.3} & \cellcolor[HTML]{FCE4D5}{45.2} & 51.0 & 20.0 & 47.2 & 42.5 \\
    \midrule
    RL-Based & \makecell{\scriptsize{GRPO-SP}} & 0.0 & 0.5 & 0.3 & \multicolumn{1}{>{\centering\arraybackslash}m{3.1em}!{\vrule width 1pt}}{\cellcolor[HTML]{EAECEF}{0.2}} & 2.0 & 2.2 & 7.0 & 0.0 & 0.0 & 1.8 \\
    \midrule
    Ours & \textbf{SoT} & \cellcolor[HTML]{FCE4D5}{\textbf{{81.8}}} & \cellcolor[HTML]{FFD4D3}{\textbf{{64.8}}} & \cellcolor[HTML]{FFD4D3}{\textbf{{40.5}}} & \multicolumn{1}{>{\centering\arraybackslash}m{3.1em}!{\vrule width 1pt}}{\cellcolor[HTML]{EAECEF}{\underline{\textbf{{65.8}}}}} & \cellcolor[HTML]{FFD4D3}{\textbf{{49.8}}} & \cellcolor[HTML]{FFD4D3}{\textbf{{64.8}}} & \cellcolor[HTML]{FFD4D3}{\textbf{{78.0}}} & \cellcolor[HTML]{FFD4D3}{\textbf{{59.1}}} & \textbf{{46.8}} & \underline{\textbf{{58.4}}} \\
    \midrule
    \end{tabularx}\par\addvspace{0.35ex}
    \begin{tabularx}{\linewidth}{@{} >{\centering\arraybackslash}p{1.65cm} >{\centering\arraybackslash}m{1.28cm} !{\vrule width 0.4pt} >{\centering\arraybackslash}X >{\centering\arraybackslash}X >{\centering\arraybackslash}X >{\centering\arraybackslash}X >{\centering\arraybackslash}X >{\centering\arraybackslash}m{3.1em} >{\centering\arraybackslash}X >{\centering\arraybackslash}X >{\centering\arraybackslash}X >{\columncolor[HTML]{EAECEF}\centering\arraybackslash}m{3.1em} @{}}
    \multirow{2}{=}{\centering\textbf{Category}} & \multirow{2}{=}{\centering\textbf{Method}}
    & \multicolumn{6}{>{\cellcolor[HTML]{E3F2FD}}c!{\vrule width 1pt}}{\textbf{General Understanding}}
    & \multicolumn{4}{>{\cellcolor[HTML]{E3F2FD}}c}{\textbf{Long-Context Reasoning}} \\
    &
    & {\small\textbf{CSQA}} & {\small\textbf{StrQA}} & {\small\textbf{BoolQ}} & {\small\textbf{MMLU}} & {\small\textbf{RACE}} & \multicolumn{1}{>{\centering\arraybackslash}m{3.1em}!{\vrule width 1pt}}{\small\textbf{avg.}}
    & {\small\textbf{HQA}} & {\small\textbf{NarQA}} & {\small\textbf{LB}} & \multicolumn{1}{>{\centering\arraybackslash}m{3.1em}}{\small\textbf{avg.}} \\
    \midrule
    Greedy & Vanilla & \cellcolor[HTML]{FFFCE0}{48.0} & \cellcolor[HTML]{FCE4D5}{68.8} & 59.2 & 45.6 & 50.7 & \multicolumn{1}{>{\centering\arraybackslash}m{3.1em}!{\vrule width 1pt}}{\cellcolor[HTML]{EAECEF}{54.2}} & 7.0 & 20.1 & \cellcolor[HTML]{FFFCE0}{32.7} & 16.2 \\
    \midrule
    \multirow{6}{=}{\centering Reasoning Paradigm}
    & CoT   & 32.8 & 31.5 & 37.2 & 34.0 & 50.3 & \multicolumn{1}{>{\centering\arraybackslash}m{3.1em}!{\vrule width 1pt}}{\cellcolor[HTML]{EAECEF}{36.3}} & 2.0 & 4.5 & 26.9 & 7.1 \\
    & PS    & 30.5 & 18.2 & 21.5 & 31.2 & 42.0 & \multicolumn{1}{>{\centering\arraybackslash}m{3.1em}!{\vrule width 1pt}}{\cellcolor[HTML]{EAECEF}{28.1}} & 1.7 & 3.0 & 20.4 & 5.3 \\
    & SR    & \cellcolor[HTML]{FCE4D5}{57.0} & 64.0 & \cellcolor[HTML]{FFFCE0}{72.2} & \cellcolor[HTML]{FCE4D5}{62.4} & \cellcolor[HTML]{FCE4D5}{62.7} & \multicolumn{1}{>{\centering\arraybackslash}m{3.1em}!{\vrule width 1pt}}{\cellcolor[HTML]{EAECEF}{63.6}} & \cellcolor[HTML]{FCE4D5}{17.9} & 21.7 & 25.5 & 20.6 \\
    & SC    & 28.5 & 34.5 & 43.2 & 31.0 & 45.7 & \multicolumn{1}{>{\centering\arraybackslash}m{3.1em}!{\vrule width 1pt}}{\cellcolor[HTML]{EAECEF}{35.8}} & 1.9 & 4.3 & 25.8 & 6.8 \\
    & CB    & 46.0 & \cellcolor[HTML]{FFD4D3}{71.5} & \cellcolor[HTML]{FCE4D5}{82.2} & \cellcolor[HTML]{FFFCE0}{51.2} & \cellcolor[HTML]{FFFCE0}{56.0} & \multicolumn{1}{>{\centering\arraybackslash}m{3.1em}!{\vrule width 1pt}}{\cellcolor[HTML]{EAECEF}{61.1}} & \cellcolor[HTML]{FFFCE0}{15.5} & \cellcolor[HTML]{FCE4D5}{27.5} & 26.2 & 21.8 \\
    & MCTS  & 42.0 & \cellcolor[HTML]{FFFCE0}{65.5} & \cellcolor[HTML]{FCE4D5}{82.2} & 46.4 & 49.7 & \multicolumn{1}{>{\centering\arraybackslash}m{3.1em}!{\vrule width 1pt}}{\cellcolor[HTML]{EAECEF}{57.0}} & 13.5 & \cellcolor[HTML]{FFFCE0}{27.1} & 27.8 & 21.0 \\
    \midrule
    \multirow{3}{=}{\centering Memory-Oriented}
    & H2O    & 33.5 & 30.8 & 39.2 & 35.0 & 19.7 & \multicolumn{1}{>{\centering\arraybackslash}m{3.1em}!{\vrule width 1pt}}{\cellcolor[HTML]{EAECEF}{32.4}} & 2.0 & 2.0 & 0.0 & 1.7 \\
    & SNAP   & 33.5 & 31.0 & 38.5 & 34.6 & 16.7 & \multicolumn{1}{>{\centering\arraybackslash}m{3.1em}!{\vrule width 1pt}}{\cellcolor[HTML]{EAECEF}{31.8}} & 2.0 & 2.0 & 0.0 & 1.7 \\
    & STREAM & 34.8 & 26.2 & 38.8 & 22.4 & 0.0 & \multicolumn{1}{>{\centering\arraybackslash}m{3.1em}!{\vrule width 1pt}}{\cellcolor[HTML]{EAECEF}{25.6}} & 2.5 & 0.0 & 0.0 & 1.1 \\
    \midrule
    Latent & COCO & 40.5 & 63.5 & 70.0 & 42.8 & 43.3 & \multicolumn{1}{>{\centering\arraybackslash}m{3.1em}!{\vrule width 1pt}}{\cellcolor[HTML]{EAECEF}{52.0}} & 4.4 & 10.8 & \cellcolor[HTML]{FCE4D5}{34.6} & 11.9 \\
    \midrule
    RL-Based & \makecell{\scriptsize{GRPO-SP}} & 10.5 & 3.8 & 9.8 & 8.2 & 12.0 & \multicolumn{1}{>{\centering\arraybackslash}m{3.1em}!{\vrule width 1pt}}{\cellcolor[HTML]{EAECEF}{8.7}} & 0.6 & 1.1 & 5.3 & 1.6 \\
    \midrule
    Ours & \textbf{SoT} & \cellcolor[HTML]{FFD4D3}{\textbf{{69.5}}} & \cellcolor[HTML]{FCE4D5}{\textbf{{68.8}}} & \cellcolor[HTML]{FFD4D3}{\textbf{{82.3}}} & \cellcolor[HTML]{FFD4D3}{\textbf{{65.0}}} & \cellcolor[HTML]{FFD4D3}{\textbf{{67.0}}} & \multicolumn{1}{>{\centering\arraybackslash}m{3.1em}!{\vrule width 1pt}}{\cellcolor[HTML]{EAECEF}{\underline{\textbf{{70.4}}}}} & \cellcolor[HTML]{FFD4D3}{\textbf{{24.9}}} & \cellcolor[HTML]{FFD4D3}{\textbf{{39.0}}} & \cellcolor[HTML]{FFD4D3}{\textbf{{34.7}}} & \underline{\textbf{{31.9}}} \\
    \bottomrule
    \end{tabularx}%
    }
    \vspace{0.35ex}
    \noindent\begin{minipage}{\linewidth}
        \footnotesize
        \textbf{Note.} Other model results are reported in Table~\ref{tab:app_main_qwen}, ~\ref{tab:app_main_mixtral}.
        Cells colored with rankings as \rankchip{FFD4D3}{1st}, \rankchip{FCE4D5}{2nd}, and \rankchip{FFFCE0}{3rd}.
    \end{minipage}
    \vspace{-7mm}
    \end{table*}

    \section{Experiments}
    \label{sec:exp}
    \vspace{-2mm}

    \subsection{Experimental Setup}
    \label{sec:setup}

    \paragraph{Backbone models.}
    We evaluate SoT on three frozen \emph{LLMs} with different scales and architectures (Llama-3.1-8B \citep{grattafiori2024llama3}, Qwen2.5-14B \citep{yang2024qwen25}, and Mixtral-8x7B \citep{jiang2024mixtral})
    {and transfer to two frozen VLM scales, Qwen2.5-VL-7B and Qwen2.5-VL-32B \citep{bai2025qwen25vl}.} Full details are deferred to Appendix~\ref{app:model_protocol}.

    \paragraph{Tasks and evaluation domains.}
    For \emph{LLMs}, we organize evaluation into \emph{four reasoning domains} comprising \emph{16 benchmarks}.
    \textbf{Quantitative Reasoning} (QS) includes GSM8K (GSM) \citep{cobbe2021gsm8k}, MATH \citep{hendrycks2021measuring}, and DROP \citep{dua2019drop}.
    \textbf{General Understanding} (GU) includes CommonsenseQA (CSQA) \citep{talmor2019commonsenseqa}, StrategyQA (StrQA) \citep{geva2021did}, BoolQ \citep{clark2019boolq}, MMLU \citep{hendrycks2021measuringmassive}, and RACE \citep{lai2017race}.
    \textbf{Symbolic and Code} (S\&C) includes FOLIO (FOL) \citep{han2024folio}, ProofWriter (PW) \citep{tafjord2021proofwriter}, BBH-Temporal (BBH) \citep{srivastava2023beyond}, HumanEval (HE) \citep{chen2021evaluating}, and MBPP \citep{austin2021program}.
    \textbf{Long-Context Reasoning} (LCR) includes HotpotQA (HQA) \citep{yang2018hotpotqa}, NarrativeQA (NarQA) \citep{kocisky2018narrativeqa}, and LongBench MultiFieldQA (LB) \citep{bai2024longbench}.
    {For \emph{VLM transfer}, we evaluate \emph{three reasoning-oriented benchmarks} that require knowledge-grounded, diagrammatic, or multimodal chain-of-thought inference: A-OKVQA \citep{schwenk2022aokvqa}, AI2D \citep{kembhavi2016ai2d}, and M$^3$CoT \citep{chen2024m3cot}.}
    Full details are deferred to Appendix~\ref{app:dataset_setup}.

    \paragraph{Baselines.}
    We compare SoT against \emph{5 baseline families}.
    \textbf{Greedy:} Vanilla.
    \textbf{Reasoning paradigm:} Chain-of-Thought (CoT) \citep{wei2022chain}, Plan-and-Solve (PS) \citep{wang2023plan}, Self-Refine (SR) \citep{madaan2023self} with externally scripted programs, and Self-Consistency (SC) \citep{wang2023selfconsistency}, Best-of-N (BoN) \citep{snell2024scaling}, Constrained Beam (CB) \citep{hokamp2017lexically}, Monte Carlo Tree Search (MCTS) \citep{xie2024mcts} with sampling or search methods.
    \textbf{Memory-oriented methods:} H$_2$O (H2O) \citep{zhang2023h2o}, SnapKV (SNAP) \citep{li2024snapkv}, and StreamingLLM (STREAM) \citep{xiao2023streamingllm}.
    \textbf{Latent reasoning:} COCONUT (COCO) \citep{hao2025training}.
    \textbf{RL-based reasoning:} \,{GRPO-inspired soft prompt (GRPO-SP)} \citep{shao2024deepseekmath}.
    {All learned components are fitted exclusively on training-side data, with evaluation examples excluded.} Full details are deferred to Appendix~\ref{app:baseline_methods} and ~\ref{app:hyperparameters}.

    \begin{wrapfigure}{r}{0.55\columnwidth}
        \vspace{-4.0mm}
        \centering
        \includegraphics[width=0.53\textwidth]{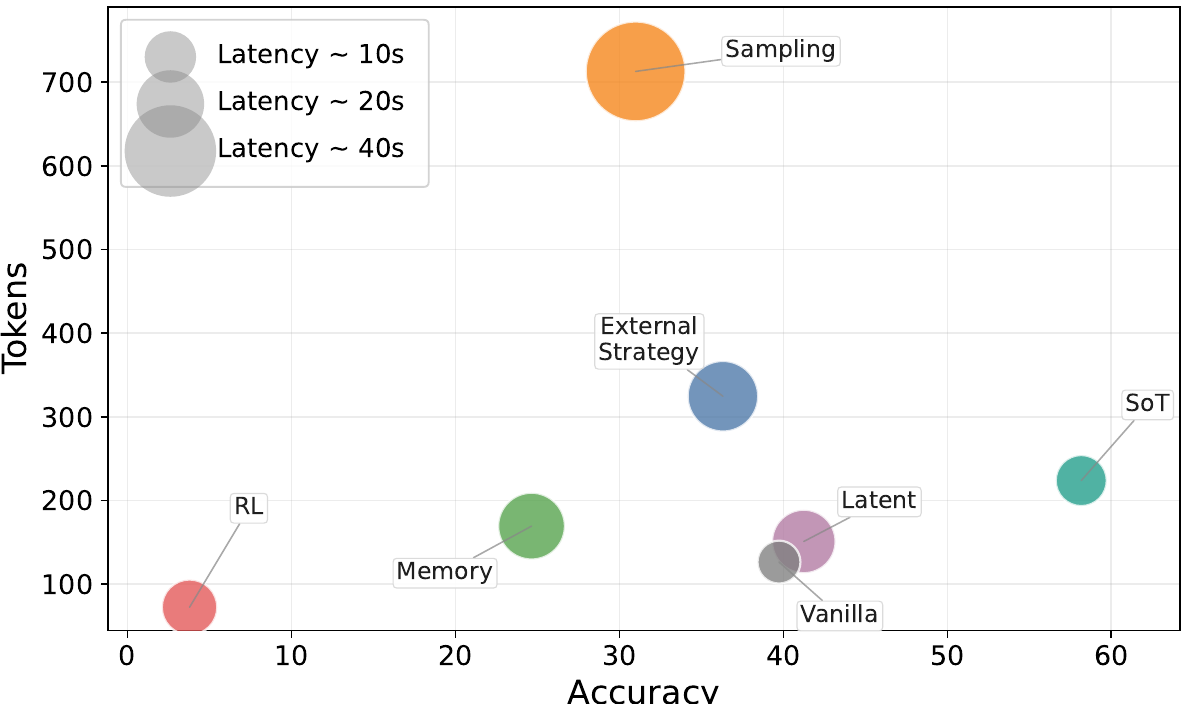}
        \caption{\textbf{Accuracy--efficiency trade-off} on Llama-3.1-8B with breakdown and other models in Figure~\ref{fig:app_tradeoff_all}.}
        \label{fig:acc_eff_tradeoff}
        \vspace{-12mm}
    \end{wrapfigure}

    \paragraph{Evaluation metrics.}
    We report task-standard accuracy metrics and efficiency cost.
    \textbf{Exact Match (EM)} is used for GSM8K, MATH, RACE, MMLU, CommonsenseQA, FOL, BBH, ProofWriter, BoolQ, StrategyQA, {A-OKVQA, AI2D, and M$^3$CoT}.
    \textbf{F1} is used for HotpotQA, DROP, NarrativeQA, and LongBench MultiFieldQA.
    \textbf{Pass@1} is used for HumanEval and MBPP.
    {\textbf{Efficiency} is measured by generated tokens and end-to-end inference latency; hardware and timing details are provided in Appendix~\ref{app:model_protocol}.}

    \begin{table*}[t]
    \centering
    \begin{minipage}[b]{0.75\textwidth}
        \centering
        \captionof{table}{\textbf{Efficiency comparison on Llama-3.1-8B.}}
        \label{tab:efficiency}
        \vspace{-2mm}
        \footnotesize
        \begingroup
        \setlength{\tabcolsep}{0.80pt}%
        \setlength{\aboverulesep}{0pt}%
        \setlength{\belowrulesep}{0pt}%
        \setlength{\extrarowheight}{0pt}%
        \renewcommand{\arraystretch}{1.1}%
        \begin{tabular}{@{}
            >{\centering\arraybackslash}p{1.32cm}
            >{\centering\arraybackslash}m{4.20em}
            !{\vrule width 0.4pt}
            >{\centering\arraybackslash}m{2.94em}
            >{\centering\arraybackslash}m{2.94em}
            !{\vrule width 1pt}
            >{\centering\arraybackslash}m{2.94em}
            >{\centering\arraybackslash}m{2.94em}
            !{\vrule width 1pt}
            >{\centering\arraybackslash}m{2.94em}
            >{\centering\arraybackslash}m{2.94em}
            !{\vrule width 1pt}
            >{\centering\arraybackslash}m{2.94em}
            >{\centering\arraybackslash}m{2.94em}
            @{}}
        \toprule
        \multirow{2}{=}{\centering\textbf{Category}} & \multirow{2}{=}{\centering\textbf{Method}}
        & \multicolumn{2}{>{\cellcolor[HTML]{E3F2FD}}c!{\vrule width 1pt}}{\textbf{QS}}
        & \multicolumn{2}{>{\cellcolor[HTML]{E3F2FD}}c!{\vrule width 1pt}}{\textbf{S\&C}}
        & \multicolumn{2}{>{\cellcolor[HTML]{E3F2FD}}c!{\vrule width 1pt}}{\textbf{GU}}
        & \multicolumn{2}{>{\cellcolor[HTML]{E3F2FD}}c}{\textbf{LCR}} \\
        \cmidrule(lr){3-4}\cmidrule(lr){5-6}\cmidrule(lr){7-8}\cmidrule(lr){9-10}
        & & {\small\textbf{Tok.}} & {\small\textbf{Lat.}} & {\small\textbf{Tok.}} & {\small\textbf{Lat.}} & {\small\textbf{Tok.}} & {\small\textbf{Lat.}} & {\small\textbf{Tok.}} & {\small\textbf{Lat.}} \\
        \midrule
        \multirow{6}{=}{\centering Reasoning\\Paradigm}
        & CoT  & 271.6 & 17.5 & 369.8 & 27.5 & 212.9 & 12.8 & 198.1 & \cellcolor[HTML]{FFFCE0}{19.6} \\
        & PS   & 314.2 & 20.7 & 372.1 & 26.8 & 319.8 & 22.2 & 282.3 & 33.7 \\
        & SR   & 368.5 & 25.9 & 391.7 & 28.8 & 292.2 & 19.2 & 278.8 & 33.2 \\
        & SC   & 736.8 & 43.1 & 953.9 & 63.5 & 626.5 & 37.6 & 533.8 & 48.9 \\
        & CB   & 423.9 & 22.0 & 363.9 & \cellcolor[HTML]{FCE4D5}{19.4} & 341.7 & 16.8 & 283.7 & \cellcolor[HTML]{FCE4D5}{19.5} \\
        & MCTS & 422.1 & 21.4 & 364.7 & \cellcolor[HTML]{FCE4D5}{19.4} & 335.2 & 16.5 & 286.6 & 20.5 \\
        \midrule
        \multirow{3}{=}{\centering Memory-\\Oriented}
        & H2O    & 226.0 & 16.7 & \cellcolor[HTML]{FFFCE0}{264.4} & 34.6 & 185.6 & 15.8 & \cellcolor[HTML]{FFFCE0}{{164.6}} & 21.5 \\
        & SNAP   & 226.0 & 16.6 & \cellcolor[HTML]{FFFCE0}{264.4} & 34.8 & 185.6 & 16.1 & \cellcolor[HTML]{FFFCE0}{{164.6}} & 20.9 \\
        & {STREAM} & \cellcolor[HTML]{FFD4D3}{92.6} & \cellcolor[HTML]{FFFCE0}{12.5} & \cellcolor[HTML]{FFD4D3}{64.2} & 24.0 & \cellcolor[HTML]{FFFCE0}{{93.4}} & \cellcolor[HTML]{FFFCE0}{9.8} & \cellcolor[HTML]{FFD4D3}{{103.1}} & \cellcolor[HTML]{FFD4D3}{10.3} \\
        \midrule
        Latent & COCO & \cellcolor[HTML]{FCE4D5}{171.5} & \cellcolor[HTML]{FCE4D5}{11.5} & \cellcolor[HTML]{FCE4D5}{214.6} & \cellcolor[HTML]{FFFCE0}{20.4} & \cellcolor[HTML]{FCE4D5}{87.2} & \cellcolor[HTML]{FCE4D5}{6.7} & \cellcolor[HTML]{FCE4D5}{131.5} & 30.3 \\
        \midrule
        RL-Based & {\scriptsize GRPO-SP} & 73.1 & 7.5 & 69.7 & 7.1 & 71.4 & 7.5 & 76.2 & 26.9 \\
        \midrule
        Ours & \textbf{SoT} & \textbf{{255.7}} & \cellcolor[HTML]{FFD4D3}{\textbf{5.8}} & \textbf{{302.2}} & \cellcolor[HTML]{FFD4D3}{\textbf{9.5}} & \cellcolor[HTML]{FFD4D3}{\textbf{{74.9}}} & \cellcolor[HTML]{FFD4D3}{\textbf{1.8}} & \textbf{{263.0}} & \textbf{22.0} \\
        \bottomrule
        \end{tabular}%
        \endgroup
        \par\vspace{1.15ex}
    \end{minipage}
    \hfill
    \begin{minipage}[b]{0.24\textwidth}
        \centering
        \includegraphics[width=0.88\textwidth]{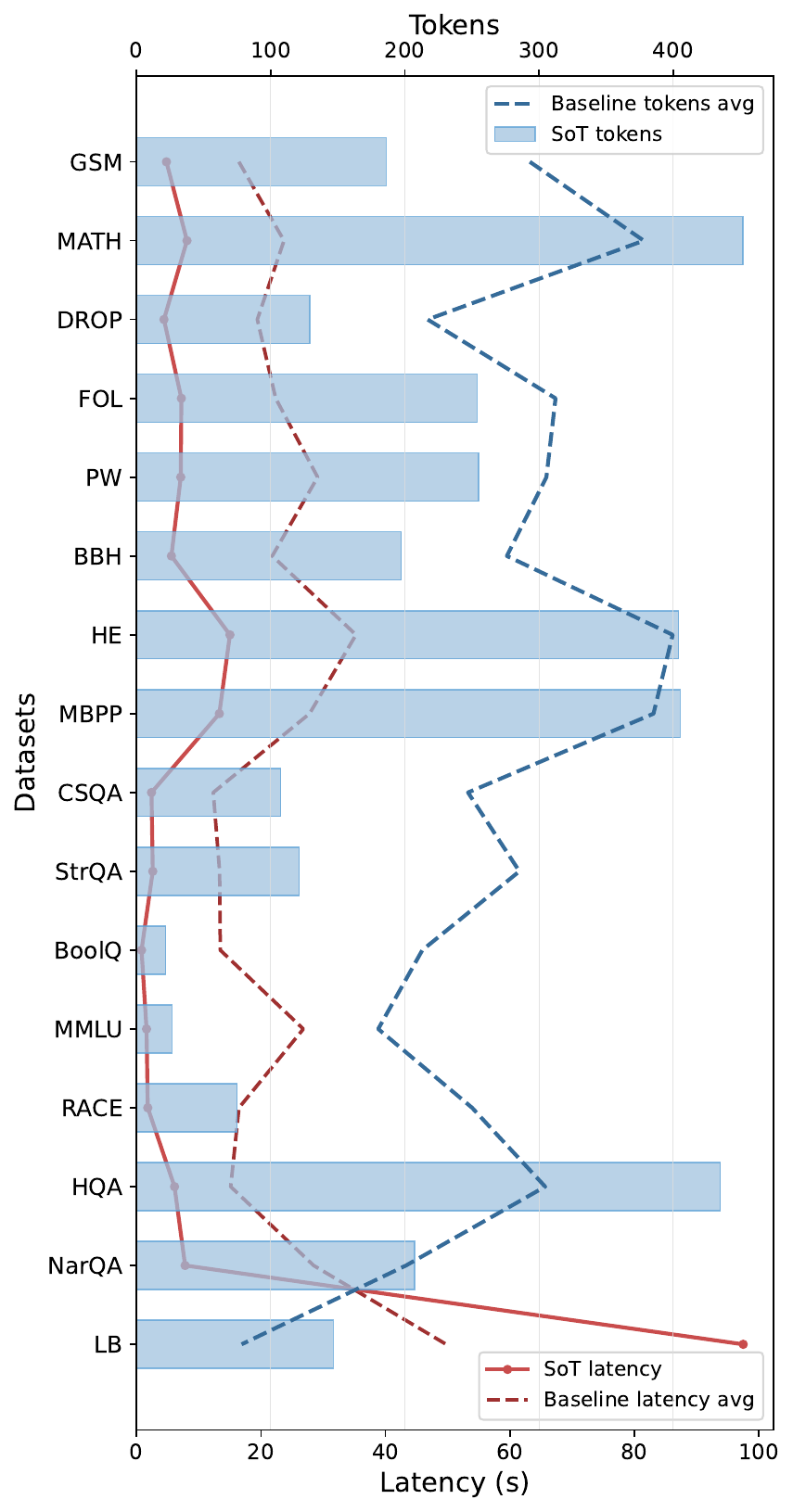}
        \vspace{-2mm}
        \captionof{figure}{\textbf{Distribution.}}
        \vspace{1mm}
        \label{fig:efficiency_datasets}
    \end{minipage}

    \vspace{-1mm}
    \begin{minipage}{\textwidth}
        \footnotesize
        \noindent\textbf{Note.}
        {Generated reasoning tokens (\textbf{Tok.}) and end-to-end latency (\textbf{Lat.}) are aggregated over four reasoning domains, with full breakdowns in Tables~\ref{tab:app_tok_llama} and~\ref{tab:app_lat_llama}. Results for the other models appear in Tables~\ref{tab:app_tok_qwen}, \ref{tab:app_lat_qwen}, \ref{tab:app_tok_mixtral}, and~\ref{tab:app_lat_mixtral}.}
    \end{minipage}
    \vspace{-4mm}
\end{table*}

    \vspace{-1mm}
    \subsection{Main Results}
    \label{sec:main_results}
    \vspace{-0.5mm}

    \paragraph{Overall performance.}
    Table~\ref{tab:main_results_llama} reports the main results on Llama-3.1-8B across four reasoning domains.
    SoT achieves the best domain-level average in all domains.
    Compared with the strongest non-SoT baseline in each domain, it improves Quantitative Reasoning, Symbolic and Code, General Understanding, and Long-Context Reasoning by \({+10.3/+15.9/+6.8/+10.1}\) points, corresponding to relative gains of \({18.6\%/37.4\%/10.7\%/46.3\%}\).
    It also obtains the best or tied-best result on {13} out of 16 datasets, with an average domain-level gain of \({10.8}\) points over the strongest competing methods.
    This pattern suggests that reasoning gains need \emph{not} come primarily from prescribing a stronger external reasoning program or from scaling search over more trajectories.
    A closed-loop controller driven by the model's own endogenous reasoning state can itself serve as an effective mechanism for improving test-time reasoning.
    We further demonstrate the stable performance gains of our method through quality sensitivity experiments on the training dataset in the Appendix~\ref{app:traj_quality}.

    \vspace{-2mm}
    \paragraph{Generalization.}
    A distinctive property of SoT is that the same controller can be applied across heterogeneous reasoning structures with stable performance.
    Compared with scripted reasoning, SoT avoids committing to a fixed external reasoning form; compared with sampling and search, it improves reasoning without relying mainly on broader trajectory expansion; compared with memory-oriented and latent reasoning baselines, it organizes historical evidence through the model's evolving endogenous state rather than generic context heuristics or hidden-space-only reasoning.
    Consistent gains across these regimes therefore indicate \emph{mechanism-level generalization}: SoT transfers as a reasoning principle, not merely as a benchmark-specific strategy.

    In general, these patterns suggest a more structural interpretation of the results.
    Within the family of externally prescribed token-chain reasoning methods, performance gains are typically obtained by strengthening scripts, increasing samples, or enlarging search.
    By contrast, SoT improves reasoning by changing the control variable itself, shifting from external token programs to endogenous state-conditioned evidence organization. This may explain why it can surpass strong token-chain baselines across heterogeneous regimes.
    Appendix~\ref{app:theory_policy_class} to~\ref{app:theory_stop} provide in-depth theoretical analysis.

    \vspace{-1.0mm}

    \subsection{Efficiency and Trade-off}
    \label{sec:efficiency}
    \vspace{-0.5mm}

    Efficiency is central to inference-time reasoning because most performance gains come from expanding computation through longer trajectories, repeated sampling, or explicit search. Instead, SoT activates only the evidence context most suitable under the current endogenous state, so its advantage should appear not only in accuracy but also in efficiency.

    \vspace{-1.0mm}
    \paragraph{Efficiency.}
    Table~\ref{tab:efficiency} and Figure~\ref{fig:efficiency_datasets} report generated reasoning tokens and end-to-end latency on Llama-3.1-8B. Averaged over the four reasoning domains, SoT uses \({223.9}\) tokens and \(9.8\)s latency, compared with \(472.7\) tokens and \(29.0\)s latency for sampling/search methods, reducing tokens by \({52.6\%}\) and latency by \(66.4\%\). The saving is especially clear in QS, S\&C, and GU, where SoT attains strong accuracy with only \(5.8\)s, \(9.5\)s, and \(1.8\)s latency, respectively. Compared with memory-oriented methods, however, SoT is not merely a more aggressive compression rule: generic pruning can reduce tokens, but often sacrifices accuracy because it does not condition retained evidence on the current reasoning state. Thus, SoT improves efficiency by removing unnecessary computation while preserving the support needed for the next decision.

    \vspace{-1.0mm}
    \paragraph{Accuracy--efficiency trade-off.}
    Figure~\ref{fig:acc_eff_tradeoff} further shows that SoT moves the accuracy--efficiency frontier rather than simply trading accuracy for shorter outputs. It reaches the highest accuracy while using far fewer tokens and lower latency than external strategies and sampling/search baselines. This pattern clarifies the core advantage of SoT: test-time compute is no longer spent on uniformly longer chains, larger candidate sets, or generic history retention, but is redirected toward state-conditioned evidence organization. The efficiency results therefore reinforce the main claim that endogenous reasoning state provides a better control signal for both reasoning quality and inference cost.

   \begin{table*}[t]
    \centering
    \caption{\textbf{Ablation results on Llama-3.1-8B,} with other models provided in {Tables~\ref{tab:app_ablation_qwen},~\ref{tab:app_ablation_mixtral}}.}
    \vspace{-2mm}
    \label{tab:ablation}
    \footnotesize
    \setlength{\tabcolsep}{1.75pt}
    \setlength{\aboverulesep}{0pt}%
    \setlength{\belowrulesep}{0pt}%
    \setlength{\extrarowheight}{0.75pt}
    \renewcommand{\arraystretch}{1.0}%
    \begin{tabular}{l%
        @{\kern0.42em}!{\vrule width 0.4pt}@{\kern0.42em}%
        ccc%
        !{\vrule width 1pt}%
        ccc%
        !{\vrule width 1pt}%
        ccc%
        !{\vrule width 1pt}%
        ccc}
    \toprule
    \multirow{2}{*}{\textbf{Variant}}
    & \multicolumn{3}{>{\columncolor[HTML]{E3F2FD}[\tabcolsep][\tabcolsep]}c!{\vrule width 1pt}}{\textbf{QS}}
    & \multicolumn{3}{>{\columncolor[HTML]{E3F2FD}[\tabcolsep][\tabcolsep]}c!{\vrule width 1pt}}{\textbf{S\&C}}
    & \multicolumn{3}{>{\columncolor[HTML]{E3F2FD}[\tabcolsep][\tabcolsep]}c!{\vrule width 1pt}}{\textbf{GU}}
    & \multicolumn{3}{>{\columncolor[HTML]{E3F2FD}[\tabcolsep][\tabcolsep]}c}{\textbf{LCR}} \\
    \cmidrule(lr){2-4} \cmidrule(lr){5-7} \cmidrule(lr){8-10} \cmidrule(lr){11-13}
    & \textbf{acc.} & \textbf{tok.} & \textbf{lat.}
    & \textbf{acc.} & \textbf{tok.} & \textbf{lat.}
    & \textbf{acc.} & \textbf{tok.} & \textbf{lat.}
    & \textbf{acc.} & \textbf{tok.} & \textbf{lat.} \\
    \midrule
    Full SoT & 62.8 & 219.4 & 5.8 & 54.5 & 294.0 & 9.5 & 71.2 & 64.0 & 1.8 & 33.0 & 181.6 & 22.0 \\
    + Threshold Tuning & 62.8 & 219.4 & 5.8 & 54.3 & 294.0 & 9.5 & 71.2 & 64.0 & 1.8 & 31.8 & 229.0 & 22.0 \\
    \midrule
    w/o Evid. Org. \(\mathcal{S}\) & 29.6 & 469.5 & 12.9 & 44.5 & 536.2 & 16.5 & \cellcolor[HTML]{EAECEF}{72.4} & 193.9 & 4.9 & 26.9 & 506.5 & \cellcolor[HTML]{EAECEF}{14.1} \\
    w/o Stopping \(\mathcal{T}\) & 60.6 & 1769.6 & 80.7 & \cellcolor[HTML]{EAECEF}{57.1} & 1705.1 & 110.3 & 70.2 & 4148.5 & 99.5 & 31.6 & 2937.4 & 157.3 \\
    w/o Geometry \(\delta_t\) & 42.5 & \cellcolor[HTML]{EAECEF}{68.6} & \cellcolor[HTML]{EAECEF}{3.3} & 46.4 & \cellcolor[HTML]{EAECEF}{48.0} & \cellcolor[HTML]{EAECEF}{2.9} & \cellcolor[HTML]{EAECEF}{73.0} & \cellcolor[HTML]{EAECEF}{25.4} & \cellcolor[HTML]{EAECEF}{0.9} & 24.5 & \cellcolor[HTML]{EAECEF}{47.2} & \cellcolor[HTML]{EAECEF}{2.2} \\
    w/o Dynamics \((v_t,c_t)\) & 57.0 & 1039.2 & 29.1 & 46.2 & 846.7 & 25.7 & 71.2 & 1272.3 & 33.7 & 31.6 & 1000.3 & 30.8 \\
    w/o Uncertainty \(H_t\) & 54.1 & 244.9 & 6.4 & 52.1 & \cellcolor[HTML]{EAECEF}{281.8} & \cellcolor[HTML]{EAECEF}{8.8} & \cellcolor[HTML]{EAECEF}{73.6} & \cellcolor[HTML]{EAECEF}{61.3} & 1.9 & 32.8 & 555.5 & \cellcolor[HTML]{EAECEF}{16.6} \\
    w/o Sent.-level Units & 54.1 & 244.9 & 7.2 & 52.1 & \cellcolor[HTML]{EAECEF}{281.8} & \cellcolor[HTML]{EAECEF}{8.8} & \cellcolor[HTML]{EAECEF}{73.6} & \cellcolor[HTML]{EAECEF}{61.3} & 1.9 & \cellcolor[HTML]{EAECEF}{35.1} & 224.1 & \cellcolor[HTML]{EAECEF}{11.9} \\
    \bottomrule
    \end{tabular}
    \vspace{-2mm}
\end{table*}

    \vspace{-1mm}
    \subsection{Ablation Studies}
    \label{sec:ablation}
    \vspace{-1.0mm}

    The ablations in Table~\ref{tab:ablation} indicate that the components of SoT interact in a way that controls distinct reasoning failure modes rather than acting as independent optimizations. For instance, removing the evidence-organization operator \(S\) leads to a significant drop in performance, especially in QS (from 62.8 to 29.6) and LCR (from 33.0 to 26.9), highlighting the critical role of state-conditioned evidence selection. Conversely, removing the stopping operator \(T\) improves S\&C (from 54.5 to 57.1), but this comes at the expense of drastically increased tokens (from 189.8 to 2640.2) and latency (from 9.8s to 112.0s), making the improvement unsustainable due to excessive computation. Ablating state components such as geometry \(\delta_t\), dynamics \((v_t, c_t)\), and uncertainty \(H_t\) shows that they contribute complementary signals: removing geometry \(\delta_t\) reduces performance (from 62.8 to 42.5 in QS), while removing dynamics \((v_t, c_t)\) increases computational cost and leads to a drop in LCR accuracy (from 33.0 to 31.6). Removing uncertainty \(H_t\) results in only a minor performance change in most domains (from 54.5 to 52.1 in S\&C), but still weakens the overall accuracy--cost balance. Finally, replacing the sentence-level reasoning units does not significantly improve performance, but suggests that the default sentence-level granularity provides stable control across domains. SoT's strength lies in the synergy between its components, rather than in any single one.

    \subsection{Transfer to Vision-Language Reasoning}
    \label{sec:vlm_results}
    \vspace{-1.0mm}

    {
    \begin{table*}[t]
    \centering
    \caption{{\textbf{Vision--language reasoning across scales} (accuracy, \%; completion tokens; latency, s).}}
    \vspace{-2mm}
    \label{tab:vlm_scale_transfer}
    \footnotesize
    \setlength{\tabcolsep}{2.5pt}
    \setlength{\aboverulesep}{0pt}
    \setlength{\belowrulesep}{0pt}
    \setlength{\extrarowheight}{0.55pt}
    \begin{tabular}{@{}l!{\vrule width 0.4pt}ccc!{\vrule width 0.8pt}ccc!{\vrule width 0.8pt}ccc!{\vrule width 1.3pt}ccc!{\vrule width 0.8pt}ccc!{\vrule width 0.8pt}ccc@{}}
    \toprule
    \multirow{3}{*}{\textbf{Method}}
      & \multicolumn{9}{>{\cellcolor[HTML]{E3F2FD}}c!{\vrule width 1.3pt}}{\textbf{Qwen2.5-VL-7B}}
      & \multicolumn{9}{>{\cellcolor[HTML]{E3F2FD}}c}{\textbf{Qwen2.5-VL-32B}} \\
    \cmidrule(lr){2-10}\cmidrule(lr){11-19}
      & \multicolumn{3}{>{\cellcolor[HTML]{E3F2FD}}c!{\vrule width 0.8pt}}{\textbf{A-OKVQA}}
      & \multicolumn{3}{>{\cellcolor[HTML]{E3F2FD}}c!{\vrule width 0.8pt}}{\textbf{AI2D}}
      & \multicolumn{3}{>{\cellcolor[HTML]{E3F2FD}}c!{\vrule width 1.3pt}}{\textbf{M$^3$CoT}}
      & \multicolumn{3}{>{\cellcolor[HTML]{E3F2FD}}c!{\vrule width 0.8pt}}{\textbf{A-OKVQA}}
      & \multicolumn{3}{>{\cellcolor[HTML]{E3F2FD}}c!{\vrule width 0.8pt}}{\textbf{AI2D}}
      & \multicolumn{3}{>{\cellcolor[HTML]{E3F2FD}}c}{\textbf{M$^3$CoT}} \\
    \cmidrule(lr){2-4}\cmidrule(lr){5-7}\cmidrule(lr){8-10}
    \cmidrule(lr){11-13}\cmidrule(lr){14-16}\cmidrule(lr){17-19}
      & \textbf{acc.} & \textbf{tok.} & \textbf{lat.}
      & \textbf{acc.} & \textbf{tok.} & \textbf{lat.}
      & \textbf{acc.} & \textbf{tok.} & \textbf{lat.}
      & \textbf{acc.} & \textbf{tok.} & \textbf{lat.}
      & \textbf{acc.} & \textbf{tok.} & \textbf{lat.}
      & \textbf{acc.} & \textbf{tok.} & \textbf{lat.} \\
    \midrule
    CoT
      & 81.0 & \cellcolor[HTML]{FFD4D3}137 & \cellcolor[HTML]{FFD4D3}3.0
      & 82.7 & \cellcolor[HTML]{FFD4D3}189 & \cellcolor[HTML]{FFD4D3}3.9
      & 80.7 & \cellcolor[HTML]{FFD4D3}178 & \cellcolor[HTML]{FFD4D3}4.4
      & 84.0 & \cellcolor[HTML]{FFD4D3}267 & \cellcolor[HTML]{FFD4D3}12.8
      & \cellcolor[HTML]{FFFCE0}87.3 & \cellcolor[HTML]{FFD4D3}321 & \cellcolor[HTML]{FFD4D3}17.0
      & \cellcolor[HTML]{FCE4D5}87.3 & \cellcolor[HTML]{FFD4D3}329 & \cellcolor[HTML]{FFD4D3}16.0 \\
    PS
      & \cellcolor[HTML]{FFFCE0}84.0 & \cellcolor[HTML]{FCE4D5}169 & \cellcolor[HTML]{FCE4D5}3.5
      & 82.7 & \cellcolor[HTML]{FFFCE0}245 & \cellcolor[HTML]{FFFCE0}5.9
      & 81.3 & \cellcolor[HTML]{FFFCE0}211 & \cellcolor[HTML]{FFFCE0}5.8
      & 80.5 & \cellcolor[HTML]{FCE4D5}311 & \cellcolor[HTML]{FCE4D5}16.9
      & 72.7 & \cellcolor[HTML]{FFFCE0}436 & 27.1
      & 78.7 & \cellcolor[HTML]{FCE4D5}387 & \cellcolor[HTML]{FCE4D5}17.9 \\
    SR
      & 78.0 & 213 & 5.1
      & 80.0 & 264 & 6.1
      & 83.3 & 254 & 6.4
      & \cellcolor[HTML]{FFFCE0}84.5 & \cellcolor[HTML]{FFFCE0}414 & \cellcolor[HTML]{FFFCE0}22.2
      & \cellcolor[HTML]{FCE4D5}89.3 & 465 & \cellcolor[HTML]{FFFCE0}26.1
      & 84.0 & 469 & 24.3 \\
    SC
      & 83.5 & 691 & 15.9
      & \cellcolor[HTML]{FCE4D5}87.3 & 955 & 19.9
      & \cellcolor[HTML]{FFFCE0}84.0 & 909 & 21.3
      & 84.0 & 1319 & 64.0
      & 86.0 & 1669 & 87.6
      & \cellcolor[HTML]{FCE4D5}87.3 & 1618 & 78.2 \\
    BoN
      & \cellcolor[HTML]{FCE4D5}84.5 & 691 & 14.9
      & \cellcolor[HTML]{FFD4D3}\textbf{88.7} & 961 & 21.7
      & \cellcolor[HTML]{FCE4D5}86.7 & 919 & 21.0
      & \cellcolor[HTML]{FCE4D5}85.0 & 1322 & 65.5
      & \cellcolor[HTML]{FFFCE0}87.3 & 1680 & 88.8
      & \cellcolor[HTML]{FFFCE0}85.3 & 1633 & 80.0 \\
    \textbf{SoT}
      & \cellcolor[HTML]{FFD4D3}\textbf{87.5} & \cellcolor[HTML]{FFFCE0}191 & \cellcolor[HTML]{FFFCE0}4.4
      & \cellcolor[HTML]{FFFCE0}86.7 & \cellcolor[HTML]{FCE4D5}215 & \cellcolor[HTML]{FCE4D5}5.4
      & \cellcolor[HTML]{FFD4D3}\textbf{87.3} & \cellcolor[HTML]{FCE4D5}198 & \cellcolor[HTML]{FCE4D5}4.5
      & \cellcolor[HTML]{FFD4D3}\textbf{85.5} & 425 & 23.2
      & \cellcolor[HTML]{FFD4D3}\textbf{90.0} & \cellcolor[HTML]{FCE4D5}364 & \cellcolor[HTML]{FCE4D5}19.2
      & \cellcolor[HTML]{FFD4D3}\textbf{88.0} & \cellcolor[HTML]{FFFCE0}409 & \cellcolor[HTML]{FFFCE0}20.1 \\
    \bottomrule
    \end{tabular}
    \end{table*}

    {Across these 6 model--reasoning-task settings, SoT averages $87.5\%$ accuracy and is best in 5. On 7B AI2D, it is $2.0$ points below BoN but uses $4.5\times$ fewer completion tokens. At 32B, it exceeds the strongest alternative by $0.5$--$0.7$ points on all 3 tasks while using $3.1$--$4.6\times$ fewer tokens than search-based methods. Across both scales, it also reduces end-to-end latency by $73.5\%$ relative to SC and BoN. The gains across knowledge-grounded, diagrammatic, and multimodal chain-of-thought tasks show that state-conditioned evidence organization transfers beyond text-only reasoning and persists across scale without expanding the controller or relying on wider sampling.}
    }

    \vspace{-1.0mm}
    \subsection{Case study}
    \label{sec:case_study}
    \vspace{-0.5mm}

    Figure~\ref{fig:app_case} shows a representative example with activated historical evidence at each step, and the corresponding stop probability.
    SoT does not keep the full trajectory active, nor does it rely on a fixed external reasoning template.
    Instead, as constraints are progressively resolved, the endogenous state selectively retains the evidence still needed for the next decision, suppresses obsolete support, and raises stop readiness only when the remaining context becomes sufficient.
    {This example therefore concretely illustrates the closed-loop view of SoT: state organizes support, support shapes the next step, and the updated step in turn changes the state.}

    \paragraph{Overall findings.}
    Across accuracy, efficiency, ablations, scale transfer, and case evidence, SoT improves reasoning by reallocating existing inference computation rather than enlarging search or backbone capacity.
    Its endogenous state acts as a compact control interface that couples evidence retention with termination, translating continuous internal evolution into discrete, inspectable decisions.
    This shared mechanism explains why the same lightweight design advances the accuracy--efficiency frontier across heterogeneous text tasks and multimodal scales.

%% file: chapters/4-explore.tex
\providecommand{\AppPlaceholder}[2]{%
    \fbox{
        \begin{minipage}[c][#1][c]{0.94\textwidth}
            \centering
            \textit{#2}
        \end{minipage}
    }
}

\begin{figure*}[t]
\centering
\vspace{-2mm}
\includegraphics[width=0.98\textwidth]{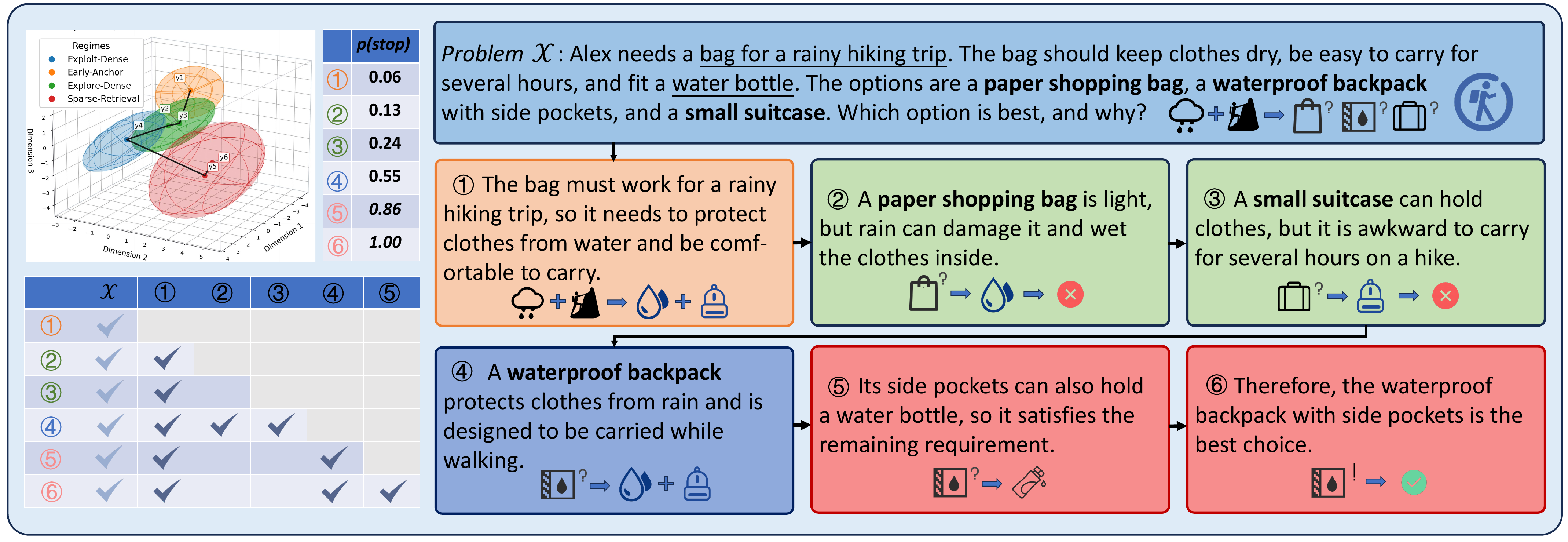}
\vspace{-2mm}
\caption{\textbf{A case study of SoT,} with more case studies in Appendix~\ref{app:case_studies}.}
\label{fig:app_case}
\vspace{-5mm}
\end{figure*}

    \vspace{-1mm}
    \section{Exploratory Extensions of State of Thought}
    \label{sec:explore}
    \vspace{-1mm}

    \subsection{{SoT under Limited Access}}
    \label{sec:training_free_sot}
    \vspace{-0.5mm}
    \paragraph{Training-Free SoT.}
    We first ask \emph{whether SoT requires learned control at all}. We instantiate the evidence-organization operator \(\mathcal{S}\) and stopping operator \(\mathcal{T}\) without training, using fixed rules derived from the same dynamics-geometric state.
    Historical evidence is selected by state compatibility with the current reasoning regime, while stopping is triggered when the trajectory becomes sufficiently stable and low-uncertainty.
    This yields a fully training-free SoT controller that preserves the same closed-loop formulation while removing learned control entirely.

    \vspace{-0.5mm}
    \paragraph{SoT-Embed under Limited Access.}
    We next ask \emph{whether SoT depends on privileged access to internal information transfer}.
    We embed the model's multi-step segmented output with a fixed encoder and feed the embedding trajectory into the control loop in place of the internal state.
    {This tests whether endogenous reasoning control remains viable when its observable interface is restricted to text.}

    SoT-Training-free evaluates whether the benefits of SoT come solely from the learned controller, while SoT-Embed tests whether it relies on white-box access.
    Table~\ref{tab:boundary_sot} summarizes that both variants maintain competitive performance, with macro-averaged accuracy lower relative to full SoT by approximately \({17.5\%}\) for SoT-Training-free and \({17.9\%}\) for SoT-Embed. In comparison, the Top-3 baseline methods show an average accuracy of \(47.5\%\) on those cells, with SoT variants achieving comparable or slightly better results in many tasks. These results suggest that SoT's core advantage lies in the principle of state-conditioned evidence organization, not in a specific implementation. Despite these reductions relative to full SoT, both variants remain highly competitive, often outperforming the Top-3 baselines on this Llama table. For further details, see Appendix~\ref{app:limitations_sot}.

    \begin{table*}[t]
    \centering
    \vspace{-1mm}
    \caption{\textbf{Boundary studies of SoT on Llama-3.1-8B,} with other models provided in {Tables~\ref{tab:app_boundary_qwen},~\ref{tab:app_boundary_mixtral}}.}
    \vspace{-2mm}
    \label{tab:boundary_sot}
    \footnotesize
    \setlength{\tabcolsep}{5pt}
    \renewcommand{\arraystretch}{1.05}
    \begingroup
    \setlength{\aboverulesep}{0pt}%
    \setlength{\belowrulesep}{0pt}%
    \setlength{\extrarowheight}{0.95pt}%
    \begin{tabular}{l!{\vrule width 0.4pt}ccc>{\columncolor[HTML]{EAECEF}\centering\arraybackslash}c!{\vrule width 1pt}ccccc>{\columncolor[HTML]{EAECEF}\centering\arraybackslash}c}
    \toprule
    & \multicolumn{4}{>{\cellcolor[HTML]{E3F2FD}}c!{\vrule width 1pt}}{\textbf{Quantitative Reasoning}} & \multicolumn{6}{>{\cellcolor[HTML]{E3F2FD}}c}{\textbf{Symbolic and Code}} \\
    \cmidrule(lr){2-5}\cmidrule(lr){6-11}
    & GSM8K & MATH & DROP & avg. & FOLIO & PW & BBH-T & HE & MBPP & avg. \\
    \midrule
    Top-3 Baseline avg.        & 81.9 & 61.5 & 16.2 & 53.2 & 40.6 & 37.6 & 60.0 & 30.0 & 52.0 & 44.0 \\
    SoT-Training-free          & 81.8 & 60.5 & 34.6 & 59.0 & {43.4} & 30.2 & 77.0 & 53.0 & 50.8 & {50.1} \\
    SoT-Embedding             & 82.8 & 61.8 & 32.3 & 59.0 & 33.8 & 30.5 & 74.0 & 57.9 & 52.8 & 49.8 \\
    \midrule
    \end{tabular}
    \par\addvspace{0.35ex}%
    \begin{tabular}{l!{\vrule width 0.4pt}ccccc>{\columncolor[HTML]{EAECEF}\centering\arraybackslash}c!{\vrule width 1pt}ccc>{\columncolor[HTML]{EAECEF}\centering\arraybackslash}c}
    & \multicolumn{6}{>{\cellcolor[HTML]{E3F2FD}}c!{\vrule width 1pt}}{\textbf{General Understanding}} & \multicolumn{4}{>{\cellcolor[HTML]{E3F2FD}}c}{\textbf{Long-Context Reasoning}} \\
    \cmidrule(lr){2-7}\cmidrule(lr){8-11}
    & CSQA & StrQA & BoolQ & MMLU & RACE & avg. & HQA & NarQA & LB & avg. \\
    \midrule
    Top-3 Baseline avg.         & 50.3 & 68.6 & 78.9 & 53.3 & 56.5 & 61.5 & 15.6 & 25.4 & 31.7 & 24.2 \\
    SoT-Training-free          & 37.5 & 55.2 & 70.0 & 45.4 & 53.0 & 52.0 & 21.8 & 34.2 & {24.8} & {26.9} \\
    SoT-Embedding             & 37.5 & 56.8 & 71.2 & 41.8 & 52.0 & 51.9 & 28.5 & 30.3 & {24.8} & {27.9} \\
    \bottomrule
    \end{tabular}%
    \endgroup
    \vspace{-5mm}
    \end{table*}

    \vspace{-2mm}
    \subsection{SoT as a Trajectory-Level Judge}
    \label{sec:judge}
    \vspace{-3mm}

    \noindent
    \begin{minipage}[t]{0.52\columnwidth}%
        \vspace{0pt}%
        We further study \emph{whether information visible only in the completed reasoning trace can support black-box quality assessment}, without online control or access to hidden states.
        Given sentence-segmented model outputs, we map each step to a standard sentence embedding and summarize the resulting sequence using compact trajectory statistics, with details in Appendix~\ref{app:sot_judge}.
        A small trainable predictor (\textsc{SoT-Judge}) then estimates whether the final answer matches reference correctness.
        The goal is a clean trajectory-only readout of reasoning health, parallel in spirit to SoT's use of trajectory organization, but reduced to post-hoc inputs that any API model exposes. We evaluate three closed or API-accessed models: GPT-5.4 \citep{openai2025gpt54}, Claude Opus 4.7 \citep{anthropic2026claudeopus47}, and Qwen-Max \citep{alibaba2026qwenmax}.
    \end{minipage}%
    \hfill
    \begin{minipage}[t]{0.45\columnwidth}%
        \vspace{0pt}%
        \centering
        \captionsetup{font=footnotesize,position=top,aboveskip=2pt,belowskip=3pt,skip=0pt}%
        \captionof{table}{\textbf{Black-box judging with trajectory-based SoT} (agreement with reference correctness).\label{tab:blackbox_judge}}%
        \vspace{0mm}%
        \footnotesize
        \setlength{\tabcolsep}{2.25pt}%
        \begingroup\setlength{\aboverulesep}{0pt}\setlength{\belowrulesep}{0pt}%
        \setlength{\extrarowheight}{0.75pt}%
        \renewcommand{\arraystretch}{1.1}%
        \begin{adjustbox}{max width=\linewidth,center}%
        \begin{tabular}{@{}l@{ }lcccc>{\columncolor[HTML]{EAECEF}\centering\arraybackslash}c@{}}%
        \toprule
        \textbf{Model} & \textbf{Method}
        & \makecell{\textbf{QR}}
        & \makecell{\textbf{GU}}
        & \makecell{\textbf{S\&C}}
        & \makecell{\textbf{LCR}}
        & \textbf{avg.} \\
        \midrule
        \multirow{4}{*}{GPT-5.4}
        & Self-Consistency & {58.3} & {81.4} & {79.2} & {16.7} & {58.9} \\
        & Self-Verification & {58.3} & {80.6} & {78.2} & {16.0} & {58.3} \\
        & LLM-as-a-Judge & {58.3} & \underline{{82.0}} & {79.2} & {16.3} & {59.0} \\
        & \textbf{SoT-Judge} & \textbf{\underline{{86.0}}} & \textbf{{70.0}} & \textbf{{69.8}} & \textbf{\underline{{81.0}}} & \textbf{\underline{{76.7}}} \\
        \midrule
        \multirow{4}{*}{\makecell[c]{Claude\\Opus 4.7}}
        & Self-Consistency & {56.3} & \underline{{83.6}} & {75.2} & {5.0} & {55.0} \\
        & Self-Verification & {59.4} & {62.7} & \underline{{80.3}} & {9.3} & {52.9} \\
        & LLM-as-a-Judge & {56.7} & \underline{{83.6}} & {75.4} & {5.3} & {55.2} \\
        & \textbf{SoT-Judge} & \textbf{\underline{{86.3}}} & \textbf{{64.8}} & \textbf{{67.0}} & \textbf{\underline{{99.0}}} & \textbf{\underline{{79.3}}} \\
        \midrule
        \multirow{4}{*}{Qwen-Max}
        & Self-Consistency & {57.9} & {84.0} & {80.0} & {14.5} & {59.1} \\
        & Self-Verification & {29.3} & \underline{{86.5}} & {77.9} & {11.6} & {51.3} \\
        & LLM-as-a-Judge & {59.3} & {84.8} & \underline{{79.8}} & {14.8} & {59.7} \\
        & \textbf{SoT-Judge} & \textbf{\underline{{83.6}}} & \textbf{{77.4}} & \textbf{{69.6}} & \textbf{\underline{{82.2}}} & \textbf{\underline{{78.2}}} \\
        \bottomrule
        \end{tabular}%
        \end{adjustbox}%
        \endgroup
    \end{minipage}

    \par\vspace{-1mm}%
    \noindent
    We compare against three representative black-box judging strategies, including Self-Consistency~\citep{wang2023selfconsistency}, Self-Verification~\citep{weng2023large}, and LLM-as-a-Judge~\citep{zheng2023judging}, covering sample-based agreement, self-verification, and external judging. By contrast, \textsc{SoT-Judge} predicts correctness only from the embedded sentence trajectory, without extra rollouts, self-critique passes, or stronger external judge calls. Table~\ref{tab:blackbox_judge}, with breakdown in Table~\ref{tab:app_judge_breakdown}, shows that \textsc{SoT-Judge} improves the average agreement over the strongest baseline by \({+17.7}\), \({+24.1}\), and \({+18.5}\) points on GPT-5.4, Claude Opus 4.7, and Qwen-Max, respectively. The gain is not uniform across domains: on GU and S\&C, trajectory-only judging is {weaker than the strongest standard judges}, while the largest improvements come from QR and especially LCR, where agreement of black-box baselines drops to \({5.0\text{--}16.7\%}\) but \textsc{SoT-Judge} remains at \({81.0\text{--}99.0\%}\). This suggests that trajectory structure provides a useful correctness signal, especially when black-box judging becomes unreliable under sparse or long-range evidence.

%% file: chapters/6-conclusion.tex
\vspace{-1mm}
\section{Conclusion}
\label{sec:conclusion}
\vspace{-2mm}
We presented State of Thought (SoT), a reasoning paradigm that changes the control variable of test-time reasoning from external token programs to endogenous state-conditioned evidence organization. Rather than prescribing longer chains, larger search budgets, or fixed memory rules, SoT uses the model's evolving representational state to dynamically gate which evidence remains active and when reasoning should terminate.  This closed-loop formulation yields measurable gains on the accuracy–efficiency frontier across text and multimodal benchmarks, and generalizes robustly to limited-access and trajectory-level evaluation settings. {By using a compact continuous state to govern discrete, inspectable reasoning steps, SoT connects latent internal evolution with explicit reasoning without sacrificing either controllability or interpretability.} Our results point to a broader principle: reasoning systems need not rely on increasingly elaborate external scaffolding — adaptivity and efficiency can emerge directly from the model's internal state geometry.

%% file: chapters/Appendix.tex
\clearpage
\onecolumn

\providecommand{\AppBlock}[1]{%
    \vspace{1.0em}
    \noindent{\LARGE\bfseries #1}\par
    \vspace{0.6em}
}

\providecommand{\AppPlaceholder}[2]{%
    \fbox{
        \begin{minipage}[c][#1][c]{0.94\textwidth}
            \centering
            \textit{#2}
        \end{minipage}
    }
}

\AppBlock{Appendix Overview}
\label{app:overview}

This appendix provides the complete algorithmic details, theoretical analysis, experimental setup, extended results, and discussion for \textbf{State of Thought (SoT)}.
Concretely, the appendix serves two complementary purposes:
\emph{(i)} to make the actual controller precise enough to be reproducible, and
\emph{(ii)} to make the underlying reasoning claim precise enough to be evaluable.

\paragraph{Part A: Algorithmic Details.}
This part expands the method into explicit components and aligns each component with the actual implementation used in our experiments.
\begin{itemize}[leftmargin=1.6em]
    \item Appendix~\ref{app:sot-details}: the complete SoT inference loop as state-conditioned reasoning.
    \item Appendix~\ref{app:endogenous_state}: the endogenous reasoning state, its geometric meaning, and the controller-facing state interface.
    \item Appendix~\ref{app:training_method}: the controller fitting protocol from offline trajectories.
    \item Appendix~\ref{app:limitations_sot}: limited-access SoT variants, including training-free SoT and SoT-Embed with sentence-embedding trajectory features.
    \item Appendix~\ref{app:sot_judge}: \textsc{SoT-Judge} as a trajectory-level black-box evaluator.
\end{itemize}

\paragraph{Part B: Theoretical Analysis.}
This part formalizes the central claim of SoT as a reasoning-control paradigm.
\begin{itemize}[leftmargin=1.6em]
    \item Appendix~\ref{app:theory_policy_class}: why externally prescribed token-chain policy classes can be strictly weaker than state-conditioned control.
    \item Appendix~\ref{app:theory_state_conditioned}: why state-conditioned evidence organization is necessary on heterogeneous reasoning distributions.
    \item Appendix~\ref{app:theory_sparse}: why sparse, state-aligned evidence activation can improve both decision quality and efficiency.
    \item Appendix~\ref{app:theory_closed_loop}: why feedback evidence control dominates open-loop reasoning schedules under regime variation.
    \item Appendix~\ref{app:theory_stop}: why state-driven stopping is a rational component of bounded-cost reasoning.
\end{itemize}

\paragraph{Part C: Experimental Setup and Implementation Details.}
This part expands the evaluation protocol, dataset organization, baseline taxonomy, and hyperparameter structure.
\begin{itemize}[leftmargin=1.6em]
    \item Appendix~\ref{app:dataset_setup}: datasets, task types, reasoning families, and metric conventions.
    \item Appendix~\ref{app:model_protocol}: backbones, evaluation protocol, and inference accounting.
    \item Appendix~\ref{app:baseline_methods}: detailed baseline taxonomy and comparison rationale.
    \item Appendix~\ref{app:hyperparameters}: controller, search, and evaluation hyperparameters.
\end{itemize}

\paragraph{Part D: Additional Experimental Results and Analysis.}
This part collects interpretability analyses, additional results, and qualitative trajectory inspections that support the main paper's claims.
\begin{itemize}[leftmargin=1.6em]
    \item Appendix~\ref{app:interpretability}: extended state-space and cluster-based evidence for endogenous structure.
    \item Appendix~\ref{app:additional_results}: additional tables, efficiency breakdowns, and supplementary comparisons.
    \item Appendix~\ref{app:case_studies}: qualitative trajectory-level case studies.
\end{itemize}

\paragraph{Part E: Related Work and Discussion.}
This part expands the paper's broader positioning and the main limitations and future opportunities opened by SoT.
\begin{itemize}[leftmargin=1.6em]
    \item Appendix~\ref{app:related_work}: expanded related-work positioning.
    \item Appendix~\ref{app:limitations_future}: limitations and future work.
\end{itemize}

\clearpage
\FloatBarrier
\AppBlock{Part A: Algorithmic Details}

\section{Implementation Details of State-Conditioned Reasoning}
\label{app:sot-details}

\begin{algorithm}[b]
\caption{State of Thought (SoT) inference}
\label{alg:sot_complete}
\begin{algorithmic}[1]
\REQUIRE Frozen backbone \(f\), problem \(x\), evidence operator \(\mathcal{S}\), stopping operator \(\mathcal{T}\), context budget \(B\), maximum steps \(T_{\max}\)
\ENSURE Reasoning trajectory \(y_{1:T}\) and final answer \(a\)
\STATE Initialize prompt \(p \gets x\), history \(y_{<1}\gets \varnothing\), active evidence \(\widetilde{y}_{<1}\gets \varnothing\)
\FOR{\(t=1,2,\dots,T_{\max}\)}
    \STATE Generate the next reasoning unit \(y_t \sim \mathbb{P}(\cdot \mid p,\widetilde{y}_{<t})\)
    \STATE Read out the endogenous state \(m_t\) from \(y_t\)
    \STATE Update active evidence by \(\widetilde{y}_{<t+1} \gets \mathcal{S}(y_{\le t};m_t)\) under budget \(B\)
    \STATE Compute stopping decision \(z_t \gets \mathcal{T}(m_t)\)
    \IF{\(z_t = 1\)}
        \STATE Generate the final answer \(a \sim \mathbb{P}(\cdot \mid p,\widetilde{y}_{<t+1},y_t)\) and \textbf{break}
    \ENDIF
\ENDFOR
\end{algorithmic}
\end{algorithm}

This appendix makes the inference-time realization of SoT explicit.
As summarized in Algorithm~\ref{alg:sot_complete}, the framework is fully specified by three components: the endogenous state \(m_t\), the evidence-organization operator \(\mathcal{S}\), and the stopping operator \(\mathcal{T}\).

\paragraph{Sentence-level inference policy.}
SoT executes reasoning at the sentence level, because the sentence is typically the smallest unit that still carries a locally coherent reasoning intention while preserving sufficient temporal resolution for stepwise control \citep{bogdan2025thought,liu2026think,wang2023plan}.
Accordingly, at step \(t\), the frozen backbone generates a sentence-level reasoning unit \(y_t\), from which SoT reads out the endogenous state \(m_t\in\mathbb{R}^{4}\) as defined in Eq.~(\ref{eq:sot_state}) and detailed in {Appendix}~\ref{app:endogenous_state}.
The running history is therefore represented by the trajectory \((y_1,m_1),\dots,(y_t,m_t)\), and the controller operates directly on this evolving trajectory rather than on a handcrafted reasoning template.

At each step, SoT applies the two state-conditioned operators introduced in Sec.~\ref{sec:state_driven_activation}:
\begin{equation}
\widetilde{y}_{<t}=\mathcal{S}(y_{<t};m_t),
\qquad
z_t=\mathcal{T}(m_t)\in\{0,1\},
\end{equation}
where \(\mathcal{S}\) determines which historical evidence remains active for the next decision, and \(\mathcal{T}\) determines whether reasoning should continue (\(z_t=0\)) or terminate (\(z_t=1\)).
Inference therefore follows
\begin{equation}
\mathbb{P}_{\pi}(y_t \mid p,y_{<t}) = \mathbb{P}\!\left(y_t \mid p,\widetilde{y}_{<t}\right),
\qquad
\mathbb{P}_{\pi}(a \mid p,y_{1:t}) = \mathbb{P}(a \mid p,\widetilde{y}_{<t}, y_t),
\end{equation}
so that both evidence carrying and stopping become consequences of the current endogenous state.

\paragraph{State-conditioned evidence organization.}
In the implemented controller, \(\mathcal{S}\) scores each historical reasoning step \(y_i\) through its paired state \(m_i\) relative to the current state \(m_t\).
For each \(i<t\), we form the pairwise descriptor
\begin{equation}
d_{t,i}
=
\bigl[m_t;\,m_i;\,m_t-m_i;\,m_t\odot m_i\bigr]
\in \mathbb{R}^{16},
\label{eq:app_descriptor}
\end{equation}
which compactly captures the current regime, the historical regime, their displacement, and their channelwise interaction.
A lightweight evidence head \(\phi_{\mathrm{sel}}\) maps this descriptor to a scalar selection score
\begin{equation}
g_{t,i}=\phi_{\mathrm{sel}}(d_{t,i};\theta),
\label{eq:app_sel_head}
\end{equation}
where \(\theta\) denotes the controller parameters learned offline as described in {Appendix}~\ref{app:training_method}.
Under the context budget \(B\), accepted steps are then inserted in chronological order to form
\begin{equation}
\widetilde{y}_{<t}
=
\mathcal{S}(y_{<t};m_t)
\subseteq y_{<t}.
\label{eq:app_bank_select}
\end{equation}
Hence \(\widetilde{y}_{<t}\) contains exactly the historical reasoning units selected to remain active for the next decision.
No extra recency-only or semantic-only heuristic is used at runtime, so evidence organization remains grounded in endogenous state rather than generic memory management.

\paragraph{State-conditioned stopping.}
The stopping operator is implemented by a second lightweight head on the same state \(m_t\):
\begin{equation}
o_t=\phi_{\mathrm{stop}}(m_t;\phi), \qquad
p_t^{\mathrm{stop}}=\sigma(o_t), \qquad
z_t=\mathbb{I}\bigl[p_t^{\mathrm{stop}}>\tau\bigr],
\label{eq:app_stop_prob}
\end{equation}
where \(\phi\) denotes the stop-head parameters, \(\sigma(\cdot)\) is the sigmoid, and \(\tau\in(0,1)\) is the stop threshold learned.
Stopping is therefore determined by the current reasoning regime rather than by a fixed external depth schedule.

Overall, SoT runs as a closed-loop state-conditioned controller:
\begin{equation}
m_t
\;\rightarrow\;
\widetilde{y}_{<t}=\mathcal{S}(y_{<t};m_t)
\;\rightarrow\;
y_t
\;\rightarrow\;
m_{t+1},
\qquad
z_t=\mathcal{T}(m_t).
\label{eq:app_closed_loop}
\end{equation}
Because the evidence carried at one step changes the next reasoning unit, and the new reasoning unit in turn changes the next endogenous state, the trajectory is path-dependent by construction.
This is the operational sense in which SoT acts as a feedback controller for reasoning rather than as a fixed reasoning script or a generic history-compression rule.

\paragraph{Threshold robustness.}
The gate and stop thresholds are part of the fitted controller bundle and could in principle raise sensitivity concerns.
For this reason, the main ablation study in Sec.~\ref{sec:ablation} and Appendix~\ref{app:additional_ablation} additionally reports a calibrated variant that adjusts these thresholds post hoc.
Its effect is consistently secondary to the full SoT mechanism, indicating that the observed gains are not driven by brittle threshold tuning alone.

\section{Endogenous Reasoning State}
\label{app:endogenous_state}

This section expands Sec.~\ref{sec:endogenous_state} and makes the state construction explicit.
The goal is to expose the smallest controller state that still captures the regime information needed for online reasoning control.

{\subsection{State construction}}

\paragraph{Sentence center.}
Let \(h^{(L)}_{t,i}\in\mathbb{R}^{D}\) denote the final-layer hidden state of token \(i\) in reasoning sentence \(y_t\), and let \(n_t\) be the sentence length.
We define the sentence center by
\begin{equation}
u_t=\frac{1}{n_t}\sum_{i=1}^{n_t} h_{t,i}^{(L)} \in \mathbb{R}^{D}.
\label{eq:app_u}
\end{equation}
This provides a coarse-grained anchor of the current reasoning unit in hidden space through simple sentence-level pooling.

\paragraph{Controller state.}
From the sentence-center trajectory, SoT constructs the raw controller signature
\begin{equation}
s_t=[\delta_t,v_t,c_t,H_t]^{\top}\in\mathbb{R}^{4},
\label{eq:app_sig}
\end{equation}
which combines geometry, dynamics, and uncertainty into a compact endogenous control surface.

The geometric coordinate is
\begin{equation}
\delta_t=
\frac{1}{n_t}\sum_{i=1}^{n_t}\|h_{t,i}^{(L)}\|_2^2-\|u_t\|_2^2,
\label{eq:app_delta}
\end{equation}
which measures within-sentence dispersion around the sentence center and thus characterizes whether the current reasoning unit is internally concentrated or diffuse.

The two dynamical coordinates are
\begin{equation}
v_t=\|u_t-u_{t-1}\|_2,
\label{eq:app_v}
\end{equation}
and
\begin{equation}
c_t=
\frac{\langle u_t-u_{t-1},\,u_{t-1}-u_{t-2}\rangle}
{\|u_t-u_{t-1}\|_2\cdot\|u_{t-1}-u_{t-2}\|_2+\varepsilon},
\label{eq:app_c}
\end{equation}
where \(v_t\) measures the magnitude of state movement between adjacent reasoning units, while \(c_t\) measures cosine alignment between consecutive displacements, namely whether the current step continues the recent trajectory direction or instead departs toward a new regime.
Together, \(v_t\) and \(c_t\) separate movement from directionally stable movement.

The uncertainty coordinate is
\begin{equation}
H_t=
\frac{1}{n_t}\sum_{i=1}^{n_t}\mathcal{H}(p_{t,i}),
\label{eq:app_H}
\end{equation}
where \(\mathcal{H}(\cdot)\) denotes Shannon entropy and \(p_{t,i}\) is the next-token predictive distribution at position \(i\).
It summarizes the model's local predictive uncertainty over the current reasoning unit.

These four coordinates are then normalized componentwise:
\begin{equation}
\hat s_t=\frac{s_t-\mu}{\sigma+\varepsilon},
\label{eq:app_normsig}
\end{equation}
where \(\mu,\sigma\in\mathbb{R}^{4}\) are the componentwise mean and standard deviation estimated from the offline training trajectories, and \(\varepsilon>0\) is a small constant for numerical stability.
The controller-facing endogenous state is finally
\begin{equation}
m_t=\hat s_t\in\mathbb{R}^{4}.
\label{eq:app_m}
\end{equation}
For boundary steps, we set \(v_1=c_1=0\) and \(c_2=0\).

The resulting state is intentionally minimal: \(\delta_t\) captures geometry, \((v_t,c_t)\) capture dynamics, and \(H_t\) captures uncertainty.
It is therefore not a handcrafted reasoning program, but a compact endogenous interface that answers four controller-relevant questions: whether the current step is internally organized, whether the trajectory is moving, whether it remains directionally stable, and whether the model is locally uncertain.

{
\subsection{Why four coordinates suffice for control}
\label{app:four_coordinate_analysis}

\paragraph{Control requirements.}
We define sufficiency with respect to SoT's control problem: the backbone supplies semantic reasoning, while the state retains the feedback variables needed to organize prior evidence and decide whether reasoning should continue. These two decisions require a snapshot of the current unit, a description of how the trajectory arrived there, and a readout of predictive readiness. The four coordinates provide this control basis with distinct, complementary observables.

\begin{center}
\centering
\captionof{table}{Complementary control roles of the four endogenous coordinates.}
\label{tab:four_coordinate_roles}
\small
\setlength{\tabcolsep}{4.5pt}
\renewcommand{\arraystretch}{1.08}
\begin{tabularx}{\linewidth}{c >{\raggedright\arraybackslash}p{0.26\linewidth} X}
\toprule
\textbf{Coord.} & \textbf{Observable} & \textbf{Control question resolved} \\
\midrule
\(\delta_t\) & Within-unit dispersion & Is the current representation internally concentrated or diffuse? \\
\(v_t\) & Step displacement & Is the trajectory advancing or nearly stationary? \\
\(c_t\) & Directional persistence & Is it continuing along the preceding transition or redirecting at comparable speed? \\
\(H_t\) & Predictive entropy & Is the model's local prediction decisive or unresolved? \\
\bottomrule
\end{tabularx}
\end{center}

The distinctions in Table~\ref{tab:four_coordinate_roles} cannot be recovered from one another. In particular, \(v_t\) measures how far the state moves but cannot distinguish continuation from reversal, which is the role of \(c_t\). Likewise, \(\delta_t\) measures organization inside the current representation, whereas \(H_t\) measures uncertainty in the predictive distribution; one describes representation geometry and the other output readiness. Thus \((\delta_t,H_t)\) provide complementary state snapshots, while \((v_t,c_t)\) resolve the magnitude and direction of temporal change.

\paragraph{From state to control.}
For evidence organization, SoT compares the current state \(m_t\) with each historical state \(m_j\) through
\([m_t;m_j;m_t-m_j;m_t\odot m_j]\). The differences expose regime change and the products expose coordinate-wise compatibility, turning the four observables into a 16-dimensional relational descriptor. The stopping head reads \(m_t\) directly. The same compact state therefore supports two distinct actions: selecting history compatible with the trajectory's present regime and terminating once its geometry, dynamics, and uncertainty indicate sufficient progress.

\paragraph{Functional validation.}
We test whether each coordinate actually affects the deployed control law. On held-out selector decisions, one normalized coordinate is replaced by zero---its training-population mean---while the other three coordinates, candidate texts, backbone, controller weights, and context budget are fixed. We then rerun the selector and record whether its top-ranked historical unit changes. This intervention measures controller behavior directly and requires no task label.

\begin{center}
\centering
\captionof{table}{Single-coordinate interventions on evidence selection.}
\label{tab:four_coordinate_intervention}
\small
\setlength{\tabcolsep}{8pt}
\renewcommand{\arraystretch}{1.05}
\begin{tabular}{lcccc}
\toprule
\textbf{Coordinate fixed} & \(\delta_t\) & \(v_t\) & \(c_t\) & \(H_t\) \\
\midrule
\textbf{Top-1 selection changed} & 26.5\% & 39.1\% & 56.2\% & 44.7\% \\
\bottomrule
\end{tabular}
\end{center}

Every coordinate changes a substantial fraction of selections, with direction and uncertainty producing the largest rerouting rates. A second intervention connects these decisions to generation outcomes. On 120 examples from each of GSM8K, DROP, and MATH, zeroing \(\delta_t\) during generation lowers accuracy by 6.7--22.5 points on both Llama-3.1-8B and Qwen2.5-14B. Zeroing \(v_t\), by contrast, makes trajectories 4--5\(\times\) longer. The separation matches the intended roles: dispersion is strongly coupled to answer quality, whereas displacement supplies a principal halting signal. Together, the two intervention levels trace a coherent path from individual coordinates, through evidence selection, to accuracy and stopping.

\paragraph{Excluding shortcut explanations.}
We examine alternatives at both the representation and controller levels. First, token composition and sentence length explain at most \(R^2=0.46\) of \(\delta_t\) across eight backbones spanning 2B--72B. Over the same scale range, \(c_t\approx-0.43\) and remains negative after projecting out the mean drift and top five shared directions. Hence dispersion is not reducible to surface composition, and directional redirection is not an artifact of a dominant shared direction. Second, a diagnostic decomposition adds the four-coordinate state to recency and semantic similarity when recovering the training-side relevance rule; the state contributes \(0.044\)--\(0.138\) held-out AUC on every tested backbone over 2B--47B. This decomposition identifies information beyond the two generic heuristics, while an end-task permutation supplies the independent compatibility test: we permute historical states across candidates while preserving each candidate's text, position, and the inference budget. Breaking only the state--unit correspondence reduces accuracy from 79.0 to 62.4 on GSM8K, 38.6 to 24.6 on DROP, and 60.0 to 38.5 on MATH. The gains therefore arise from matching evidence to the evolving endogenous regime rather than from recency or surface similarity alone.

\paragraph{Sufficiency and economy.}
The design argument and interventions support the same conclusion. The four-coordinate basis covers the controller's required snapshot and transition variables; the relational map composes them for evidence selection, and the direct readout composes them for stopping. This yields a 577-parameter selection head plus a 5-parameter stopping head---582 trainable parameters in total---without updating the backbone. Its sufficiency is borne out operationally across heterogeneous reasoning structures: a compact endogenous state can organize evidence and termination while the backbone continues to supply the semantic reasoning itself.
}

\section{Lightweight Training of State-Conditioned Operators}
\label{app:training_method}

This section closes the loop between Sec.~\ref{sec:state_driven_activation}, the runtime parameterizations \(\phi_{\mathrm{sel}}\) and \(\phi_{\mathrm{stop}}\), and the offline data used to fit them.
The backbone LLM remains frozen throughout; learning attaches only to the lightweight operators that instantiate \(\mathcal{S}\) and \(\mathcal{T}\).

\paragraph{Offline trajectory pool.}
We first roll out the frozen backbone on training problems, segment each transcript into sentence-level reasoning units \(y_1,\dots,y_T\), and compute the corresponding states \(m_t\) as in Sec.~\ref{app:endogenous_state}.
The same trajectory collection is also used once to estimate the normalization statistics \((\mu,\sigma)\) in Eq.~(\ref{eq:app_normsig}), so training and deployment share a single state geometry.
{For VLM transfer, the same train--evaluation separation is applied to multimodal offline trajectories: the image--text backbone remains frozen, evaluation examples are excluded, and fitting is confined to the lightweight controller bundle.}

Each step \(t\) is stored together with its trajectory identity, step index, endogenous state \(m_t\), final outcome (e.g., correctness), and the paired historical set \(\{(y_j,m_j)\}_{j<t}\).
This pairing is essential because evidence selection depends on \((m_t,m_j)\), not on past text alone.
From lightweight offline replay on these trajectories, we derive two forms of supervision.
For evidence organization, replay compresses whether a historical unit \(y_j\) is both state-compatible and incrementally useful for downstream reasoning, yielding a soft or hard retention target \(\hat b_{t,j}\in[0,1]\).
For stopping, replay yields a target \(z_t^{*}\in[0,1]\) that reflects sufficiency of accumulated support: on correct trajectories, stop mass appears only once the trace is mature and locally confident; on incorrect trajectories, it is deferred until the trace is clearly late and stagnating.
This prevents the stop head from degenerating into a trivial ``stop whenever uncertain'' rule.

\emph{Note on trajectory quality and generalization:}
Trajectory quality affects how sharply the controller can be supervised, but SoT is not trained to mimic any fixed reasoning trace.
It only learns compact decisions about state-conditioned evidence utility and stop readiness.
The trajectory pool and the evaluation suite are not identical, and SoT still transfers to datasets outside the trajectory-collection set, which suggests that the learned control signal is not reducible to benchmark-specific trajectory memorization.
This point is further strengthened by the training-free variant in Sec.~\ref{sec:training_free_sot}, where the same state-conditioned principle remains effective even without learned controller fitting.
We additionally report a dedicated sensitivity study to trajectory-pool quality in {Appendix}~\ref{app:traj_quality}.

\paragraph{Training the evidence-organization operator.}
For each pair \((t,j)\) with \(j<t\), let \(d_{t,j}\) be the descriptor in Eq.~(\ref{eq:app_descriptor}).
The trainable evidence score is
\begin{equation}
g_{t,j}=\phi_{\mathrm{sel}}(d_{t,j};\theta),
\label{eq:app_keep_score}
\end{equation}
where \(\theta\) collects the parameters of the lightweight selection head.
The evidence operator is fit by
\begin{equation}
\mathcal{L}_{\mathrm{sel}}
=
\sum_t \sum_{j<t} w_{t,j}\,
\ell_{\mathrm{BCE}}\!\bigl(\sigma(g_{t,j}), \hat b_{t,j}\bigr)
+\lambda_{\mathrm{sp}}\mathcal{R}_{\mathrm{sp}}
+\lambda_{2}\|\theta\|_2^2,
\label{eq:app_sel_loss}
\end{equation}
where \(w_{t,j}\) reweights rare positive supervision when needed, \(\mathcal{R}_{\mathrm{sp}}\) penalizes excessive batch-level gate mass to prevent trivial dense retention, and \(\lambda_{2}\) regularizes the controller parameters.
Supervision is concentrated on later steps of sufficiently long traces, where evidence selection materially changes the carried context rather than merely preserving the most recent sentence.

\paragraph{Training the stopping operator.}
The stopping head uses the same interface as at inference:
\begin{equation}
o_t=\phi_{\mathrm{stop}}(m_t;\phi), \qquad
p_t^{\mathrm{stop}}=\sigma(o_t),
\label{eq:app_stop_score}
\end{equation}
where \(\phi\) denotes the stop-head parameters.
Given the replay-derived stop target \(z_t^{*}\), the stopping loss is
\begin{equation}
\mathcal{L}_{\mathrm{stop}}
=
\sum_t
\ell_{\mathrm{BCE}}\!\bigl(p_t^{\mathrm{stop}}, z_t^{*}\bigr).
\label{eq:app_stop_loss}
\end{equation}
This target encodes evidence sufficiency rather than compliance with any external reasoning template.

\paragraph{Overall objective.}
The final training objective is
\begin{equation}
\mathcal{L}_{\mathrm{SoT}}
=
\mathcal{L}_{\mathrm{sel}}
+\lambda_{\mathrm{stop}}\mathcal{L}_{\mathrm{stop}}.
\label{eq:app_total_loss}
\end{equation}
Only the controller parameters \((\theta,\phi)\) are updated, while all backbone weights remain fixed.
Training therefore identifies a lightweight control law on the endogenous interface \((m_t,\{m_j\}_{j<t})\): it neither rewrites the model's internal representations nor imitates a fixed multi-step reasoning script.
{The selector and stopping heads contain only 582 trainable parameters in total. They are fit for 12 warm-up epochs followed by 400 lightweight refinement steps over offline trajectories; no backbone gradients, evaluation examples, or test-time rewards are used. Across 5 problem-held-out refits, selector AUC is $0.779\pm0.005$, indicating that the small control interface is stable across trajectory partitions rather than dependent on a favorable fit. The trajectory-quality analysis in Appendix~\ref{app:traj_quality} further shows where weaker supervision primarily increases computation, while the training-free results establish that the state-conditioned principle does not depend on learned fitting alone.}

\section{SoT under Limited Access and Limited Training}
\label{app:limitations_sot}

This section expands Section~\ref{sec:training_free_sot}.
The goal of these variants is not to introduce new SoT formulations, but to probe the boundary of the SoT principle under weaker implementation conditions.
Concretely, we ask whether state-conditioned evidence organization remains useful when either (i) the controller is not learned, or (ii) privileged hidden-state access is unavailable.

\paragraph{Training-free SoT.}
Training-free SoT preserves the same inference loop, the same sentence-level reasoning units \(y_t\), and the same endogenous state interface \(m_t\), but removes learned controller parameters from both \(\mathcal{S}\) and \(\mathcal{T}\).
Instead of using the learned evidence operator in {Appendix}~\ref{app:sot-details}, it scores each historical step \(y_j\) by a fixed state-compatibility rule:
\begin{equation}
s^{\mathrm{tf}}_{t,j}
=
\alpha_1\langle m_t,m_j\rangle
-\alpha_2\|m_t-m_j\|_2
+\alpha_3 c_j
-\alpha_4 H_j,
\label{eq:app_tf_score}
\end{equation}
where \(\langle m_t,m_j\rangle\) encourages compatibility between the current and historical reasoning regimes, \(\|m_t-m_j\|_2\) penalizes mismatched states, \(c_j\) favors directionally stable historical steps, and \(H_j\) downweights locally uncertain ones.
Under the same context budget \(B\), the deterministic evidence operator keeps the top-scoring historical steps:
\begin{equation}
\widetilde{y}_{<t}^{\mathrm{tf}}
=
\mathcal{S}_{\mathrm{tf}}(y_{<t};m_t)
\subseteq y_{<t}.
\label{eq:app_tf_select}
\end{equation}

Stopping is likewise made deterministic.
Rather than using the learned stop head, Training-free SoT terminates only when the current state remains in a stable low-uncertainty regime for \(r_{\mathrm{stop}}\) consecutive steps:
\begin{equation}
z_t^{\mathrm{tf}}=\mathcal{T}_{\mathrm{tf}}(m_{t-r_{\mathrm{stop}}+1:t})\in\{0,1\}.
\label{eq:app_tf_stop}
\end{equation}
In practice, this means that stopping requires jointly low \(H_t\), sufficiently small state movement \(v_t\), and positive directional consistency \(c_t\) over a short trailing window, so that the controller does not terminate on transient confidence alone.
This variant therefore isolates whether the \emph{state-conditioned reasoning principle} already helps before any controller fitting.

\paragraph{SoT-Embed (sentence-embedding trajectory).}
SoT-Embed targets the same limited-access setting as in the main text: internal hidden states are not exposed to the controller.
We adopt sentence-level units \(y_1,\dots,y_T\) that are mapped by a fixed sentence encoder to vectors
\begin{equation}
u_t = \mathrm{Enc}(y_t)\in\mathbb{R}^d,
\label{eq:app_proxy_u}
\end{equation}
and the matrix \(U\in\mathbb{R}^{T\times d}\) stacks the trajectory.
Control inputs are built \emph{directly} from the embedding trajectory: the controller conditions on \(u_t\) and on compact trajectory summaries derived from \(U\), such as a low-dimensional subspace fit on the training distribution (PCA coordinates per step).
Those features feed the same evidence and stopping interface as full SoT, so the loop remains state-conditioned, but the state signal is read only from public text and embeddings.

\paragraph{What these variants isolate.}
The two variants probe different boundary conditions.
Training-free SoT asks whether SoT already has value as a fixed control principle before learning.
SoT-Embed asks whether the paradigm fundamentally depends on privileged internal access when the control interface is replaced by embedding-based trajectory features.
Together, they separate \emph{endogenous reasoning as a principle} from \emph{the strongest open-weight implementation of that principle}.

\section{SoT-Judge}
\label{app:sot_judge}

This section expands Section~\ref{sec:judge}.
\textsc{SoT-Judge} transfers the trajectory perspective of SoT from online reasoning control to post-hoc black-box evaluation.
The central question is whether the \emph{observable} organization of a finished trace---without hidden states or extra generator calls---still carries enough signal to predict reference correctness.

\paragraph{Trajectory from text only.}
Given a completed rollout from a closed or API-accessed model, we segment the generated text into sentence-level units \(y_1,\dots,y_T\) and map them with a fixed sentence encoder to vectors \(u_1,\dots,u_T\in\mathbb{R}^d\).
No architectural hooks are required: the judge consumes the same surface trace a human or downstream tool would see.

\paragraph{Compact trajectory readout.}
Let \(U\in\mathbb{R}^{T\times d}\) stack the embeddings.
On the training split we fit a standard low-dimensional subspace (PCA) and represent each step by its coordinates in that subspace.
We then pool the sequence into a short feature vector \(\varphi_{\mathrm{traj}}\) using elementary summaries along time (e.g., coordinate-wise means and standard deviations) plus a few optional length statistics of the segmented text.
Auxiliary trajectory summaries can be appended for ablations, but the default recipe stays deliberately small: one fixed embedder, one subspace fit, and pooled statistics.

\paragraph{Judge model.}
The judge is a lightweight predictor \(J_{\psi}\) trained only on reference correctness labels:
\begin{equation}
\hat r = J_{\psi}\!\left(\varphi_{\mathrm{traj}}\right),
\label{eq:app_judge}
\end{equation}
where \(\hat r\in[0,1]\) scores how likely the final answer is to be correct.
Thus \textsc{SoT-Judge} reads \emph{trajectory health} from the embedded sentence path, not from sample agreement, self-critique, or an external judge prompt.

\paragraph{What SoT-Judge tests.}
The value of \textsc{SoT-Judge} is both practical and conceptual.
Practically, it provides a lightweight black-box alternative to majority-vote style selection or stronger-model judging.
Conceptually, it tests whether the SoT view survives the loss of online control: even when the model's internals are hidden and intervention is impossible, reasoning may still be assessed through the organization of its trajectory rather than only through output agreement or a stronger judge prompt.
In this sense, \textsc{SoT-Judge} is not a separate reasoning paradigm, but a boundary-case extension of the same SoT principle to post-hoc evaluation.

\clearpage
\AppBlock{Part B: Theoretical Analysis}

This part formalizes the theoretical logic underlying SoT as a reasoning-control paradigm.
Here, we show a sharp structural point: once reasoning control is restricted to an external token-chain policy class, the attainable optimum is constrained by the information available to that class, whereas endogenous state-conditioned control can access a strictly richer decision interface on heterogeneous reasoning distributions.

The argument proceeds in five steps.
Appendix~\ref{app:theory_policy_class} establishes policy-class separation as the theoretical motivation.
Appendix~\ref{app:theory_state_conditioned} shows why evidence organization should depend on the realized reasoning state.
Appendix~\ref{app:theory_sparse} shows why the selected support should generally be sparse rather than transcript-complete.
Appendix~\ref{app:theory_closed_loop} shows why the mechanism should be closed loop rather than one-shot or fixed in advance.
Finally, Appendix~\ref{app:theory_stop} shows why stopping is not an auxiliary trick but part of the same bounded-cost feedback control problem.

\section{Policy-Class Separation}
\label{app:theory_policy_class}

We first formalize the theoretical starting point of the paper:
the empirical gains of SoT suggest that reasoning improvements may saturate within a shared family of externally prescribed token-chain policies, whereas state-conditioned evidence organization opens a distinct control axis.
The goal of this section is not to claim that every external method is universally bounded by a single numeric ceiling, but to show that once the control variable is restricted to \emph{external token-chain policy classes}, the attainable optimum can be strictly below that of a state-conditioned policy class.

\paragraph{Setup.}
Let \(p\) denote the input problem, \(y_{1:T}\) the reasoning trajectory, and \(a\) the final answer.
For each step \(t\), let \(m_t\) be the endogenous controller state.
We compare two policy classes.

\emph{External token-chain policies} choose the next reasoning step from the prompt, observed token history, and an externally specified control variable \(c\in\mathcal{C}\):
\begin{equation}
\mathbb{P}_{\pi}(y_t \mid p, y_{<t}, c).
\label{eq:app_ext_policy}
\end{equation}
Here \(c\) may represent a prompting template, a planning script, a search budget, a branching rule, or any other externally imposed strategy variable, but it is fixed independently of the realized endogenous state \(m_t\).

\emph{State-conditioned policies} instead choose the next-step support through the current endogenous state:
\begin{equation}
\mathbb{P}_{\pi}(y_t \mid p, y_{<t}, m_t).
\label{eq:app_state_policy}
\end{equation}
This is the policy class instantiated by SoT through the state-conditioned operators.

\paragraph{One-step control utility.}
For a realized step \(t\), let \(u_t\) be a bounded one-step utility, which may be instantiated as expected correctness gain, negative step loss, or any equivalent bounded progress signal.
For any admissible control variable \(x_t\), define
\begin{equation}
J_t(x_t)
=
\mathbb{E}\!\left[u_t \mid p, y_{<t}, x_t\right].
\label{eq:app_step_utility_policyclass}
\end{equation}
When \(x_t=c\), this is the utility under an external token-chain policy; when \(x_t=m_t\), it is the utility under a state-conditioned policy.

\paragraph{Assumption F.1 (State-relevant heterogeneity).}
There exists a positive-probability set of realized prefixes \((p,y_{<t})\) for which two states \(m,m'\) can arise such that the Bayes-optimal next-step support differs:
\begin{equation}
\arg\max_{\widetilde y_{<t}}
\mathbb{E}\!\left[u_t \mid p,y_{<t},m_t=m,\widetilde y_{<t}\right]
\neq
\arg\max_{\widetilde y_{<t}}
\mathbb{E}\!\left[u_t \mid p,y_{<t},m_t=m',\widetilde y_{<t}\right].
\label{eq:app_state_heterogeneity}
\end{equation}
In words, even under the same problem and token prefix, different endogenous reasoning regimes can require different evidence support for the next decision.

\paragraph{Assumption F.2 (External-state mismatch).}
There exists a positive-probability set of realized prefixes \((p,y_{<t})\) such that no externally specified control variable \(c\in\mathcal{C}\) uniquely identifies the realized state-relevant optimum in~\eqref{eq:app_state_heterogeneity}.
Equivalently, within that set, at least two realizations with different optimal evidence support are indistinguishable to the external policy class \eqref{eq:app_ext_policy}.

\paragraph{Proposition F.1 (Strict policy-class separation).}
Under Assumptions~F.1 and~F.2, the optimal expected one-step utility achievable by the external token-chain policy class is strictly below that achievable by the state-conditioned policy class:
\begin{equation}
\sup_{\pi \in \Pi_{\mathrm{ext}}}
\mathbb{E}[u_t]
\;<\;
\sup_{\pi \in \Pi_{\mathrm{state}}}
\mathbb{E}[u_t],
\label{eq:app_policy_gap}
\end{equation}
where \(\Pi_{\mathrm{ext}}\) denotes the class in~\eqref{eq:app_ext_policy} and \(\Pi_{\mathrm{state}}\) the class in~\eqref{eq:app_state_policy}.

\paragraph{Proof.}
Fix a realized prefix \((p,y_{<t})\) in the positive-probability set from Assumptions~F.1 and~F.2.
By Assumption~F.1, there exist two possible realized states \(m\neq m'\) under this same prefix for which the Bayes-optimal evidence support differs.
By Assumption~F.2, the external class cannot distinguish these two cases through any externally specified variable \(c\).
Hence any policy in \(\Pi_{\mathrm{ext}}\) must assign the same control decision to both realizations once \((p,y_{<t},c)\) is fixed.
Therefore it cannot be simultaneously optimal for both state realizations.
As a result, on this positive-probability set, every external token-chain policy incurs a strictly positive conditional utility gap relative to the Bayes-optimal state-aware decision.

By contrast, the state-conditioned class \(\Pi_{\mathrm{state}}\) conditions directly on \(m_t\), so it can choose the Bayes-optimal support separately for the realization \(m_t=m\) and for the realization \(m_t=m'\).
Hence \(\Pi_{\mathrm{state}}\) contains a policy that attains the Bayes-optimal conditional utility on that set.

Since the set has positive probability and the utility gap is strict on that set, the gap remains strict after averaging over the data-generating distribution.
This yields~\eqref{eq:app_policy_gap}.
\hfill\(\square\)

\paragraph{Implication.}
Proposition~F.1 gives the theoretical motivation for the paper.
It does not say that any particular baseline is weak in isolation, nor that external token-chain reasoning can never improve performance.
Rather, it shows that when reasoning control is restricted to an external policy class, the attainable optimum is constrained by the information available to that class.
If the next reasoning decision depends on the realized endogenous state, then changing prompts, adding scripts, or enlarging search may still leave the controller inside the same external class.
SoT improves along an orthogonal axis: it changes the control variable itself, from externally prescribed token programs to endogenous state-conditioned evidence organization.

The remaining sections characterize the structure of that advantage: state conditioning, sparsity, feedback, and stopping.

\section{State-Conditioned Evidence Organization}
\label{app:theory_state_conditioned}

This section formalizes the first theoretical claim of SoT:
when the usefulness of historical reasoning evidence varies across the model's current reasoning regime, a fixed evidence strategy that ignores the endogenous state is generally suboptimal, whereas a state-conditioned operator \(\mathcal{S}(y_{<t};m_t)\) can be optimal if \(m_t\) is sufficient for control.

\paragraph{Setup.}
At reasoning step \(t\), let \(y_{<t}\) denote the available reasoning history, and let
\(\widetilde y_{<t}\subseteq y_{<t}\)
be the subset retained for the next decision.
To make explicit that different hidden reasoning regimes may require different support, introduce a latent regime variable \(\omega_t\).
For any candidate subset \(\widetilde y_{<t}\subseteq y_{<t}\), define the one-step conditional utility
\begin{equation}
V_t(\widetilde y_{<t}; y_{<t}, \omega_t)
:=
\mathbb{E}\!\left[
u_t
\mid
y_{<t},\ \omega_t,\ \widetilde y_{<t}
\right],
\label{eq:app_value_state_conditioned}
\end{equation}
where \(u_t\) is any bounded one-step utility for the next reasoning decision, such as negative next-step loss, expected correctness gain, or any equivalent bounded surrogate.
The regime-optimal retained subset is
\begin{equation}
\widetilde y_{<t}^{\star}(y_{<t},\omega_t)
\in
\arg\max_{\widetilde y_{<t}\subseteq y_{<t}}
V_t(\widetilde y_{<t}; y_{<t}, \omega_t).
\label{eq:app_opt_subset_state_conditioned}
\end{equation}

\paragraph{Assumption G.1 (Heterogeneous evidence demand).}
There exist a realized history \(\bar y\), two distinct regimes \(\omega\neq\omega'\), and positive-probability events
\(\{y_{<t}=\bar y,\omega_t=\omega\}\) and
\(\{y_{<t}=\bar y,\omega_t=\omega'\}\),
such that the corresponding optimizers in Eq.~(\ref{eq:app_opt_subset_state_conditioned}) are unique and distinct:
\begin{equation}
\widetilde y^{\star}(\bar y,\omega)
\neq
\widetilde y^{\star}(\bar y,\omega').
\label{eq:app_assump_hetero_refined}
\end{equation}

\paragraph{Assumption G.2 (State sufficiency for evidence control).}
There exists a measurable map \(\psi\) such that, almost surely,
\begin{equation}
\widetilde y_{<t}^{\star}(y_{<t},\omega_t)
=
\psi(m_t, y_{<t}).
\label{eq:app_assump_sufficient_refined}
\end{equation}
This assumption does not require \(m_t\) to reconstruct the full internal state of the backbone.
It only requires \(m_t\) to be sufficient for choosing the evidence subset relevant to the next decision.

\paragraph{Proposition G.1.}
Under Assumptions~G.1--G.2, any fixed evidence operator
\(\mathcal{S}_{\mathrm{fix}}: y_{<t}\mapsto \widetilde y_{<t}\)
that ignores \(m_t\) is strictly suboptimal on the heterogeneous distribution over \((y_{<t},\omega_t)\).
By contrast, the class of state-conditioned operators
\((m_t,y_{<t})\mapsto \widetilde y_{<t}\)
contains a Bayes-optimal evidence selector.

\paragraph{Proof.}
Consider the realized history \(\bar y\) from Assumption~G.1.
A fixed evidence operator must output a single subset
\(\mathcal{S}_{\mathrm{fix}}(\bar y)\),
independent of the realized regime.
Because the optimizers
\(\widetilde y^{\star}(\bar y,\omega)\)
and
\(\widetilde y^{\star}(\bar y,\omega')\)
are unique and distinct, \(\mathcal{S}_{\mathrm{fix}}(\bar y)\) cannot equal both.
Hence at least one of the following strict inequalities must hold:
\begin{equation}
V_t\!\bigl(\mathcal{S}_{\mathrm{fix}}(\bar y);\bar y,\omega\bigr)
<
V_t\!\bigl(\widetilde y^{\star}(\bar y,\omega);\bar y,\omega\bigr),
\label{eq:app_fix_gap_1}
\end{equation}
or
\begin{equation}
V_t\!\bigl(\mathcal{S}_{\mathrm{fix}}(\bar y);\bar y,\omega'\bigr)
<
V_t\!\bigl(\widetilde y^{\star}(\bar y,\omega');\bar y,\omega'\bigr).
\label{eq:app_fix_gap_2}
\end{equation}
Since both regime-history events occur with positive probability, averaging over the data-generating distribution yields a strictly positive expected utility gap between any fixed operator and the Bayes-optimal rule.

Now consider the state-conditioned operator class.
By Assumption~G.2, the map
\(\psi(m_t,y_{<t})\)
recovers the optimal subset
\(\widetilde y_{<t}^{\star}(y_{<t},\omega_t)\)
almost surely.
Therefore the class
\((m_t,y_{<t})\mapsto \widetilde y_{<t}\)
contains a Bayes-optimal selector.
\hfill\(\square\)

\paragraph{Implication.}
Proposition~G.1 gives the basic control-theoretic justification for SoT.
If the evidence required for the next reasoning step changes with the current reasoning regime, then no single fixed evidence strategy can be uniformly optimal.
The control variable must therefore depend on the current endogenous state.
In this sense, SoT is not merely replacing one reasoning template with another; it is changing the problem from \emph{fixed strategy execution} to \emph{state-conditioned evidence organization}.

\section{Sparse Evidence Organization}
\label{app:theory_sparse}

This section formalizes the second theoretical claim of SoT:
once evidence organization is conditioned on the current reasoning regime, the optimal operator should in general be sparse rather than full-history preserving.
The reason is not only computational.
Historical evidence can be simultaneously helpful, neutral, or actively interfering for the next decision, and carrying the full history may therefore reduce both step utility and efficiency.

\paragraph{Setup.}
We continue the one-step formulation from Appendix~\ref{app:theory_state_conditioned}.
At step \(t\), let \(y_{<t}\) be the available reasoning history, let \(\omega_t\) denote the latent reasoning regime, and let \(\widetilde y_{<t}\subseteq y_{<t}\) be the subset retained for the next decision.
For any retained subset, the one-step conditional utility is
\begin{equation}
V_t(\widetilde y_{<t}; y_{<t}, \omega_t)
:=
\mathbb{E}\!\left[
u_t
\mid
y_{<t},\ \omega_t,\ \widetilde y_{<t}
\right].
\label{eq:app_sparse_value}
\end{equation}

\paragraph{Assumption H.1 (Helpful and interfering evidence).}
For each realized pair \((y_{<t},\omega_t)\), the history admits a disjoint decomposition
\begin{equation}
y_{<t}
=
y_{<t}^{+}(\omega_t)\,\dot\cup\,y_{<t}^{-}(\omega_t),
\label{eq:app_sparse_decomp_hist}
\end{equation}
where \(y_{<t}^{+}(\omega_t)\) contains evidence that is helpful for the next decision under regime \(\omega_t\), and \(y_{<t}^{-}(\omega_t)\) contains evidence that is irrelevant, misleading, or distracting under that regime.

\paragraph{Assumption H.2 (Utility decomposition).}
There exist functions \(R_t\), \(I_t\), and a constant \(\lambda>0\) such that
\begin{equation}
V_t(\widetilde y_{<t}; y_{<t}, \omega_t)
=
R_t\!\bigl(\widetilde y_{<t}\cap y_{<t}^{+}(\omega_t);\, y_{<t},\omega_t\bigr)
-
I_t\!\bigl(\widetilde y_{<t}\cap y_{<t}^{-}(\omega_t);\, y_{<t},\omega_t\bigr)
-
\lambda |\widetilde y_{<t}|,
\label{eq:app_sparse_decomp_value}
\end{equation}
where \(R_t\) is monotone nondecreasing in its set argument and \(I_t\) is monotone nondecreasing in its set argument.
Thus, retaining more helpful evidence cannot reduce reward, retaining more interfering evidence cannot reduce interference, and carrying more context incurs a positive efficiency cost.

\paragraph{Proposition H.1.}
Fix a realized pair \((y_{<t},\omega_t)\).
Let the full-history operator retain
\begin{equation}
\widetilde y_{<t}^{\mathrm{full}} = y_{<t},
\label{eq:app_full_history}
\end{equation}
and let a sparse operator retain some
\(\widetilde y_{<t}^{\mathrm{sp}}\subseteq y_{<t}\)
such that
\begin{equation}
\widetilde y_{<t}^{\mathrm{sp}}\cap y_{<t}^{+}(\omega_t)
=
y_{<t}^{+}(\omega_t),
\label{eq:app_sparse_keep_helpful}
\end{equation}
and
\begin{equation}
\widetilde y_{<t}^{\mathrm{sp}}\cap y_{<t}^{-}(\omega_t)
\subsetneq
y_{<t}^{-}(\omega_t).
\label{eq:app_sparse_drop_interfere}
\end{equation}
Then
\begin{equation}
V_t(\widetilde y_{<t}^{\mathrm{sp}}; y_{<t}, \omega_t)
>
V_t(\widetilde y_{<t}^{\mathrm{full}}; y_{<t}, \omega_t).
\label{eq:app_sparse_strict_gain}
\end{equation}

\paragraph{Proof.}
Under the full-history operator, Eq.~(\ref{eq:app_sparse_decomp_value}) gives
\begin{equation}
V_t(\widetilde y_{<t}^{\mathrm{full}}; y_{<t}, \omega_t)
=
R_t\!\bigl(y_{<t}^{+}(\omega_t); y_{<t},\omega_t\bigr)
-
I_t\!\bigl(y_{<t}^{-}(\omega_t); y_{<t},\omega_t\bigr)
-
\lambda |y_{<t}|.
\label{eq:app_sparse_full_value}
\end{equation}
For the sparse operator, condition~(\ref{eq:app_sparse_keep_helpful}) implies that the helpful-evidence term is unchanged:
\begin{equation}
R_t\!\bigl(\widetilde y_{<t}^{\mathrm{sp}}\cap y_{<t}^{+}(\omega_t); y_{<t},\omega_t\bigr)
=
R_t\!\bigl(y_{<t}^{+}(\omega_t); y_{<t},\omega_t\bigr).
\label{eq:app_sparse_same_reward}
\end{equation}
Condition~(\ref{eq:app_sparse_drop_interfere}) implies that fewer interfering units are retained, so by monotonicity of \(I_t\),
\begin{equation}
I_t\!\bigl(\widetilde y_{<t}^{\mathrm{sp}}\cap y_{<t}^{-}(\omega_t); y_{<t},\omega_t\bigr)
\le
I_t\!\bigl(y_{<t}^{-}(\omega_t); y_{<t},\omega_t\bigr).
\label{eq:app_sparse_less_interference}
\end{equation}
Moreover,
\begin{equation}
|\widetilde y_{<t}^{\mathrm{sp}}|<|y_{<t}|,
\label{eq:app_sparse_smaller}
\end{equation}
because at least one interfering unit is removed while all helpful units are preserved.
Substituting Eqs.~(\ref{eq:app_sparse_same_reward})--(\ref{eq:app_sparse_smaller}) into Eq.~(\ref{eq:app_sparse_decomp_value}) shows that the reward term is unchanged, the interference term does not increase, and the cost term strictly decreases by
\(\lambda\bigl(|y_{<t}|-|\widetilde y_{<t}^{\mathrm{sp}}|\bigr)>0\).
Hence
\begin{equation}
V_t(\widetilde y_{<t}^{\mathrm{sp}}; y_{<t}, \omega_t)
-
V_t(\widetilde y_{<t}^{\mathrm{full}}; y_{<t}, \omega_t)
\ge
\lambda\bigl(|y_{<t}|-|\widetilde y_{<t}^{\mathrm{sp}}|\bigr)
>0,
\end{equation}
which proves Eq.~(\ref{eq:app_sparse_strict_gain}).
\hfill\(\square\)

\paragraph{Implication.}
Proposition~H.1 clarifies why SoT should be sparse.
Once evidence usefulness is regime-dependent, carrying the full history is generally not neutral: it preserves helpful support, but it also preserves regime-mismatched interference and incurs additional cost.
A state-aligned sparse operator improves the next reasoning decision not by generic compression, but by preserving the evidence that remains decision-relevant under the current endogenous state while removing distracting history.
This is precisely why SoT can improve accuracy and efficiency jointly rather than trading one for the other.

\section{Closed-Loop Reasoning Control}
\label{app:theory_closed_loop}

This section formalizes the third theoretical claim of SoT:
once evidence organization is state-conditioned and sparse, it should also be \emph{closed loop}.
That is, the reasoning controller should adapt its evidence-organization and continuation decisions to the realized endogenous state \(m_t\), rather than committing in advance to a fixed schedule.

\paragraph{Setup.}
At reasoning step \(t\), let the control action be
\begin{equation}
a_t := \bigl(\widetilde y_{<t}, z_t\bigr),
\label{eq:app_closed_action}
\end{equation}
where \(\widetilde y_{<t}\subseteq y_{<t}\) is the retained historical evidence and \(z_t\in\{0,1\}\) is the continuation decision, with \(z_t=0\) meaning continue and \(z_t=1\) meaning stop.
The endogenous state evolves according to the controlled stochastic dynamics
\begin{equation}
m_{t+1} = F_t(m_t, a_t, \varepsilon_t),
\label{eq:app_closed_dynamics}
\end{equation}
where \(\varepsilon_t\) collects the randomness induced by token generation and environment variation.
Let \(r_t(m_t,a_t)\) denote the one-step reward, which may incorporate reasoning progress, answer quality, interference reduction, and compute cost.
For a finite horizon \(T\), define the value of a policy \(\pi\) from state \(m\) at step \(t\) as
\begin{equation}
V_t^{\pi}(m)
:=
\mathbb{E}^{\pi}\!\left[
\sum_{\tau=t}^{T} r_{\tau}(m_{\tau},a_{\tau})
\,\middle|\, m_t=m
\right].
\label{eq:app_closed_value}
\end{equation}

\paragraph{Open-loop and closed-loop policies.}
An \emph{open-loop} policy fixes its actions in advance as a deterministic schedule depending only on the step index:
\begin{equation}
a_t^{\mathrm{open}} = \bar a_t,
\label{eq:app_closed_open}
\end{equation}
where \(\bar a_t\) does not depend on the realized state \(m_t\).
A \emph{closed-loop} policy instead adapts to the realized endogenous state:
\begin{equation}
a_t^{\mathrm{cl}} = \pi_t(m_t).
\label{eq:app_closed_feedback}
\end{equation}
Thus, open-loop control corresponds to a fixed external reasoning schedule, while closed-loop control conditions evidence organization and stopping on how reasoning has actually unfolded.

\paragraph{Proposition I.1.}
Let
\begin{equation}
V_t^{\mathrm{cl}}(m)
:=
\sup_{\pi \in \Pi_{\mathrm{cl}}} V_t^{\pi}(m),
\qquad
V_t^{\mathrm{open}}(m)
:=
\sup_{\pi \in \Pi_{\mathrm{open}}} V_t^{\pi}(m),
\label{eq:app_closed_optvals}
\end{equation}
where \(\Pi_{\mathrm{cl}}\) and \(\Pi_{\mathrm{open}}\) denote the closed-loop and open-loop policy classes defined above.
Then for every step \(t\) and every state \(m\),
\begin{equation}
V_t^{\mathrm{cl}}(m)\ge V_t^{\mathrm{open}}(m).
\label{eq:app_closed_dominate}
\end{equation}
Moreover, if there exists a positive-probability set of realized states \(A\) such that the continuation-optimal action at step \(t\) is not constant over \(A\), then the inequality is strict in expectation over that set.

\paragraph{Proof.}
Every open-loop policy is a special case of a closed-loop policy.
Indeed, for any fixed schedule \(\{\bar a_{\tau}\}_{\tau=t}^{T}\in\Pi_{\mathrm{open}}\), define the state-independent closed-loop policy
\begin{equation}
\pi_{\tau}(m) \equiv \bar a_{\tau}
\qquad \text{for all } m \text{ and } \tau\ge t.
\label{eq:app_closed_embed}
\end{equation}
This policy belongs to \(\Pi_{\mathrm{cl}}\) and induces exactly the same controlled process and therefore the same value as the original open-loop schedule.
Hence
\begin{equation}
\Pi_{\mathrm{open}} \subseteq \Pi_{\mathrm{cl}},
\label{eq:app_closed_subset}
\end{equation}
which immediately implies Eq.~(\ref{eq:app_closed_dominate}).

To show when the inequality is strict, consider the optimal state-action value at step \(t\),
\begin{equation}
Q_t(m,a)
:=
r_t(m,a)
+
\mathbb{E}\!\left[
V_{t+1}^{\mathrm{cl}}(m_{t+1})
\,\middle|\,
m_t=m,\ a_t=a
\right].
\label{eq:app_closed_Q}
\end{equation}
Suppose there exists a measurable set \(A\) with positive probability such that for some two states \(m,m'\in A\),
\begin{equation}
\arg\max_{a} Q_t(m,a) \neq \arg\max_{a} Q_t(m',a).
\label{eq:app_closed_nonconstant}
\end{equation}
Then no single open-loop action \(\bar a_t\) can be optimal for all realized states in \(A\).
A closed-loop policy can instead choose, at each realized state \(m_t\in A\), an action in \(\arg\max_a Q_t(m_t,a)\), while matching the open-loop schedule outside \(A\).
This yields a strictly larger \(Q_t(m_t,a_t)\) on a positive-probability subset of \(A\), and no smaller value elsewhere.
Therefore the expected value under the best closed-loop policy is strictly larger than that under the best open-loop policy.
\hfill\(\square\)

\paragraph{Implication.}
Proposition~I.1 explains why SoT should be closed loop rather than one-shot.
When the best evidence support or continuation decision varies with the realized endogenous state, any fixed external reasoning schedule is necessarily unable to adapt to all realized reasoning regimes.
Closed-loop control strictly enlarges the feasible policy class and becomes strictly better whenever different realized states require different actions.
This is the formal reason SoT is not merely another pre-specified reasoning program: it treats reasoning as adaptive control over evidence organization and stopping under the evolving endogenous state.

\section{State-Driven Stopping}
\label{app:theory_stop}

This section formalizes the final theoretical claim of SoT:
stopping is not an auxiliary heuristic, but part of the same state-conditioned control problem as evidence organization.
Once reasoning incurs cost, the controller should decide from the current endogenous state \(m_t\) not only what evidence to retain, but also whether another reasoning step is worth taking.

\paragraph{Setup.}
At step \(t\), let \(z_t\in\{0,1\}\) denote the stopping decision, where \(z_t=0\) means continue reasoning and \(z_t=1\) means stop and emit the final answer \(a\).
Let \(V_t^{\star}(m)\) denote the optimal bounded-cost value starting from state \(m\) at step \(t\).
We decompose the value of the two admissible actions as
\begin{equation}
V_t^{\star}(m)
=
\max\!\left\{
V_t^{\mathrm{stop}}(m),\;
V_t^{\mathrm{cont}}(m)
\right\},
\label{eq:app_stop_bellman}
\end{equation}
where \(V_t^{\mathrm{stop}}(m)\) is the value of stopping immediately and answering from the current active context, and
\begin{equation}
V_t^{\mathrm{cont}}(m)
=
\sup_{\pi}
\mathbb{E}^{\pi}\!\left[
r_t^{\mathrm{cont}}
+
V_{t+1}^{\star}(m_{t+1})
\,\middle|\, m_t=m,\ z_t=0
\right]
\label{eq:app_stop_continue}
\end{equation}
is the optimal value of continuing for one more step and then behaving optimally thereafter.
Here \(r_t^{\mathrm{cont}}\) already includes the marginal computation cost of taking another reasoning step.

\paragraph{Continuation advantage.}
Define the continuation advantage
\begin{equation}
\Delta_t(m)
:=
V_t^{\mathrm{cont}}(m)-V_t^{\mathrm{stop}}(m).
\label{eq:app_stop_delta}
\end{equation}
The sign of \(\Delta_t(m)\) determines whether an additional step is worthwhile under the current endogenous state:
continuation is beneficial when \(\Delta_t(m)>0\), and stopping is optimal when \(\Delta_t(m)\le 0\).

\paragraph{Proposition J.1.}
Assume that \(m_t\) is sufficient for the stopping decision in the sense that both \(V_t^{\mathrm{stop}}\) and \(V_t^{\mathrm{cont}}\) are measurable functions of \(m_t\).
Then an optimal bounded-cost stopping rule is state-conditioned and can be written as
\begin{equation}
z_t^{\star}(m_t)
=
\mathbf{1}\!\left\{\Delta_t(m_t)\le 0\right\}.
\label{eq:app_stop_rule}
\end{equation}
Equivalently, the stop region
\begin{equation}
\Omega_{\mathrm{stop}}
:=
\{m\in\mathbb{R}^{4}:\Delta_t(m)\le 0\}
\label{eq:app_stop_region}
\end{equation}
is optimal: the controller should stop exactly on \(\Omega_{\mathrm{stop}}\) and continue on its complement.

\paragraph{Proof.}
By Eq.~(\ref{eq:app_stop_bellman}), the optimal action at state \(m\) is whichever attains the larger of \(V_t^{\mathrm{stop}}(m)\) and \(V_t^{\mathrm{cont}}(m)\).
If \(\Delta_t(m)\le 0\), then by Eq.~(\ref{eq:app_stop_delta}),
\begin{equation}
V_t^{\mathrm{cont}}(m)\le V_t^{\mathrm{stop}}(m),
\end{equation}
so stopping weakly dominates continuation and is therefore optimal.
If \(\Delta_t(m)>0\), then
\begin{equation}
V_t^{\mathrm{cont}}(m)>V_t^{\mathrm{stop}}(m),
\end{equation}
so continuation is strictly optimal.
Hence the optimal decision is exactly
\begin{equation}
z_t^{\star}(m)=1 \iff \Delta_t(m)\le 0,
\end{equation}
which proves Eq.~(\ref{eq:app_stop_rule}).
Because \(V_t^{\mathrm{stop}}\) and \(V_t^{\mathrm{cont}}\) are measurable in \(m_t\), the resulting stopping rule is a measurable state-conditioned policy.
\hfill\(\square\)

\paragraph{Corollary J.1.}
Any stopping rule that depends only on a fixed external depth budget, a pre-specified schedule, or the step index \(t\), but not on the realized endogenous state \(m_t\), is generally suboptimal whenever \(\Delta_t(m)\) changes sign over a positive-probability set of realized states.

\paragraph{Proof.}
If \(\Delta_t(m)\) changes sign across realized states, then there exist states \(m\) and \(m'\) with positive probability such that stopping is optimal at \(m\) but continuation is optimal at \(m'\).
Any state-agnostic rule must assign the same decision to both states and therefore errs on at least one positive-probability subset.
By Proposition~I.1, the state-conditioned rule in Eq.~(\ref{eq:app_stop_rule}) attains the optimal action on both.
Thus the state-agnostic rule is strictly suboptimal in expectation.
\hfill\(\square\)

\paragraph{Implication.}
Proposition~J.1 shows that stopping is the bounded-cost counterpart of evidence organization.
Both are decisions about whether additional reasoning support is worth carrying forward under the current endogenous state.
Corollary~J.1 further explains why SoT does not use a fixed reasoning depth or an external stopping schedule: when the marginal value of another step varies across realized reasoning regimes, rational stopping must itself be state-conditioned.
This is why the stop controller is part of the same closed-loop reasoning mechanism rather than an afterthought.

\clearpage
\AppBlock{Part C: Experimental Setup and Implementation}
\vspace{-4mm}
\section{Dataset}
\label{app:dataset_setup}
\vspace{-2mm}

\begin{table*}[!b]
\centering
\vspace{-2mm}
\caption{\textbf{Dataset-level organization, split size, and primary metric.}}
\vspace{-2mm}
\label{tab:app_dataset_metrics}
\footnotesize
\setlength{\tabcolsep}{4.2pt}
\renewcommand{\arraystretch}{0.95}
\begin{tabular}{llcccl}
\toprule
\textbf{Category} & \textbf{Dataset} & \textbf{Task type} & \textbf{Train} & \textbf{Test} & \textbf{Metric} \\
\midrule

\multirow{3}{*}{\makecell[l]{Quantitative\\Reasoning (QS)}}
& GSM8K & Math QA & --- & 500 & EM \\
& MATH & Competition math & --- & 400 & EM \\
& DROP & Discrete reasoning QA & --- & 300 & F1 \\
\midrule

\multirow{5}{*}{\makecell[l]{General\\Understanding\\(GU)}}
& CommonsenseQA (CSQA) & Multiple choice QA & 52 & 400 & EM \\
& StrategyQA (StrQA) & Binary QA & 52 & 400 & EM \\
& BoolQ & Boolean QA & --- & 400 & EM \\
& MMLU & Multi-domain MCQA & 52 & 500 & EM \\
& RACE & Reading comprehension & --- & 300 & EM \\
\midrule

\multirow{5}{*}{\makecell[l]{Symbolic and\\Code (S\&C)}}
& FOLIO & Logical entailment & --- & 203 & EM \\
& ProofWriter (PW) & Deductive QA & 52 & 400 & EM \\
& BBH-Temporal (BBH-T) & Temporal reasoning & 52 & 100 & EM \\
& HumanEval (HE) & Code generation & --- & 164 & P@1 \\
& MBPP & Code generation & 52 & 250 & P@1 \\
\midrule

\multirow{3}{*}{\makecell[l]{Long-Context\\Reasoning (LCR)}}
& HotpotQA (HQA) & Multi-hop QA & 52 & 300 & F1 \\
& NarrativeQA (NarQA) & Narrative QA & --- & 250 & F1 \\
& LongBench MultiFieldQA (LB) & Long-Context QA & --- & 110 & F1 \\
\midrule

\multirow{3}{*}{\makecell[l]{Vision--language\\multimodal (VLM)}}
& {A-OKVQA} & {Knowledge-intensive VQA} & --- & {200} & {EM} \\
& {AI2D} & {Science-diagram reasoning} & --- & {150} & {EM} \\
& {M$^3$CoT} & {Multi-step multimodal reasoning} & --- & {150} & {EM} \\
\bottomrule
\end{tabular}
\vspace{-1mm}
\begin{flushleft}
    \scriptsize
    \textbf{Note}:
    \textbf{EM} = exact match, \textbf{F1} = token-level F1, \textbf{P@1} = pass@1.
\end{flushleft}
\vspace{-3mm}
\end{table*}

\subsection{Category organization}
\vspace{-1mm}
The LLM datasets are grouped into four categories:

\begin{itemize}[leftmargin=1.5em]
    \item \textbf{Quantitative Reasoning (QS):}
    GSM8K~\citep{cobbe2021gsm8k} (grade-school arithmetic word problems), MATH~\citep{hendrycks2021measuring} (competition-level mathematical reasoning), and DROP~\citep{dua2019drop} (discrete reading comprehension with numerical operations).

    \item \textbf{General Understanding (GU):}
    CommonsenseQA~\citep{talmor2019commonsenseqa} (commonsense multiple-choice QA), StrategyQA~\citep{geva2021did} (implicit multi-hop yes/no reasoning), BoolQ~\citep{clark2019boolq} (passage-grounded boolean QA), MMLU~\citep{hendrycks2021measuringmassive} (multi-domain professional and academic knowledge), and RACE~\citep{lai2017race} (exam-style reading comprehension).

    \item \textbf{Symbolic and Code (S\&C):}
    FOLIO~\citep{han2024folio} (first-order logical entailment), ProofWriter~\citep{tafjord2021proofwriter} (multi-step rule-based deduction), BBH-Temporal~\citep{srivastava2023beyond} (temporal symbolic reasoning), HumanEval~\citep{chen2021evaluating} (function-level code generation), and MBPP~\citep{austin2021program} (program synthesis from natural-language specifications).

    \item \textbf{Long-Context Reasoning (LCR):}
    HotpotQA~\citep{yang2018hotpotqa} (multi-hop evidence aggregation), NarrativeQA~\citep{kocisky2018narrativeqa} (long-form story understanding), and LongBench MultiFieldQA~\citep{bai2024longbench} (long-context question answering across heterogeneous documents).
\end{itemize}

{For multimodal transfer, we use A-OKVQA~\citep{schwenk2022aokvqa} for knowledge-intensive visual question answering, AI2D~\citep{kembhavi2016ai2d} for science-diagram reasoning, and M$^3$CoT~\citep{chen2024m3cot} for multi-domain, multi-step multimodal reasoning.}

\vspace{-2mm}
\subsection{Metrics}
\vspace{-1mm}
Table~\ref{tab:app_dataset_metrics} summarizes the dataset-level organization, task type, split size, and primary metric used in our experiments.
The headline ``accuracy'' in the main text is a unified label for the \emph{task-standard final metric} of each benchmark.

We use the following primary metrics:
\begin{itemize}[leftmargin=1.5em]
    \item \textbf{EM} for classification-style, short-answer, logical-label, and visual recognition tasks;
    \item \textbf{F1} for long-form QA and evidence-aggregation tasks;
    \item \textbf{P@1} for code-generation benchmarks.
\end{itemize}

{For efficiency, we report generated tokens and end-to-end latency for all models:}
\begin{itemize}[leftmargin=1.5em]
    \item \textbf{Tokens}: the number of generated reasoning tokens per example, averaged over the evaluation set;
    \item \textbf{Latency}: the end-to-end wall-clock inference time under the same hardware and decoding.
\end{itemize}

\section{Model Configuration and Evaluation Protocol}
\label{app:model_protocol}

\paragraph{Backbone configuration.}
For text-only reasoning, we evaluate frozen Llama-3.1-8B~\citep{grattafiori2024llama3}, Qwen2.5-14B~\citep{yang2024qwen25}, and Mixtral-8$\times$7B~\citep{jiang2024mixtral}.
{For multimodal transfer, we evaluate frozen Qwen2.5-VL-7B and Qwen2.5-VL-32B backbones~\citep{bai2025qwen25vl}; the 32B block uses BF16 weights throughout.}
In all cases, backbone weights remain fixed throughout; any learned component is confined to the lightweight SoT controller introduced in Section~\ref{sec:state_driven_activation}.

\paragraph{Unified evaluation protocol.}
{All methods are evaluated from the same benchmark splits under a shared protocol that fixes decoding settings, inference budgets, answer extraction, and metric computation.}
{For each run, the evaluation subset is fixed before inference, with the same decoding caps and comparable search or sampling budgets when applicable; completed-example counts are retained with the evaluation artifacts and used for sample-weighted aggregates.}
Task outputs are then passed through the same dataset-specific answer extraction, normalization, and scoring functions.
Accordingly, SoT changes only the internal organization of evidence during reasoning; it does not benefit from a different evaluator, parser, or metric definition relative to the baselines.
{Text-backbone latency is measured on H200 GPUs. VLM latency is measured on one dedicated 80GB A100.}

\paragraph{Consistency across baseline families.}
The comparison is controlled at three levels.
First, all methods use the same frozen backbone within each experiment.
Second, all methods follow the same prompt--generation--parsing pipeline except for the reasoning mechanism specific to that baseline family.
Third, efficiency statistics are collected under the same runtime interface.
Generated tokens are counted as the total reasoning tokens produced per example, including internal generations for multi-pass methods, and latency is measured as end-to-end wall-clock inference time under the same hardware and decoding setup.
This ensures that both quality and efficiency comparisons are mechanism-level rather than implementation-level.

\paragraph{Training-side consistency for methods with learned components.}
{Whenever a method introduces a learned component beyond the frozen backbone, it is fit only on a pre-specified training-side corpus, with evaluation questions excluded in advance.}
SoT follows the same rule: its evidence-selection and stopping operators are trained only from offline reasoning trajectories derived from the frozen backbone, without access to evaluation instances.
This keeps train--evaluation separation uniform across all methods requiring lightweight fitting and prevents gains from arising from unequal data exposure.

\paragraph{Prompting and post-processing.}
For baseline families that share a common reasoning prompt style, SoT uses the same user-level prompt format; differences are confined to the internal control policy rather than surface prompt advantages.
Final predictions for all methods are extracted by the same dataset-specific post-processing rules, and the primary metrics reported in Table~\ref{tab:app_dataset_metrics} are always computed by the same evaluation routines.

\vspace{-1mm}
\section{Baseline Methods}
\label{app:baseline_methods}
\vspace{-1mm}
\subsection{Direct generation}
\vspace{-1mm}

\paragraph{\textbf{Vanilla}.}
Vanilla performs direct greedy answering without explicit intermediate reasoning control.
It serves as the minimal frozen-backbone reference.

\vspace{-1mm}
\subsection{Reasoning paradigms}
\vspace{-1mm}

\paragraph{\textbf{CoT}.}
Chain-of-Thought (CoT) prompting elicits reasoning through an externally specified step-by-step textual scaffold \citep{wei2022chain}.
It is the canonical baseline for scripted single-trajectory reasoning.

\paragraph{\textbf{PS}.}
Plan-and-Solve (PS) separates reasoning into an explicit planning stage followed by execution \citep{wang2023plan}.
Compared with CoT, it imposes a more structured external decomposition before answering.

\paragraph{\textbf{SR}.}
Self-Refine (SR) uses an externally scripted draft--feedback--revision loop to iteratively improve an initial response \citep{madaan2023self}.
Its improvement comes from explicit textual self-correction rather than endogenous control of carried evidence.

\paragraph{\textbf{SC}.}
Self-Consistency (SC) samples multiple reasoning traces and selects the final answer by agreement across sampled chains \citep{wang2023selfconsistency}.
It represents sample-based test-time scaling through breadth rather than state-conditioned control.

\paragraph{\textbf{BoN}.}
Best-of-$N$ (BoN) samples five candidate solutions and uses a fixed zero-temperature selector pass to choose the final answer from those candidates \citep{snell2024scaling}.
It therefore tests candidate-level selection beyond SC's answer-agreement rule; all candidate and selector completions are included in its reported cost.

\paragraph{\textbf{CB}.}
Constrained Beam (CB) allocates extra inference budget through beam-style structured decoding under explicit search constraints \citep{hokamp2017lexically}.
It provides a search-oriented comparison point for whether better reasoning must come from broader trajectory exploration.

\paragraph{\textbf{MCTS}.}
Monte Carlo Tree Search (MCTS) performs explicit search over reasoning branches using rollout-based lookahead and selection \citep{xie2024mcts}.
It is a strong representative of search-heavy reasoning methods that trade additional compute for broader exploration.

\vspace{-1mm}
\subsection{Memory-oriented methods}
\vspace{-1mm}

\paragraph{\textbf{H\(_2\)O}.}
H\(_2\)O prunes attention history by retaining tokens estimated to be important for generation \citep{zhang2023h2o}, reducing effective context usage.

\paragraph{\textbf{SNAP}.}
SnapKV accelerates long-context inference by compressing KV cache content through observation of stable attention patterns \citep{li2024snapkv}.
Its objective is efficient context management rather than reasoning-specific evidence control.

\paragraph{\textbf{STREAM}.}
StreamingLLM maintains stable long-context decoding by preserving a small set of attention anchors together with a streaming context window \citep{xiao2023streamingllm}.
It is an important efficiency baseline because it also avoids full-history accumulation.

\begin{table*}[!t]
\centering
\caption{\textbf{Deployment-side inference hyperparameters of SoT.} We report only externally specified quantities used at test time.}
\vspace{-2mm}
\label{tab:app_hparam_sot}
\footnotesize
\setlength{\tabcolsep}{5.2pt}
\renewcommand{\arraystretch}{0.95}
\begin{tabular}{lll}
\toprule
\textbf{Parameter} & \textbf{Value} & \textbf{Notes} \\
\midrule
Context budget & 1536 & Active evidence budget \\
Maximum reasoning steps & 32 & Safety cap only \\
Observation window & 4096 & Visible context span \\
Final answer budget & 512 & Answer generation cap \\
Stop threshold & 0.5 & Default stop rule \\
Gate threshold & 0.5 & Default keep rule \\
Score mixing coefficient & 0.5 & Score fusion only \\
Score temperature & 0.8 & Ranking smoothing only \\
\bottomrule
\end{tabular}
\vspace{1mm}

\centering
\caption{\textbf{Inference hyperparameters of baseline methods.}}
\vspace{-2mm}
\label{tab:app_hparam_baseline}
\footnotesize
\setlength{\tabcolsep}{6pt}
\renewcommand{\arraystretch}{0.93}

\begin{minipage}[t]{0.49\textwidth}
\centering
\begin{tabular}{lll}
\toprule
\textbf{Baseline} & \textbf{Parameter} & \textbf{Value} \\
\midrule

\multirow{2}{*}{Vanilla}
& Maximum new tokens & 512 \\
& Temperature & 0 \\
\midrule

\multirow{2}{*}{CoT}
& Maximum new tokens & 768 \\
& Temperature & 0 \\
\midrule

\multirow{3}{*}{PS}
& Planning tokens & 96 \\
& Solving tokens & 320 \\
& Temperature & 0 \\
\midrule

\multirow{4}{*}{SR}
& Draft tokens & 256 \\
& Critique tokens & 96 \\
& Revision tokens & 256 \\
& Temperature & 0 \\
\midrule

\multirow{3}{*}{SC}
& Number of samples & {3 (LLM); 5 (VLM)} \\
& Tokens per sample & 384 \\
& Temperature & 0.7 \\
\midrule

\multirow{5}{*}{BoN}
& Number of candidates & 5 \\
& Candidate token cap & 768 \\
& Candidate temperature & 0.7 \\
& Selector token cap & 64 \\
& Selector temperature & 0 \\
\midrule

\multirow{8}{*}{CB}
& Search depth & 3 \\
& Beam width & 3 \\
& Branches per beam & 2 \\
& Step tokens & 56 \\
& Scoring tokens & 12 \\
& Answer tokens & 256 \\
& Expansion temp. & 0.4 \\
& Tree token budget & 3200 \\
\midrule

MCTS & Search iterations & 6 \\
\bottomrule
\end{tabular}
\end{minipage}
\hfill
\begin{minipage}[t]{0.49\textwidth}
\centering
\begin{tabular}{lll}
\toprule
\textbf{Baseline} & \textbf{Parameter} & \textbf{Value} \\
\midrule

\multirow{7}{*}{MCTS}
& Expansion branches & 2 \\
& Step tokens & 56 \\
& Scoring tokens & 12 \\
& Answer tokens & 256 \\
& Exploration const. & 1.2 \\
& Expansion temp. & 0.5 \\
& Tree token budget & 3200 \\
\midrule

\multirow{5}{*}{H2O}
& Sink tokens & 4 \\
& Recent tokens & 256 \\
& Context budget & 512 \\
& Maximum new tokens & 512 \\
& Temperature & 0 \\
\midrule

\multirow{5}{*}{SNAP}
& Sink tokens & 4 \\
& Recent tokens & 256 \\
& Context budget & 512 \\
& Maximum new tokens & 512 \\
& Temperature & 0 \\
\midrule

\multirow{5}{*}{STREAM}
& Sink tokens & 4 \\
& Recent tokens & 256 \\
& Context budget & 512 \\
& Maximum new tokens & 512 \\
& Temperature & 0 \\
\midrule

\multirow{3}{*}{COCO}
& Latent steps & 3 \\
& Maximum new tokens & 512 \\
& Temperature & 0 \\
\midrule

\multirow{2}{*}{\makecell{\scriptsize{GRPO-SP}}}
& Maximum new tokens & 768 \\
& Temperature & 0 \\
\bottomrule
\end{tabular}
\end{minipage}
\vspace{-3mm}
\end{table*}

\vspace{-1mm}
\subsection{Latent and RL-based reasoning}
\vspace{-1mm}

\paragraph{\textbf{COCO}.}
COCONUT performs reasoning partly in continuous latent space rather than purely through explicit textual chains \citep{hao2025training}.
It is the closest latent-reasoning comparison, but differs from SoT in that it shifts the reasoning substrate itself rather than using endogenous state to organize evidence.

\paragraph{\textbf{\,{GRPO-SP}}.}
{GRPO-SP is a lightweight RL-based comparison inspired by grouped relative policy optimization \citep{shao2024deepseekmath}. It freezes the backbone and applies grouped reward-normalized policy-gradient updates only to a four-token continuous prompt, testing whether a compact reward-trained inference prefix can recover reasoning gains without backbone updates.}

\paragraph{\textbf{Comparison logic}.}
These baselines probe complementary alternatives to SoT.
Vanilla tests whether explicit reasoning control is necessary.
CoT, PS, and SR test externally prescribed single-trajectory programs.
SC, BoN, CB, and MCTS test whether gains instead come from larger test-time search.
H\(_2\)O, SNAP, and STREAM test whether generic context reduction is sufficient.
COCO and \,{GRPO-SP} test latent-space or optimization-based alternatives to inference-time state-conditioned control.
Together, they provide the comparison surface needed to show that SoT is not reducible to prompting, search, memory compression, latent reasoning, or RL-style behavioral optimization.

\paragraph{Implementation and fairness.}
{Names follow the implemented algorithmic roles: SC selects by answer agreement, BoN uses a fixed selector over five sampled candidates, CB and MCTS perform structured search, and GRPO-SP is the grouped-advantage soft-prompt variant defined above. Each family uses one pre-specified configuration across datasets; sample counts, temperatures, and search budgets are reported in Table~\ref{tab:app_hparam_baseline}. All generated stages count toward efficiency, and all methods share the same task-specific answer extraction and metric.}

\vspace{-1mm}
\section{Hyperparameter Configuration}
\label{app:hyperparameters}
\vspace{-1mm}

Table~\ref{tab:app_hparam_sot} reports the deployment-side hyperparameters of SoT, while Table~\ref{tab:app_hparam_baseline} summarizes the corresponding hyperparameters for all baselines.

{The two score-smoothing parameters used by SoT---the attention mixing coefficient and attention temperature---are deployment-side stabilizers rather than additional reasoning modules: they only smooth or fuse ranking signals before thresholding, and do not change the state interface, control form, or backbone generation budget.}
The maximum reasoning-step budget is a deployment-side safeguard only.
Empirically, nearly all trajectories terminate through the learned stop policy before the hard cap is reached (about 99.6\% in our runs), so the reported efficiency gains are not driven by forced truncation.

For fairness, all methods are evaluated under the same backbone, data split, decoding interface, and scoring pipeline.
Each method is assigned only the minimal family-specific hyperparameters required for execution.
In particular, SoT does not receive additional model capacity, longer unrestricted generation, or a privileged prompt path relative to the baselines.
Accordingly, any improvement should be attributed to its state-conditioned control mechanism rather than to favorable hyperparameter budgeting.
Supplementary ablations also help verify that the observed gains are not driven by fragile score calibration or deployment-side smoothing choices, but by the full state-conditioned control mechanism.

\clearpage
\AppBlock{Part D: Additional Experimental Results and Analysis}

\section{Interpretable Evidence for State-Conditioned Reasoning}
\label{app:interpretability}

This section extends the interpretability evidence in Secs.~\ref{sec:endogenous_state} and~\ref{sec:state_driven_activation}, together with SoT as a trajectory-level judge in {Sec.}~\ref{sec:judge}.
Our goal is to show that SoT is not merely a strong inference-time heuristic: its gains are accompanied by stable and interpretable state-space structure, regime-dependent evidence organization, and measurable differences between successful and unsuccessful reasoning trajectories.

\subsection{State-space trajectories across reasoning paradigms}
\label{app:traj_paradigms}

We first extend the state-space visualization in Sec.~\ref{sec:endogenous_state} by covering all four task categories and a broader set of reasoning paradigms.
For each method, we project the trajectory of \(m_t\) into a shared two-dimensional space and visualize the stepwise evolution across representative examples.
The comparison is designed to answer two questions:
whether different external reasoning paradigms induce distinct trajectory regions or motion patterns, and whether SoT occupies a broader and more adaptive portion of the state space rather than collapsing to a fixed scripted mode.

\begin{figure*}[!b]
\centering
\includegraphics[width=1.0\textwidth]{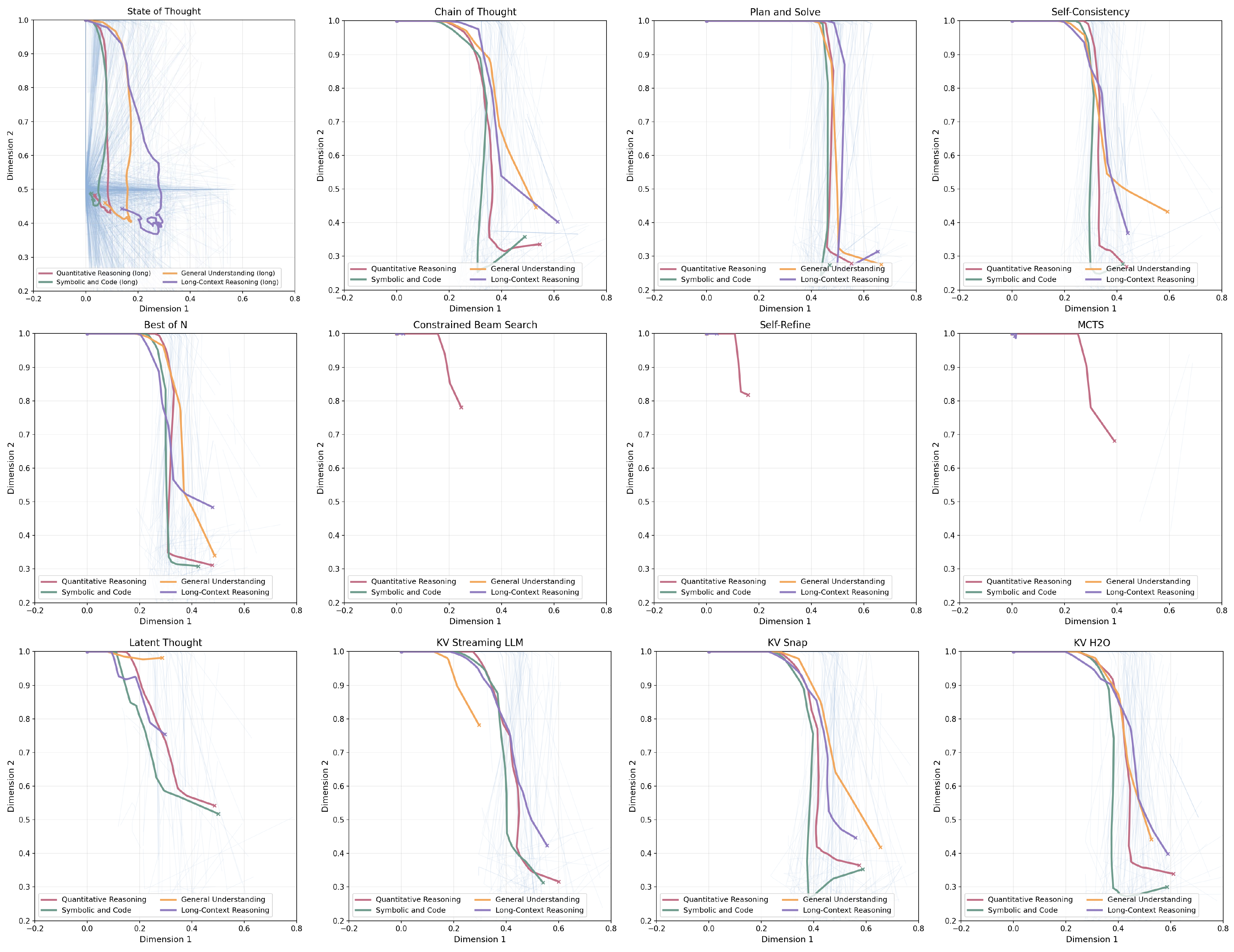}
\caption{
\textbf{Extended state-space trajectories across reasoning paradigms.}
This figure supplements the main-text trajectory view with additional paradigms and task categories.
}
\label{fig:app_state_regimes}
\end{figure*}

\paragraph{Interpretation.}
As shown in Figure~\ref{fig:app_state_regimes}, this analysis does not claim that each paradigm maps to a discrete symbolic state label.
Instead, it shows that the compact state \(m_t\) carries stable geometric structure: different paradigms trace distinguishable trajectory regions, while SoT occupies a broader and more adaptive regime rather than collapsing to a fixed externally scripted pattern.
The key insight is that SoT's gain is associated with controllable state coverage---reaching multiple useful regimes when needed---rather than with a single pre-committed reasoning template.

\subsection{State clusters, sparse memory, and stopping}
\label{app:cluster_ops}

We next extend the operator-level analysis of Sec.~\ref{sec:state_driven_activation}.
After aggregating SoT states across tasks, we cluster the observed \(m_t\) and summarize, for each cluster,
(i) its occupancy across task categories,
(ii) the average sparsity pattern of \(\mathcal{S}(y_{<t};m_t)\), and
(iii) the stop tendency induced by \(\mathcal{T}(m_t)\).
This links the geometry of the state space to concrete control behavior.

\begin{figure}[H]
\centering
\includegraphics[width=1.0\textwidth]{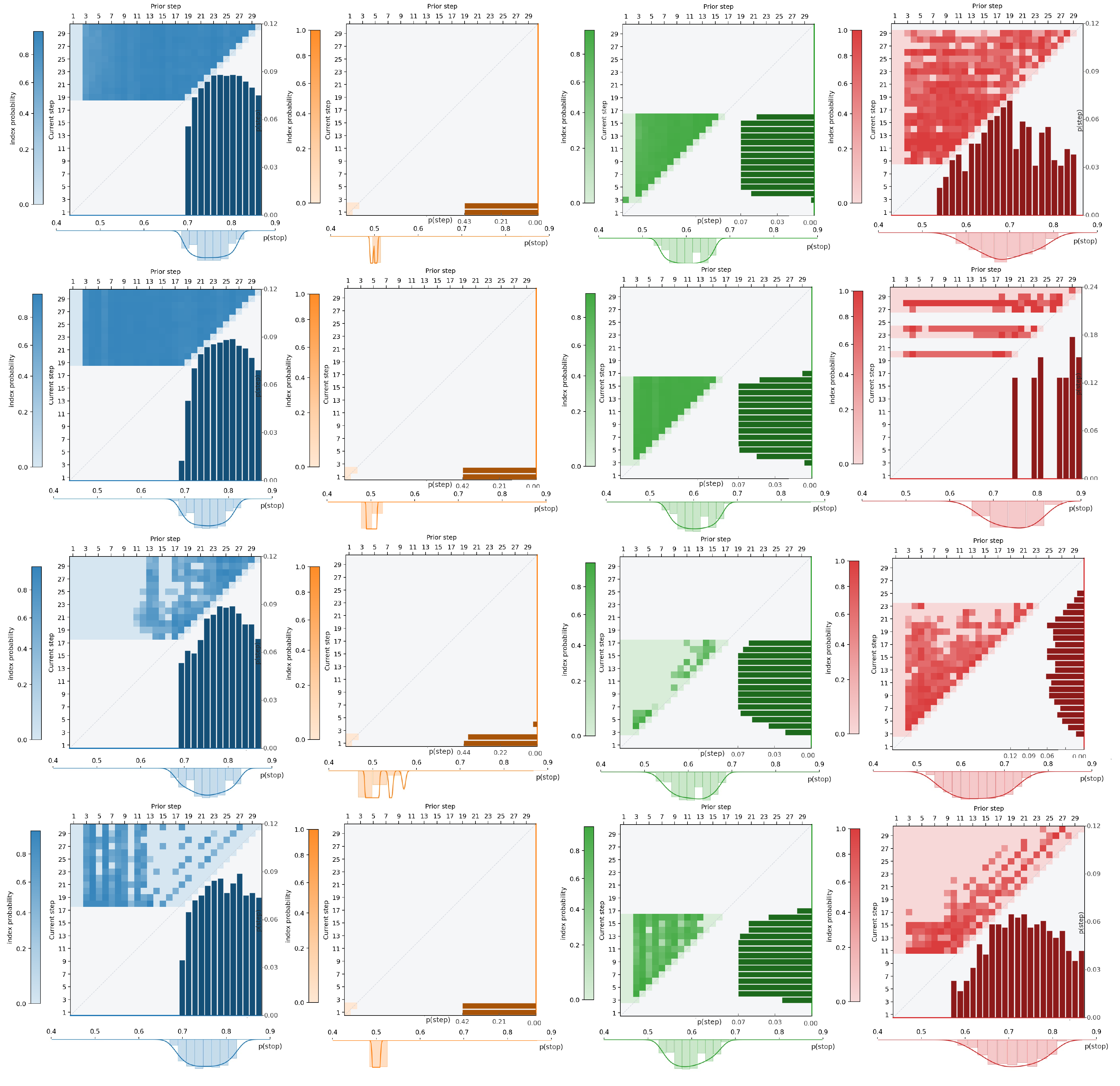}
\vspace{-4mm}
\caption{
\textbf{Clustered state regimes and operator profiles.}
}
\vspace{-4mm}
\label{fig:app_state_clusters}
\end{figure}

\paragraph{Interpretation.}
Figure~\ref{fig:app_state_clusters} tests whether clustered states correspond to distinct control behavior, not just visual separability.
The observed cluster-specific sparsity and stop profiles indicate regime-dependent evidence organization and stopping, rather than generic recency bias, static compression, or a fixed-depth schedule.
This suggests that SoT's policy is conditionally compositional: the state first selects \emph{what} evidence remains active and then implicitly calibrates \emph{when} to terminate.
{The four labels---\emph{exploit-dense}, \emph{explore-dense}, \emph{early-anchor}, and \emph{sparse-retrieval}---are assigned only after clustering. We inspect each normalized centroid $(\delta,v,c,H)$ together with three measured properties: activation density, the temporal location of retained units, and stop probability. ``Dense'' and ``sparse'' therefore describe the activation matrix; ``early-anchor'' denotes concentration on early units; and the explore--exploit distinction additionally requires the centroid dynamics and stop profile to indicate continued movement or consolidation. These labels never enter fitting or evaluation. The evidence for distinct regimes is the operator profiles themselves, not the names attached to them.}

\subsection{Correct versus incorrect trajectories under internal and embedding-trajectory SoT}
\label{app:correct_wrong_traj}

Finally, we compare successful and unsuccessful trajectories under both the full SoT state and the SoT-Embed variant from Sec.~\ref{sec:endogenous_state} and {Sec.}~\ref{sec:judge}.
For each of the four task categories, we aggregate trajectories by correctness and visualize the resulting state-space distributions.
We perform the same analysis twice:
once using the internal state derived from {internal information transfer},
and once using features derived from the sentence-embedding trajectory ({Appendix}~\ref{app:limitations_sot}).
This allows us to test whether the separation between healthy and unhealthy reasoning trajectories is preserved even when privileged internal access is removed.

\begin{figure*}[t]
\centering
\includegraphics[width=1.0\textwidth]{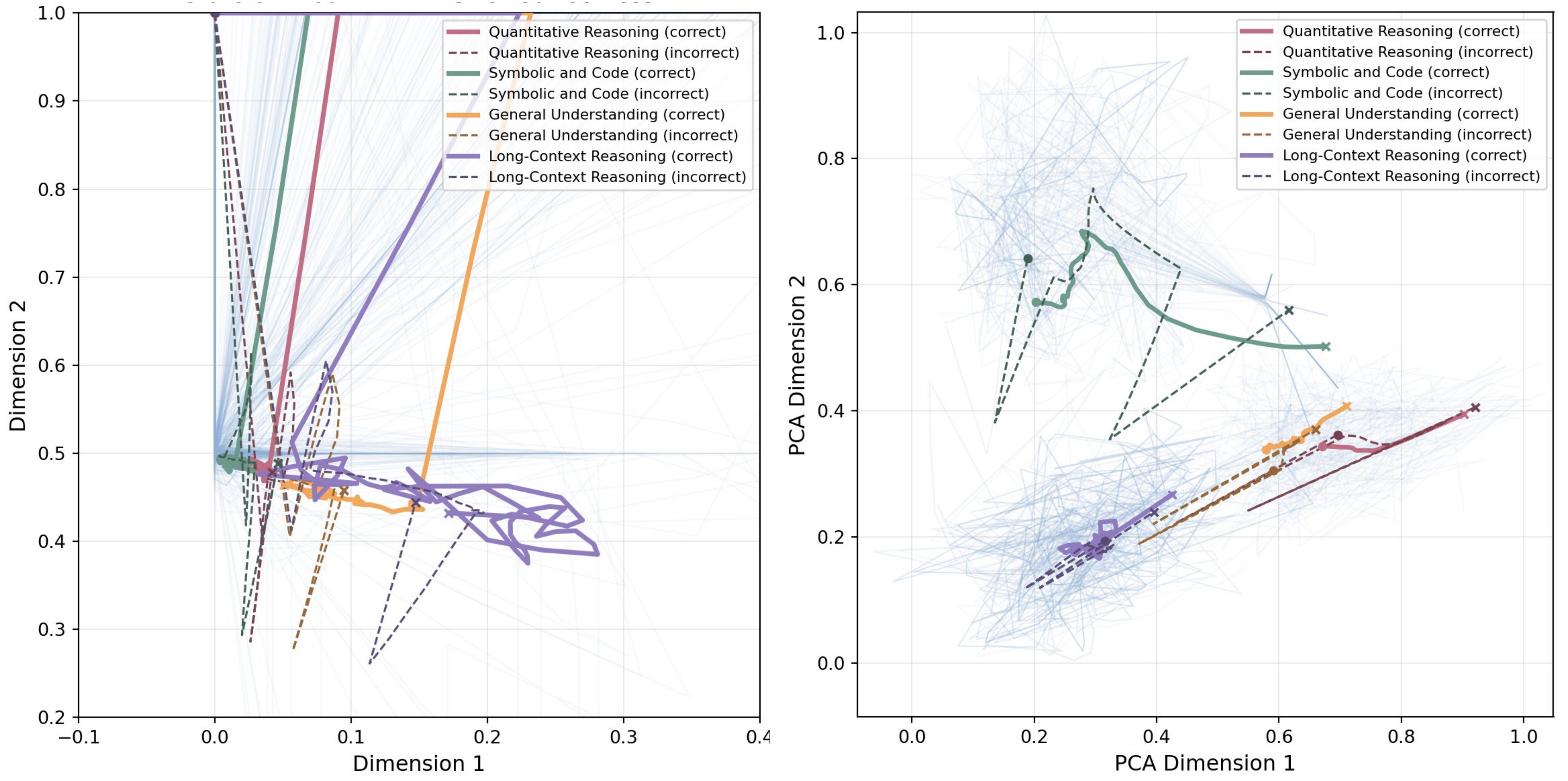}
\vspace{-5mm}
\caption{
\textbf{Correct and incorrect trajectory regimes under full-state and embedding-trajectory SoT.}
This figure compares successful and unsuccessful reasoning trajectories across all four task categories,
under both the internal-state SoT and SoT-Embed.
}
\vspace{-5mm}
\label{fig:app_correct_incorrect}
\end{figure*}

\paragraph{Interpretation.}
Figure~\ref{fig:app_correct_incorrect} serves two roles.
For full SoT, it links gains to trajectory quality by showing whether correct runs occupy more stable and favorable state regimes than incorrect runs.
For SoT-Embed, it tests whether this correct--incorrect separation persists under limited access; the preserved gap supports that SoT is a transferable trajectory-level principle rather than a privileged internal-state trick.
The practical insight is that trajectory-state separability can function as a lightweight health signal for reliability monitoring, even when white-box internals are unavailable.

\section{Additional Results}
\label{app:additional_results}

This section provides supplementary result tables and figures omitted from the main text for space.
{Specifically, we provide backbone-specific main-result tables, token and latency breakdowns, and expanded boundary studies for the limited-access variants.}

\subsection{Additional backbone results}

Table~\ref{tab:app_main_qwen} and Table~\ref{tab:app_main_mixtral} report the full per-dataset accuracy results for the two LLM backbones not shown in the main text.
Two patterns are notable.
On Qwen2.5-14B, SoT achieves the best domain averages in QS (\(66.1\)), S\&C (\(76.8\)), and GU (\(81.4\)); the GU gain is especially clear, with all five GU datasets reaching the highest scores in the table (CSQA \(80.5\), StrQA \(75.0\), BoolQ \(86.8\), MMLU \(80.0\), RACE \(86.0\)).
On Mixtral-8x7B, SoT is near-best in QS (\(46.3\), close to \(46.4\)) and leads clearly in S\&C (\(51.7\)), GU (\(72.6\)), and LCR (\(23.5\)), indicating that the same state-conditioned control remains effective across substantially different backbone characteristics.
Taken together, these cross-backbone results support a mechanism-level claim: SoT's advantage is not tied to one architecture or one dataset family, but to a transferable control principle that improves diverse reasoning regimes under heterogeneous model priors.

\begin{table*}[!t]
\centering
\caption{\textbf{Main results on Qwen2.5-14B (accuracy, \%).}}
\label{tab:app_main_qwen}
\vspace{-2mm}
\footnotesize
{\setlength{\tabcolsep}{1.3pt}%
\setlength{\aboverulesep}{0pt}%
\setlength{\belowrulesep}{0pt}%
\setlength{\extrarowheight}{1.43pt}%
\renewcommand{\arraystretch}{0.823}%
\renewcommand{\tabularxcolumn}[1]{m{#1}}%
\begin{tabularx}{\linewidth}{@{} >{\centering\arraybackslash}p{1.65cm} >{\centering\arraybackslash}m{1.28cm} !{\vrule width 0.4pt} >{\centering\arraybackslash}X >{\centering\arraybackslash}X >{\centering\arraybackslash}X >{\centering\arraybackslash}m{3.1em} >{\centering\arraybackslash}X >{\centering\arraybackslash}X >{\centering\arraybackslash}X >{\centering\arraybackslash}X >{\centering\arraybackslash}X >{\centering\arraybackslash}m{3.1em} @{}}
\toprule
\multirow{2}{=}{\centering\textbf{Category}} & \multirow{2}{=}{\centering\textbf{Method}}
& \multicolumn{4}{>{\cellcolor[HTML]{E3F2FD}}c!{\vrule width 1pt}}{\textbf{Quantitative Reasoning}}
& \multicolumn{6}{>{\cellcolor[HTML]{E3F2FD}}c}{\textbf{Symbolic and Code}} \\
&
& {\small\textbf{GSM}} & {\small\textbf{MATH}} & {\small\textbf{DROP}} & \multicolumn{1}{>{\centering\arraybackslash}m{3.1em}!{\vrule width 1pt}}{\small\textbf{avg.}}
& {\small\textbf{FOL}} & {\small\textbf{PW}} & {\small\textbf{BBH}} & {\small\textbf{HE}} & {\small\textbf{MBPP}} & \multicolumn{1}{>{\centering\arraybackslash}m{3.1em}}{\small\textbf{avg.}} \\
\midrule
\multirow{6}{=}{\centering Reasoning\\Paradigm} & CoT & \cellcolor[HTML]{FFD4D3}{94.4} & \cellcolor[HTML]{FFD4D3}{77.5} & 1.9 & \multicolumn{1}{>{\centering\arraybackslash}m{3.1em}!{\vrule width 1pt}}{\cellcolor[HTML]{EAECEF}{65.6}} & 0.0 & 55.5 & 84.0 & \cellcolor[HTML]{FCE4D5}{75.0} & \cellcolor[HTML]{FFD4D3}{66.0} & \multicolumn{1}{>{\centering\arraybackslash}m{3.1em}}{\cellcolor[HTML]{EAECEF}{49.9}} \\
 & PS & 88.0 & 51.5 & 2.1 & \multicolumn{1}{>{\centering\arraybackslash}m{3.1em}!{\vrule width 1pt}}{\cellcolor[HTML]{EAECEF}{54.4}} & 0.5 & 65.5 & \cellcolor[HTML]{FFD4D3}{92.0} & 50.0 & 47.2 & \multicolumn{1}{>{\centering\arraybackslash}m{3.1em}}{\cellcolor[HTML]{EAECEF}{49.6}} \\
 & SR & 92.0 & 68.2 & \cellcolor[HTML]{FFFCE0}{17.6} & \multicolumn{1}{>{\centering\arraybackslash}m{3.1em}!{\vrule width 1pt}}{\cellcolor[HTML]{EAECEF}{65.5}} & 61.1 & 83.4 & 79.0 & 60.0 & 50.4 & \multicolumn{1}{>{\centering\arraybackslash}m{3.1em}}{\cellcolor[HTML]{EAECEF}{68.1}} \\
 & SC & \cellcolor[HTML]{FFFCE0}{92.8} & 61.0 & 1.9 & \multicolumn{1}{>{\centering\arraybackslash}m{3.1em}!{\vrule width 1pt}}{\cellcolor[HTML]{EAECEF}{59.5}} & 0.5 & 59.5 & \cellcolor[HTML]{FFFCE0}{87.0} & 55.0 & 48.8 & \multicolumn{1}{>{\centering\arraybackslash}m{3.1em}}{\cellcolor[HTML]{EAECEF}{47.2}} \\
 & CB & 60.0 & 41.2 & \cellcolor[HTML]{FFD4D3}{30.8} & \multicolumn{1}{>{\centering\arraybackslash}m{3.1em}!{\vrule width 1pt}}{\cellcolor[HTML]{EAECEF}{46.4}} & \cellcolor[HTML]{FFFCE0}{62.1} & \cellcolor[HTML]{FFFCE0}{85.0} & 51.0 & 50.0 & 60.8 & \multicolumn{1}{>{\centering\arraybackslash}m{3.1em}}{\cellcolor[HTML]{EAECEF}{69.8}} \\
 & MCTS & 66.2 & 50.0 & \cellcolor[HTML]{FCE4D5}{19.3} & \multicolumn{1}{>{\centering\arraybackslash}m{3.1em}!{\vrule width 1pt}}{\cellcolor[HTML]{EAECEF}{49.1}} & \cellcolor[HTML]{FCE4D5}{63.1} & \cellcolor[HTML]{FFD4D3}{88.5} & 53.0 & 60.0 & \cellcolor[HTML]{FCE4D5}{63.6} & \multicolumn{1}{>{\centering\arraybackslash}m{3.1em}}{\cellcolor[HTML]{EAECEF}{72.6}} \\
\midrule
\multirow{3}{=}{\centering Memory-Oriented} & H2O & 92.6 & 67.0 & 2.6 & \multicolumn{1}{>{\centering\arraybackslash}m{3.1em}!{\vrule width 1pt}}{\cellcolor[HTML]{EAECEF}{61.6}} & 0.5 & 14.8 & 81.0 & 72.8 & 45.2 & \multicolumn{1}{>{\centering\arraybackslash}m{3.1em}}{\cellcolor[HTML]{EAECEF}{33.4}} \\
 & SNAP & \cellcolor[HTML]{FFFCE0}{92.8} & 66.5 & 2.7 & \multicolumn{1}{>{\centering\arraybackslash}m{3.1em}!{\vrule width 1pt}}{\cellcolor[HTML]{EAECEF}{61.5}} & 2.0 & 13.0 & 83.0 & \cellcolor[HTML]{FFFCE0}{73.5} & 44.0 & \multicolumn{1}{>{\centering\arraybackslash}m{3.1em}}{\cellcolor[HTML]{EAECEF}{33.1}} \\
 & STREAM & 78.0 & 45.8 & 1.8 & \multicolumn{1}{>{\centering\arraybackslash}m{3.1em}!{\vrule width 1pt}}{\cellcolor[HTML]{EAECEF}{48.2}} & 5.9 & 19.5 & 3.0 & 68.4 & 3.2 & \multicolumn{1}{>{\centering\arraybackslash}m{3.1em}}{\cellcolor[HTML]{EAECEF}{19.1}} \\
\midrule
Latent & COCO & \cellcolor[HTML]{FCE4D5}{93.6} & \cellcolor[HTML]{FCE4D5}{74.8} & 2.8 & \multicolumn{1}{>{\centering\arraybackslash}m{3.1em}!{\vrule width 1pt}}{\cellcolor[HTML]{EAECEF}{64.6}} & 2.5 & 40.5 & 75.0 & 71.2 & 60.0 & \multicolumn{1}{>{\centering\arraybackslash}m{3.1em}}{\cellcolor[HTML]{EAECEF}{45.6}} \\
\midrule
RL-Based & \makecell{\scriptsize{GRPO-SP}} & 0.6 & 5.0 & 0.0 & \multicolumn{1}{>{\centering\arraybackslash}m{3.1em}!{\vrule width 1pt}}{\cellcolor[HTML]{EAECEF}{1.9}} & 3.4 & 3.0 & 5.0 & 58.7 & 0.0 & \multicolumn{1}{>{\centering\arraybackslash}m{3.1em}}{\cellcolor[HTML]{EAECEF}{10.8}} \\
\midrule
Ours & \textbf{SoT} & \textbf{91.8} & \cellcolor[HTML]{FFFCE0}{\textbf{72.2}} & \textbf{15.1} & \multicolumn{1}{>{\centering\arraybackslash}m{3.1em}!{\vrule width 1pt}}{\cellcolor[HTML]{EAECEF}{\underline{\textbf{66.1}}}} & \cellcolor[HTML]{FFD4D3}{\textbf{66.5}} & \cellcolor[HTML]{FCE4D5}{\textbf{87.8}} & \cellcolor[HTML]{FCE4D5}{\textbf{91.0}} & \cellcolor[HTML]{FFD4D3}{\textbf{75.6}} & \cellcolor[HTML]{FFFCE0}{\textbf{62.8}} & \multicolumn{1}{>{\centering\arraybackslash}m{3.1em}}{\cellcolor[HTML]{EAECEF}{\underline{\textbf{76.8}}}} \\
\midrule\end{tabularx}\par\addvspace{0.35ex}}%
{\setlength{\tabcolsep}{1.3pt}%
\setlength{\aboverulesep}{0pt}%
\setlength{\belowrulesep}{0pt}%
\setlength{\extrarowheight}{1.43pt}%
\renewcommand{\arraystretch}{0.823}%
\renewcommand{\tabularxcolumn}[1]{m{#1}}%
\begin{tabularx}{\linewidth}{@{} >{\centering\arraybackslash}p{1.65cm} >{\centering\arraybackslash}m{1.28cm} !{\vrule width 0.4pt} >{\centering\arraybackslash}X >{\centering\arraybackslash}X >{\centering\arraybackslash}X >{\centering\arraybackslash}X >{\centering\arraybackslash}X >{\centering\arraybackslash}m{3.1em} >{\centering\arraybackslash}X >{\centering\arraybackslash}X >{\centering\arraybackslash}X >{\centering\arraybackslash}m{3.1em} @{}}
\multirow{2}{=}{\centering\textbf{Category}} & \multirow{2}{=}{\centering\textbf{Method}}
& \multicolumn{6}{>{\cellcolor[HTML]{E3F2FD}}c!{\vrule width 1pt}}{\textbf{General Understanding}}
& \multicolumn{4}{>{\cellcolor[HTML]{E3F2FD}}c}{\textbf{Long-Context Reasoning}} \\
&
& {\small\textbf{CSQA}} & {\small\textbf{StrQA}} & {\small\textbf{BoolQ}} & {\small\textbf{MMLU}} & {\small\textbf{RACE}} & \multicolumn{1}{>{\centering\arraybackslash}m{3.1em}!{\vrule width 1pt}}{\small\textbf{avg.}}
& {\small\textbf{HQA}} & {\small\textbf{NarQA}} & {\small\textbf{LB}} & \multicolumn{1}{>{\centering\arraybackslash}m{3.1em}}{\small\textbf{avg.}} \\
\midrule
\multirow{6}{=}{\centering Reasoning\\Paradigm} & CoT & 42.8 & 51.2 & 61.5 & 39.8 & 36.8 & \multicolumn{1}{>{\centering\arraybackslash}m{3.1em}!{\vrule width 1pt}}{\cellcolor[HTML]{EAECEF}{46.6}} & 2.0 & 4.6 & 18.4 & \multicolumn{1}{>{\centering\arraybackslash}m{3.1em}}{\cellcolor[HTML]{EAECEF}{5.7}} \\
 & PS & 58.0 & 58.4 & 42.0 & 48.6 & 40.2 & \multicolumn{1}{>{\centering\arraybackslash}m{3.1em}!{\vrule width 1pt}}{\cellcolor[HTML]{EAECEF}{49.9}} & 2.5 & 5.6 & 23.9 & \multicolumn{1}{>{\centering\arraybackslash}m{3.1em}}{\cellcolor[HTML]{EAECEF}{7.3}} \\
 & SR & \cellcolor[HTML]{FCE4D5}{75.0} & \cellcolor[HTML]{FCE4D5}{71.6} & 81.8 & \cellcolor[HTML]{FCE4D5}{74.0} & 63.4 & \multicolumn{1}{>{\centering\arraybackslash}m{3.1em}!{\vrule width 1pt}}{\cellcolor[HTML]{EAECEF}{73.7}} & 16.3 & 25.0 & \cellcolor[HTML]{FFD4D3}{30.8} & \multicolumn{1}{>{\centering\arraybackslash}m{3.1em}}{\cellcolor[HTML]{EAECEF}{22.0}} \\
 & SC & 40.2 & 55.8 & 61.8 & 38.6 & 48.6 & \multicolumn{1}{>{\centering\arraybackslash}m{3.1em}!{\vrule width 1pt}}{\cellcolor[HTML]{EAECEF}{48.5}} & 2.0 & 4.4 & 22.5 & \multicolumn{1}{>{\centering\arraybackslash}m{3.1em}}{\cellcolor[HTML]{EAECEF}{6.3}} \\
 & CB & 71.0 & \cellcolor[HTML]{FFFCE0}{70.2} & \cellcolor[HTML]{FFFCE0}{84.5} & \cellcolor[HTML]{FFFCE0}{67.8} & \cellcolor[HTML]{FFFCE0}{63.8} & \multicolumn{1}{>{\centering\arraybackslash}m{3.1em}!{\vrule width 1pt}}{\cellcolor[HTML]{EAECEF}{71.7}} & \cellcolor[HTML]{FFD4D3}{25.8} & \cellcolor[HTML]{FFD4D3}{33.5} & 26.3 & \multicolumn{1}{>{\centering\arraybackslash}m{3.1em}}{\cellcolor[HTML]{EAECEF}{\underline{28.8}}} \\
 & MCTS & \cellcolor[HTML]{FFFCE0}{71.2} & 69.4 & \cellcolor[HTML]{FCE4D5}{85.8} & 65.8 & 62.6 & \multicolumn{1}{>{\centering\arraybackslash}m{3.1em}!{\vrule width 1pt}}{\cellcolor[HTML]{EAECEF}{71.1}} & \cellcolor[HTML]{FFFCE0}{17.0} & \cellcolor[HTML]{FCE4D5}{29.0} & \cellcolor[HTML]{FCE4D5}{28.5} & \multicolumn{1}{>{\centering\arraybackslash}m{3.1em}}{\cellcolor[HTML]{EAECEF}{23.5}} \\
\midrule
\multirow{3}{=}{\centering Memory-Oriented} & H2O & 41.8 & 47.2 & 61.5 & 37.0 & 24.7 & \multicolumn{1}{>{\centering\arraybackslash}m{3.1em}!{\vrule width 1pt}}{\cellcolor[HTML]{EAECEF}{43.1}} & 2.0 & 2.0 & 0.0 & \multicolumn{1}{>{\centering\arraybackslash}m{3.1em}}{\cellcolor[HTML]{EAECEF}{1.7}} \\
 & SNAP & 41.8 & 47.2 & 61.0 & 37.4 & 27.0 & \multicolumn{1}{>{\centering\arraybackslash}m{3.1em}!{\vrule width 1pt}}{\cellcolor[HTML]{EAECEF}{43.4}} & 2.0 & 2.0 & 0.0 & \multicolumn{1}{>{\centering\arraybackslash}m{3.1em}}{\cellcolor[HTML]{EAECEF}{1.7}} \\
 & STREAM & 39.8 & 46.2 & 52.8 & 33.2 & 0.0 & \multicolumn{1}{>{\centering\arraybackslash}m{3.1em}!{\vrule width 1pt}}{\cellcolor[HTML]{EAECEF}{36.1}} & 2.4 & 0.0 & 0.0 & \multicolumn{1}{>{\centering\arraybackslash}m{3.1em}}{\cellcolor[HTML]{EAECEF}{1.1}} \\
\midrule
Latent & COCO & 49.0 & 45.5 & 61.8 & 53.6 & \cellcolor[HTML]{FCE4D5}{65.7} & \multicolumn{1}{>{\centering\arraybackslash}m{3.1em}!{\vrule width 1pt}}{\cellcolor[HTML]{EAECEF}{54.5}} & 2.3 & 7.4 & 25.3 & \multicolumn{1}{>{\centering\arraybackslash}m{3.1em}}{\cellcolor[HTML]{EAECEF}{8.0}} \\
\midrule
RL-Based & \makecell{\scriptsize{GRPO-SP}} & 10.0 & 5.8 & 5.2 & 8.2 & 12.7 & \multicolumn{1}{>{\centering\arraybackslash}m{3.1em}!{\vrule width 1pt}}{\cellcolor[HTML]{EAECEF}{8.2}} & 0.1 & 0.1 & 0.3 & \multicolumn{1}{>{\centering\arraybackslash}m{3.1em}}{\cellcolor[HTML]{EAECEF}{0.1}} \\
\midrule
Ours & \textbf{SoT} & \cellcolor[HTML]{FFD4D3}{\textbf{80.5}} & \cellcolor[HTML]{FFD4D3}{\textbf{75.0}} & \cellcolor[HTML]{FFD4D3}{\textbf{86.8}} & \cellcolor[HTML]{FFD4D3}{\textbf{80.0}} & \cellcolor[HTML]{FFD4D3}{\textbf{86.0}} & \multicolumn{1}{>{\centering\arraybackslash}m{3.1em}!{\vrule width 1pt}}{\cellcolor[HTML]{EAECEF}{\underline{\textbf{81.4}}}} & \cellcolor[HTML]{FCE4D5}{\textbf{24.2}} & \cellcolor[HTML]{FFFCE0}{\textbf{25.4}} & \cellcolor[HTML]{FFFCE0}{\textbf{27.3}} & \multicolumn{1}{>{\centering\arraybackslash}m{3.1em}}{\cellcolor[HTML]{EAECEF}{\textbf{25.2}}} \\
\bottomrule\end{tabularx}}%

\vspace{-2mm}
\end{table*}

\begin{table*}[!t]
\centering
\caption{\textbf{Main results on Mixtral-8x7B (accuracy, \%).}}
\label{tab:app_main_mixtral}
\vspace{-2mm}
\footnotesize
{\setlength{\tabcolsep}{1.3pt}%
\setlength{\aboverulesep}{0pt}%
\setlength{\belowrulesep}{0pt}%
\setlength{\extrarowheight}{1.43pt}%
\renewcommand{\arraystretch}{0.823}%
\renewcommand{\tabularxcolumn}[1]{m{#1}}%
\begin{tabularx}{\linewidth}{@{} >{\centering\arraybackslash}p{1.65cm} >{\centering\arraybackslash}m{1.28cm} !{\vrule width 0.4pt} >{\centering\arraybackslash}X >{\centering\arraybackslash}X >{\centering\arraybackslash}X >{\centering\arraybackslash}m{3.1em} >{\centering\arraybackslash}X >{\centering\arraybackslash}X >{\centering\arraybackslash}X >{\centering\arraybackslash}X >{\centering\arraybackslash}X >{\centering\arraybackslash}m{3.1em} @{}}
\toprule
\multirow{2}{=}{\centering\textbf{Category}} & \multirow{2}{=}{\centering\textbf{Method}}
& \multicolumn{4}{>{\cellcolor[HTML]{E3F2FD}}c!{\vrule width 1pt}}{\textbf{Quantitative Reasoning}}
& \multicolumn{6}{>{\cellcolor[HTML]{E3F2FD}}c}{\textbf{Symbolic and Code}} \\
&
& {\small\textbf{GSM}} & {\small\textbf{MATH}} & {\small\textbf{DROP}} & \multicolumn{1}{>{\centering\arraybackslash}m{3.1em}!{\vrule width 1pt}}{\small\textbf{avg.}}
& {\small\textbf{FOL}} & {\small\textbf{PW}} & {\small\textbf{BBH}} & {\small\textbf{HE}} & {\small\textbf{MBPP}} & \multicolumn{1}{>{\centering\arraybackslash}m{3.1em}}{\small\textbf{avg.}} \\
\midrule
\multirow{6}{=}{\centering Reasoning\\Paradigm}
 & CoT & 64.8 & 41.5 & 9.9 & \multicolumn{1}{>{\centering\arraybackslash}m{3.1em}!{\vrule width 1pt}}{\cellcolor[HTML]{EAECEF}{43.3}} & 23.2 & 43.0 & 47.0 & \cellcolor[HTML]{FFFCE0}{20.0} & \cellcolor[HTML]{FFFCE0}{40.4} & \multicolumn{1}{>{\centering\arraybackslash}m{3.1em}}{\cellcolor[HTML]{EAECEF}{38.1}} \\
 & PS & 60.6 & 35.8 & \cellcolor[HTML]{FCE4D5}{10.6} & \multicolumn{1}{>{\centering\arraybackslash}m{3.1em}!{\vrule width 1pt}}{\cellcolor[HTML]{EAECEF}{39.8}} & 12.8 & 38.2 & \cellcolor[HTML]{FFFCE0}{52.0} & \cellcolor[HTML]{FCE4D5}{25.0} & \cellcolor[HTML]{FCE4D5}{42.4} & \multicolumn{1}{>{\centering\arraybackslash}m{3.1em}}{\cellcolor[HTML]{EAECEF}{35.1}} \\
 & SR & 49.2 & 37.8 & 9.7 & \multicolumn{1}{>{\centering\arraybackslash}m{3.1em}!{\vrule width 1pt}}{\cellcolor[HTML]{EAECEF}{35.5}} & 27.6 & 54.5 & 51.0 & 15.0 & 32.0 & \multicolumn{1}{>{\centering\arraybackslash}m{3.1em}}{\cellcolor[HTML]{EAECEF}{41.9}} \\
 & SC & \cellcolor[HTML]{FFD4D3}{69.8} & \cellcolor[HTML]{FCE4D5}{43.0} & 9.8 & \multicolumn{1}{>{\centering\arraybackslash}m{3.1em}!{\vrule width 1pt}}{\cellcolor[HTML]{EAECEF}{45.9}} & 25.6 & 45.2 & \cellcolor[HTML]{FFD4D3}{56.0} & \cellcolor[HTML]{FFFCE0}{20.0} & 37.2 & \multicolumn{1}{>{\centering\arraybackslash}m{3.1em}}{\cellcolor[HTML]{EAECEF}{39.7}} \\
 & CB & 50.0 & 30.2 & \cellcolor[HTML]{FFFCE0}{10.1} & \multicolumn{1}{>{\centering\arraybackslash}m{3.1em}!{\vrule width 1pt}}{\cellcolor[HTML]{EAECEF}{33.4}} & \cellcolor[HTML]{FCE4D5}{48.3} & \cellcolor[HTML]{FCE4D5}{60.5} & 39.0 & 10.0 & 34.4 & \multicolumn{1}{>{\centering\arraybackslash}m{3.1em}}{\cellcolor[HTML]{EAECEF}{48.0}} \\
 & MCTS & 45.6 & 26.8 & 8.0 & \multicolumn{1}{>{\centering\arraybackslash}m{3.1em}!{\vrule width 1pt}}{\cellcolor[HTML]{EAECEF}{29.9}} & \cellcolor[HTML]{FFD4D3}{50.7} & \cellcolor[HTML]{FFFCE0}{57.2} & 39.0 & \cellcolor[HTML]{FFFCE0}{20.0} & 38.4 & \multicolumn{1}{>{\centering\arraybackslash}m{3.1em}}{\cellcolor[HTML]{EAECEF}{48.4}} \\
\midrule
\multirow{3}{=}{\centering Memory-Oriented}
 & H2O & \cellcolor[HTML]{FFFCE0}{65.2} & \cellcolor[HTML]{FCE4D5}{43.0} & 9.4 & \multicolumn{1}{>{\centering\arraybackslash}m{3.1em}!{\vrule width 1pt}}{\cellcolor[HTML]{EAECEF}{43.9}} & 23.6 & 43.8 & \cellcolor[HTML]{FFFCE0}{52.0} & 15.0 & 31.2 & \multicolumn{1}{>{\centering\arraybackslash}m{3.1em}}{\cellcolor[HTML]{EAECEF}{36.6}} \\
 & SNAP & 64.8 & \cellcolor[HTML]{FFD4D3}{43.2} & 9.1 & \multicolumn{1}{>{\centering\arraybackslash}m{3.1em}!{\vrule width 1pt}}{\cellcolor[HTML]{EAECEF}{43.7}} & 23.6 & 43.8 & \cellcolor[HTML]{FCE4D5}{55.0} & 15.0 & 30.8 & \multicolumn{1}{>{\centering\arraybackslash}m{3.1em}}{\cellcolor[HTML]{EAECEF}{36.8}} \\
 & STREAM & 59.0 & 38.5 & 1.1 & \multicolumn{1}{>{\centering\arraybackslash}m{3.1em}!{\vrule width 1pt}}{\cellcolor[HTML]{EAECEF}{37.7}} & 8.9 & 39.0 & 0.0 & 0.0 & 2.8 & \multicolumn{1}{>{\centering\arraybackslash}m{3.1em}}{\cellcolor[HTML]{EAECEF}{18.6}} \\
\midrule
Latent & COCO & 60.8 & 38.0 & 9.8 & \multicolumn{1}{>{\centering\arraybackslash}m{3.1em}!{\vrule width 1pt}}{\cellcolor[HTML]{EAECEF}{40.5}} & 29.1 & 29.0 & \cellcolor[HTML]{FCE4D5}{55.0} & 5.0 & 18.0 & \multicolumn{1}{>{\centering\arraybackslash}m{3.1em}}{\cellcolor[HTML]{EAECEF}{28.4}} \\
\midrule
RL-Based & \makecell{\scriptsize{GRPO-SP}} & 1.2 & 3.8 & 0.0 & \multicolumn{1}{>{\centering\arraybackslash}m{3.1em}!{\vrule width 1pt}}{\cellcolor[HTML]{EAECEF}{1.8}} & 4.4 & 3.8 & 8.0 & 0.0 & 0.0 & \multicolumn{1}{>{\centering\arraybackslash}m{3.1em}}{\cellcolor[HTML]{EAECEF}{3.3}} \\
\midrule
Ours & \textbf{SoT} & \cellcolor[HTML]{FCE4D5}{\textbf{69.2}} & \cellcolor[HTML]{FFFCE0}{\textbf{42.0}} & \cellcolor[HTML]{FFD4D3}{\textbf{14.0}} & \multicolumn{1}{>{\centering\arraybackslash}m{3.1em}!{\vrule width 1pt}}{\cellcolor[HTML]{EAECEF}{\textbf{46.3}}} & \cellcolor[HTML]{FFFCE0}{\textbf{43.8}} & \cellcolor[HTML]{FFD4D3}{\textbf{67.5}} & \cellcolor[HTML]{FCE4D5}{\textbf{55.0}} & \cellcolor[HTML]{FFD4D3}{\textbf{33.5}} & \cellcolor[HTML]{FFD4D3}{\textbf{43.2}} & \multicolumn{1}{>{\centering\arraybackslash}m{3.1em}}{\cellcolor[HTML]{EAECEF}{\underline{\textbf{51.7}}}} \\
\midrule\end{tabularx}\par\addvspace{0.35ex}}%
{\setlength{\tabcolsep}{1.3pt}%
\setlength{\aboverulesep}{0pt}%
\setlength{\belowrulesep}{0pt}%
\setlength{\extrarowheight}{1.43pt}%
\renewcommand{\arraystretch}{0.823}%
\renewcommand{\tabularxcolumn}[1]{m{#1}}%
\begin{tabularx}{\linewidth}{@{} >{\centering\arraybackslash}p{1.65cm} >{\centering\arraybackslash}m{1.28cm} !{\vrule width 0.4pt} >{\centering\arraybackslash}X >{\centering\arraybackslash}X >{\centering\arraybackslash}X >{\centering\arraybackslash}X >{\centering\arraybackslash}X >{\centering\arraybackslash}m{3.1em} >{\centering\arraybackslash}X >{\centering\arraybackslash}X >{\centering\arraybackslash}X >{\centering\arraybackslash}m{3.1em} @{}}
\multirow{2}{=}{\centering\textbf{Category}} & \multirow{2}{=}{\centering\textbf{Method}}
& \multicolumn{6}{>{\cellcolor[HTML]{E3F2FD}}c!{\vrule width 1pt}}{\textbf{General Understanding}}
& \multicolumn{4}{>{\cellcolor[HTML]{E3F2FD}}c}{\textbf{Long-Context Reasoning}} \\
&
& {\small\textbf{CSQA}} & {\small\textbf{StrQA}} & {\small\textbf{BoolQ}} & {\small\textbf{MMLU}} & {\small\textbf{RACE}} & \multicolumn{1}{>{\centering\arraybackslash}m{3.1em}!{\vrule width 1pt}}{\small\textbf{avg.}}
& {\small\textbf{HQA}} & {\small\textbf{NarQA}} & {\small\textbf{LB}} & \multicolumn{1}{>{\centering\arraybackslash}m{3.1em}}{\small\textbf{avg.}} \\
\midrule
\multirow{6}{=}{\centering Reasoning\\Paradigm}
 & CoT & 46.5 & 53.5 & 72.8 & 49.8 & 47.7 & \multicolumn{1}{>{\centering\arraybackslash}m{3.1em}!{\vrule width 1pt}}{\cellcolor[HTML]{EAECEF}{54.1}} & \cellcolor[HTML]{FFFCE0}{14.5} & 15.3 & \cellcolor[HTML]{FFFCE0}{37.8} & \multicolumn{1}{>{\centering\arraybackslash}m{3.1em}}{\cellcolor[HTML]{EAECEF}{18.7}} \\
 & PS & 51.5 & 38.5 & 46.8 & 49.6 & 47.3 & \multicolumn{1}{>{\centering\arraybackslash}m{3.1em}!{\vrule width 1pt}}{\cellcolor[HTML]{EAECEF}{46.9}} & 11.1 & \cellcolor[HTML]{FFFCE0}{17.4} & 33.2 & \multicolumn{1}{>{\centering\arraybackslash}m{3.1em}}{\cellcolor[HTML]{EAECEF}{17.2}} \\
 & SR & \cellcolor[HTML]{FFFCE0}{65.5} & 62.0 & 80.2 & \cellcolor[HTML]{FCE4D5}{63.2} & \cellcolor[HTML]{FCE4D5}{61.0} & \multicolumn{1}{>{\centering\arraybackslash}m{3.1em}!{\vrule width 1pt}}{\cellcolor[HTML]{EAECEF}{66.5}} & 9.9 & 12.2 & 33.0 & \multicolumn{1}{>{\centering\arraybackslash}m{3.1em}}{\cellcolor[HTML]{EAECEF}{14.6}} \\
 & SC & 44.5 & 57.2 & 75.8 & 55.0 & 50.0 & \multicolumn{1}{>{\centering\arraybackslash}m{3.1em}!{\vrule width 1pt}}{\cellcolor[HTML]{EAECEF}{56.8}} & \cellcolor[HTML]{FFFCE0}{14.5} & 15.3 & \cellcolor[HTML]{FCE4D5}{38.1} & \multicolumn{1}{>{\centering\arraybackslash}m{3.1em}}{\cellcolor[HTML]{EAECEF}{18.8}} \\
 & CB & \cellcolor[HTML]{FCE4D5}{67.2} & \cellcolor[HTML]{FFD4D3}{69.5} & \cellcolor[HTML]{FFD4D3}{82.0} & \cellcolor[HTML]{FFFCE0}{58.6} & 49.3 & \multicolumn{1}{>{\centering\arraybackslash}m{3.1em}!{\vrule width 1pt}}{\cellcolor[HTML]{EAECEF}{65.8}} & 13.0 & 16.6 & 37.4 & \multicolumn{1}{>{\centering\arraybackslash}m{3.1em}}{\cellcolor[HTML]{EAECEF}{18.4}} \\
 & MCTS & 60.0 & \cellcolor[HTML]{FCE4D5}{68.2} & \cellcolor[HTML]{FFFCE0}{81.0} & 53.6 & \cellcolor[HTML]{FFFCE0}{54.0} & \multicolumn{1}{>{\centering\arraybackslash}m{3.1em}!{\vrule width 1pt}}{\cellcolor[HTML]{EAECEF}{63.4}} & 9.5 & 17.3 & 36.4 & \multicolumn{1}{>{\centering\arraybackslash}m{3.1em}}{\cellcolor[HTML]{EAECEF}{16.9}} \\
\midrule
\multirow{3}{=}{\centering Memory-Oriented}
 & H2O & 45.8 & 51.7 & 73.5 & 52.0 & 11.7 & \multicolumn{1}{>{\centering\arraybackslash}m{3.1em}!{\vrule width 1pt}}{\cellcolor[HTML]{EAECEF}{49.0}} & 13.1 & 6.6 & 0.0 & \multicolumn{1}{>{\centering\arraybackslash}m{3.1em}}{\cellcolor[HTML]{EAECEF}{8.5}} \\
 & SNAP & 45.8 & 51.7 & 73.8 & 52.4 & 12.0 & \multicolumn{1}{>{\centering\arraybackslash}m{3.1em}!{\vrule width 1pt}}{\cellcolor[HTML]{EAECEF}{49.1}} & 13.1 & 6.7 & 0.0 & \multicolumn{1}{>{\centering\arraybackslash}m{3.1em}}{\cellcolor[HTML]{EAECEF}{8.5}} \\
 & STREAM & 47.0 & 50.2 & 30.2 & 39.4 & 0.0 & \multicolumn{1}{>{\centering\arraybackslash}m{3.1em}!{\vrule width 1pt}}{\cellcolor[HTML]{EAECEF}{35.4}} & \cellcolor[HTML]{FCE4D5}{15.2} & 0.0 & 0.0 & \multicolumn{1}{>{\centering\arraybackslash}m{3.1em}}{\cellcolor[HTML]{EAECEF}{6.9}} \\
\midrule
Latent & COCO & 48.0 & 41.2 & 57.5 & 48.0 & 48.3 & \multicolumn{1}{>{\centering\arraybackslash}m{3.1em}!{\vrule width 1pt}}{\cellcolor[HTML]{EAECEF}{48.6}} & \cellcolor[HTML]{FCE4D5}{15.2} & \cellcolor[HTML]{FCE4D5}{20.4} & 17.4 & \multicolumn{1}{>{\centering\arraybackslash}m{3.1em}}{\cellcolor[HTML]{EAECEF}{17.5}} \\
\midrule
RL-Based & \makecell{\scriptsize{GRPO-SP}} & 11.8 & 8.2 & 7.2 & 11.6 & 13.3 & \multicolumn{1}{>{\centering\arraybackslash}m{3.1em}!{\vrule width 1pt}}{\cellcolor[HTML]{EAECEF}{10.3}} & 0.1 & 0.2 & 0.6 & \multicolumn{1}{>{\centering\arraybackslash}m{3.1em}}{\cellcolor[HTML]{EAECEF}{0.2}} \\
\midrule
Ours & \textbf{SoT} & \cellcolor[HTML]{FFD4D3}{\textbf{71.0}} & \cellcolor[HTML]{FFFCE0}{\textbf{67.8}} & \cellcolor[HTML]{FCE4D5}{\textbf{81.5}} & \cellcolor[HTML]{FFD4D3}{\textbf{68.4}} & \cellcolor[HTML]{FFD4D3}{\textbf{76.0}} & \multicolumn{1}{>{\centering\arraybackslash}m{3.1em}!{\vrule width 1pt}}{\cellcolor[HTML]{EAECEF}{\underline{\textbf{72.6}}}} & \cellcolor[HTML]{FFD4D3}{\textbf{17.2}} & \cellcolor[HTML]{FFD4D3}{\textbf{22.2}} & \cellcolor[HTML]{FFD4D3}{\textbf{43.7}} & \multicolumn{1}{>{\centering\arraybackslash}m{3.1em}}{\cellcolor[HTML]{EAECEF}{\underline{\textbf{23.5}}}} \\
\bottomrule\end{tabularx}}%

\vspace{-2mm}
\end{table*}

\subsection{Extended efficiency and trade-off results}

Here we provide full per-dataset efficiency breakdowns for Llama-3.1-8B (Tables~\ref{tab:app_tok_llama}, \ref{tab:app_lat_llama}), Qwen2.5-14B (Tables~\ref{tab:app_tok_qwen}, \ref{tab:app_lat_qwen}), and Mixtral-8x7B (Tables~\ref{tab:app_tok_mixtral}, \ref{tab:app_lat_mixtral}).
Two consistent trends emerge: (i) search-heavy methods incur very large token/latency overhead, especially on long-context tasks (e.g., Qwen LCR token average \(4901.5\) for SC vs.\ \(254.0\) for SoT; Mixtral LCR token average \(7546.2\) for CB vs.\ \(260.6\) for SoT), and (ii) SoT remains in a low-cost regime across backbones, with particularly strong Llama efficiency in GU (\({74.9}\) tokens, \(1.8\)s) and QS (\({255.7}\) tokens, \(5.8\)s).
This indicates that SoT improves the accuracy--efficiency frontier through selective state-conditioned computation rather than brute-force trajectory expansion, so its efficiency gains reflect better control allocation, not merely shorter outputs by default.

\begin{table*}[!b]
\centering
\caption{\textbf{Generated reasoning tokens on Llama-3.1-8B.}}
\vspace{-2mm}
\label{tab:app_tok_llama}
\footnotesize
{\setlength{\tabcolsep}{1.3pt}%
\setlength{\aboverulesep}{0pt}%
\setlength{\belowrulesep}{0pt}%
\setlength{\extrarowheight}{1.43pt}%
\renewcommand{\arraystretch}{0.823}%
\renewcommand{\tabularxcolumn}[1]{m{#1}}%
\begin{tabularx}{\linewidth}{@{} >{\centering\arraybackslash}p{1.65cm} >{\centering\arraybackslash}m{1.28cm} !{\vrule width 0.4pt} >{\centering\arraybackslash}X >{\centering\arraybackslash}X >{\centering\arraybackslash}X >{\centering\arraybackslash}m{3.1em} >{\centering\arraybackslash}X >{\centering\arraybackslash}X >{\centering\arraybackslash}X >{\centering\arraybackslash}X >{\centering\arraybackslash}X >{\centering\arraybackslash}m{3.1em} @{}}
\toprule
\multirow{2}{=}{\centering\textbf{Category}} & \multirow{2}{=}{\centering\textbf{Method}}
& \multicolumn{4}{>{\cellcolor[HTML]{E3F2FD}}c!{\vrule width 1pt}}{\textbf{Quantitative Reasoning}}
& \multicolumn{6}{>{\cellcolor[HTML]{E3F2FD}}c}{\textbf{Symbolic and Code}} \\
&
& {\small\textbf{GSM}} & {\small\textbf{MATH}} & {\small\textbf{DROP}} & \multicolumn{1}{>{\centering\arraybackslash}m{3.1em}!{\vrule width 1pt}}{\small\textbf{avg.}}
& {\small\textbf{FOL}} & {\small\textbf{PW}} & {\small\textbf{BBH}} & {\small\textbf{HE}} & {\small\textbf{MBPP}} & \multicolumn{1}{>{\centering\arraybackslash}m{3.1em}}{\small\textbf{avg.}} \\
\midrule
\multirow{6}{=}{\centering Reasoning\\Paradigm}
 & CoT & 230.1 & 381.5 & 194.1 & \multicolumn{1}{>{\centering\arraybackslash}m{3.1em}!{\vrule width 1pt}}{\cellcolor[HTML]{EAECEF}{271.6}} & 342.9 & 332.0 & 264.4 & 490.1 & 415.4 & \multicolumn{1}{>{\centering\arraybackslash}m{3.1em}}{\cellcolor[HTML]{EAECEF}{369.8}} \\
 & PS & 311.0 & 358.9 & 259.9 & \multicolumn{1}{>{\centering\arraybackslash}m{3.1em}!{\vrule width 1pt}}{\cellcolor[HTML]{EAECEF}{314.2}} & 369.9 & 354.4 & 346.0 & 404.6 & 391.4 & \multicolumn{1}{>{\centering\arraybackslash}m{3.1em}}{\cellcolor[HTML]{EAECEF}{372.1}} \\
 & SR & 346.5 & 441.6 & 307.6 & \multicolumn{1}{>{\centering\arraybackslash}m{3.1em}!{\vrule width 1pt}}{\cellcolor[HTML]{EAECEF}{368.5}} & 337.8 & 330.4 & 336.4 & 517.6 & 473.0 & \multicolumn{1}{>{\centering\arraybackslash}m{3.1em}}{\cellcolor[HTML]{EAECEF}{391.7}} \\
\end{tabularx}\par\addvspace{0.35ex}}%
{\setlength{\tabcolsep}{1.3pt}%
\setlength{\aboverulesep}{0pt}%
\setlength{\belowrulesep}{0pt}%
\setlength{\extrarowheight}{1.43pt}%
\renewcommand{\arraystretch}{0.823}%
\renewcommand{\tabularxcolumn}[1]{m{#1}}%
\begin{tabularx}{\linewidth}{@{} >{\centering\arraybackslash}p{1.65cm} >{\centering\arraybackslash}m{1.28cm} !{\vrule width 0.4pt} >{\centering\arraybackslash}X >{\centering\arraybackslash}X >{\centering\arraybackslash}X >{\centering\arraybackslash}X >{\centering\arraybackslash}X >{\centering\arraybackslash}m{3.1em} >{\centering\arraybackslash}X >{\centering\arraybackslash}X >{\centering\arraybackslash}X >{\centering\arraybackslash}m{3.1em} @{}}
\multirow{2}{=}{\centering\textbf{Category}} & \multirow{2}{=}{\centering\textbf{Method}}
& \multicolumn{6}{>{\cellcolor[HTML]{E3F2FD}}c!{\vrule width 1pt}}{\textbf{General Understanding}}
& \multicolumn{4}{>{\cellcolor[HTML]{E3F2FD}}c}{\textbf{Long-Context Reasoning}} \\
&
& {\small\textbf{CSQA}} & {\small\textbf{StrQA}} & {\small\textbf{BoolQ}} & {\small\textbf{MMLU}} & {\small\textbf{RACE}} & \multicolumn{1}{>{\centering\arraybackslash}m{3.1em}!{\vrule width 1pt}}{\small\textbf{avg.}}
& {\small\textbf{HQA}} & {\small\textbf{NarQA}} & {\small\textbf{LB}} & \multicolumn{1}{>{\centering\arraybackslash}m{3.1em}}{\small\textbf{avg.}} \\
\midrule
\multirow{6}{=}{\centering Reasoning\\Paradigm}
 & CoT & 194.7 & 221.7 & 164.8 & 279.4 & 178.8 & \multicolumn{1}{>{\centering\arraybackslash}m{3.1em}!{\vrule width 1pt}}{\cellcolor[HTML]{EAECEF}{212.9}} & 275.4 & 165.9 & 60.7 & \multicolumn{1}{>{\centering\arraybackslash}m{3.1em}}{\cellcolor[HTML]{EAECEF}{198.1}} \\
 & PS & 283.7 & 344.4 & 310.1 & 353.0 & 292.8 & \multicolumn{1}{>{\centering\arraybackslash}m{3.1em}!{\vrule width 1pt}}{\cellcolor[HTML]{EAECEF}{319.8}} & 316.1 & 310.6 & 125.8 & \multicolumn{1}{>{\centering\arraybackslash}m{3.1em}}{\cellcolor[HTML]{EAECEF}{282.3}} \\
 & SR & 278.3 & 304.0 & 260.9 & 326.8 & 278.8 & \multicolumn{1}{>{\centering\arraybackslash}m{3.1em}!{\vrule width 1pt}}{\cellcolor[HTML]{EAECEF}{292.2}} & 326.9 & 284.1 & 135.3 & \multicolumn{1}{>{\centering\arraybackslash}m{3.1em}}{\cellcolor[HTML]{EAECEF}{278.8}} \\
\bottomrule\end{tabularx}}%

\vspace{-3mm}
\end{table*}

\begin{table*}[t]
\centering
\vspace{-4mm}
\caption{\textbf{End-to-end latency on Llama-3.1-8B.}}
\vspace{-2mm}
\label{tab:app_lat_llama}
\footnotesize
{\setlength{\tabcolsep}{1.3pt}%
\setlength{\aboverulesep}{0pt}%
\setlength{\belowrulesep}{0pt}%
\setlength{\extrarowheight}{1.43pt}%
\renewcommand{\arraystretch}{0.823}%
\renewcommand{\tabularxcolumn}[1]{m{#1}}%
\begin{tabularx}{\linewidth}{@{} >{\centering\arraybackslash}p{1.65cm} >{\centering\arraybackslash}m{1.28cm} !{\vrule width 0.4pt} >{\centering\arraybackslash}X >{\centering\arraybackslash}X >{\centering\arraybackslash}X >{\centering\arraybackslash}m{3.1em} >{\centering\arraybackslash}X >{\centering\arraybackslash}X >{\centering\arraybackslash}X >{\centering\arraybackslash}X >{\centering\arraybackslash}X >{\centering\arraybackslash}m{3.1em} @{}}
\toprule
\multirow{2}{=}{\centering\textbf{Category}} & \multirow{2}{=}{\centering\textbf{Method}}
& \multicolumn{4}{>{\cellcolor[HTML]{E3F2FD}}c!{\vrule width 1pt}}{\textbf{Quantitative Reasoning}}
& \multicolumn{6}{>{\cellcolor[HTML]{E3F2FD}}c}{\textbf{Symbolic and Code}} \\
&
& {\small\textbf{GSM}} & {\small\textbf{MATH}} & {\small\textbf{DROP}} & \multicolumn{1}{>{\centering\arraybackslash}m{3.1em}!{\vrule width 1pt}}{\small\textbf{avg.}}
& {\small\textbf{FOL}} & {\small\textbf{PW}} & {\small\textbf{BBH}} & {\small\textbf{HE}} & {\small\textbf{MBPP}} & \multicolumn{1}{>{\centering\arraybackslash}m{3.1em}}{\small\textbf{avg.}} \\
\midrule
\multirow{6}{=}{\centering Reasoning\\Paradigm}
 & CoT & 12.1 & 25.6 & 15.7 & \multicolumn{1}{>{\centering\arraybackslash}m{3.1em}!{\vrule width 1pt}}{\cellcolor[HTML]{EAECEF}{17.5}} & 24.0 & 23.9 & 18.9 & 41.3 & 30.4 & \multicolumn{1}{>{\centering\arraybackslash}m{3.1em}}{\cellcolor[HTML]{EAECEF}{27.5}} \\
 & PS & 18.7 & 22.1 & 22.3 & \multicolumn{1}{>{\centering\arraybackslash}m{3.1em}!{\vrule width 1pt}}{\cellcolor[HTML]{EAECEF}{20.7}} & 26.2 & 24.4 & 27.7 & 32.3 & 27.3 & \multicolumn{1}{>{\centering\arraybackslash}m{3.1em}}{\cellcolor[HTML]{EAECEF}{26.8}} \\
 & SR & 21.6 & 30.3 & 27.2 & \multicolumn{1}{>{\centering\arraybackslash}m{3.1em}!{\vrule width 1pt}}{\cellcolor[HTML]{EAECEF}{25.9}} & 22.6 & 21.6 & 26.2 & 45.1 & 35.5 & \multicolumn{1}{>{\centering\arraybackslash}m{3.1em}}{\cellcolor[HTML]{EAECEF}{28.8}} \\
\end{tabularx}\par\addvspace{0.35ex}}%
{\setlength{\tabcolsep}{1.3pt}%
\setlength{\aboverulesep}{0pt}%
\setlength{\belowrulesep}{0pt}%
\setlength{\extrarowheight}{1.43pt}%
\renewcommand{\arraystretch}{0.823}%
\renewcommand{\tabularxcolumn}[1]{m{#1}}%
\begin{tabularx}{\linewidth}{@{} >{\centering\arraybackslash}p{1.65cm} >{\centering\arraybackslash}m{1.28cm} !{\vrule width 0.4pt} >{\centering\arraybackslash}X >{\centering\arraybackslash}X >{\centering\arraybackslash}X >{\centering\arraybackslash}X >{\centering\arraybackslash}X >{\centering\arraybackslash}m{3.1em} >{\centering\arraybackslash}X >{\centering\arraybackslash}X >{\centering\arraybackslash}X >{\centering\arraybackslash}m{3.1em} @{}}
\multirow{2}{=}{\centering\textbf{Category}} & \multirow{2}{=}{\centering\textbf{Method}}
& \multicolumn{6}{>{\cellcolor[HTML]{E3F2FD}}c!{\vrule width 1pt}}{\textbf{General Understanding}}
& \multicolumn{4}{>{\cellcolor[HTML]{E3F2FD}}c}{\textbf{Long-Context Reasoning}} \\
&
& {\small\textbf{CSQA}} & {\small\textbf{StrQA}} & {\small\textbf{BoolQ}} & {\small\textbf{MMLU}} & {\small\textbf{RACE}} & \multicolumn{1}{>{\centering\arraybackslash}m{3.1em}!{\vrule width 1pt}}{\small\textbf{avg.}}
& {\small\textbf{HQA}} & {\small\textbf{NarQA}} & {\small\textbf{LB}} & \multicolumn{1}{>{\centering\arraybackslash}m{3.1em}}{\small\textbf{avg.}} \\
\midrule
\multirow{6}{=}{\centering Reasoning\\Paradigm}
 & CoT & 9.3 & \cellcolor[HTML]{FFFCE0}{9.8} & 9.6 & 16.7 & 19.1 & \multicolumn{1}{>{\centering\arraybackslash}m{3.1em}!{\vrule width 1pt}}{\cellcolor[HTML]{EAECEF}{12.8}} & 15.2 & \cellcolor[HTML]{FFFCE0}{18.5} & 33.8 & \multicolumn{1}{>{\centering\arraybackslash}m{3.1em}}{\cellcolor[HTML]{EAECEF}{19.6}} \\
 & PS & 15.8 & 18.7 & 21.8 & 22.9 & 34.8 & \multicolumn{1}{>{\centering\arraybackslash}m{3.1em}!{\vrule width 1pt}}{\cellcolor[HTML]{EAECEF}{22.2}} & 16.6 & 38.0 & 70.7 & \multicolumn{1}{>{\centering\arraybackslash}m{3.1em}}{\cellcolor[HTML]{EAECEF}{33.7}} \\
 & SR & 14.7 & 15.1 & 17.0 & 20.0 & 32.1 & \multicolumn{1}{>{\centering\arraybackslash}m{3.1em}!{\vrule width 1pt}}{\cellcolor[HTML]{EAECEF}{19.2}} & 17.0 & 33.7 & 76.3 & \multicolumn{1}{>{\centering\arraybackslash}m{3.1em}}{\cellcolor[HTML]{EAECEF}{33.2}} \\
\bottomrule\end{tabularx}}%

\vspace{-2mm}
\end{table*}

\begin{table*}[t]
\centering
\caption{\textbf{Generated reasoning tokens on Qwen2.5-14B.}}
\vspace{-2mm}
\label{tab:app_tok_qwen}
\footnotesize
{\setlength{\tabcolsep}{1.3pt}%
\setlength{\aboverulesep}{0pt}%
\setlength{\belowrulesep}{0pt}%
\setlength{\extrarowheight}{1.43pt}%
\renewcommand{\arraystretch}{0.823}%
\renewcommand{\tabularxcolumn}[1]{m{#1}}%
\begin{tabularx}{\linewidth}{@{} >{\centering\arraybackslash}p{1.65cm} >{\centering\arraybackslash}m{1.28cm} !{\vrule width 0.4pt} >{\centering\arraybackslash}X >{\centering\arraybackslash}X >{\centering\arraybackslash}X >{\centering\arraybackslash}m{3.1em} >{\centering\arraybackslash}X >{\centering\arraybackslash}X >{\centering\arraybackslash}X >{\centering\arraybackslash}X >{\centering\arraybackslash}X >{\centering\arraybackslash}m{3.1em} @{}}
\toprule
\multirow{2}{=}{\centering\textbf{Category}} & \multirow{2}{=}{\centering\textbf{Method}}
& \multicolumn{4}{>{\cellcolor[HTML]{E3F2FD}}c!{\vrule width 1pt}}{\textbf{Quantitative Reasoning}}
& \multicolumn{6}{>{\cellcolor[HTML]{E3F2FD}}c}{\textbf{Symbolic and Code}} \\
&
& {\small\textbf{GSM}} & {\small\textbf{MATH}} & {\small\textbf{DROP}} & \multicolumn{1}{>{\centering\arraybackslash}m{3.1em}!{\vrule width 1pt}}{\small\textbf{avg.}}
& {\small\textbf{FOL}} & {\small\textbf{PW}} & {\small\textbf{BBH}} & {\small\textbf{HE}} & {\small\textbf{MBPP}} & \multicolumn{1}{>{\centering\arraybackslash}m{3.1em}}{\small\textbf{avg.}} \\
\midrule
\multirow{6}{=}{\centering Reasoning\\Paradigm} & CoT & 399.4 & 521.3 & 544.5 & \multicolumn{1}{>{\centering\arraybackslash}m{3.1em}!{\vrule width 1pt}}{\cellcolor[HTML]{EAECEF}{476.3}} & 503.3 & 416.7 & 456.0 & 644.4 & 597.8 & \multicolumn{1}{>{\centering\arraybackslash}m{3.1em}}{\cellcolor[HTML]{EAECEF}{490.0}} \\
 & PS & 653.4 & 698.4 & 1021.3 & \multicolumn{1}{>{\centering\arraybackslash}m{3.1em}!{\vrule width 1pt}}{\cellcolor[HTML]{EAECEF}{760.4}} & 639.7 & 633.2 & 896.3 & 935.2 & 829.9 & \multicolumn{1}{>{\centering\arraybackslash}m{3.1em}}{\cellcolor[HTML]{EAECEF}{718.3}} \\
 & SR & 1326.7 & 1454.9 & 1804.9 & \multicolumn{1}{>{\centering\arraybackslash}m{3.1em}!{\vrule width 1pt}}{\cellcolor[HTML]{EAECEF}{1489.0}} & 1496.6 & 1528.6 & 1601.1 & 1891.5 & 1717.5 & \multicolumn{1}{>{\centering\arraybackslash}m{3.1em}}{\cellcolor[HTML]{EAECEF}{1625.0}} \\
\end{tabularx}\par\addvspace{0.35ex}}%
{\setlength{\tabcolsep}{1.3pt}%
\setlength{\aboverulesep}{0pt}%
\setlength{\belowrulesep}{0pt}%
\setlength{\extrarowheight}{1.43pt}%
\renewcommand{\arraystretch}{0.823}%
\renewcommand{\tabularxcolumn}[1]{m{#1}}%
\begin{tabularx}{\linewidth}{@{} >{\centering\arraybackslash}p{1.65cm} >{\centering\arraybackslash}m{1.28cm} !{\vrule width 0.4pt} >{\centering\arraybackslash}X >{\centering\arraybackslash}X >{\centering\arraybackslash}X >{\centering\arraybackslash}X >{\centering\arraybackslash}X >{\centering\arraybackslash}m{3.1em} >{\centering\arraybackslash}X >{\centering\arraybackslash}X >{\centering\arraybackslash}X >{\centering\arraybackslash}m{3.1em} @{}}
\multirow{2}{=}{\centering\textbf{Category}} & \multirow{2}{=}{\centering\textbf{Method}}
& \multicolumn{6}{>{\cellcolor[HTML]{E3F2FD}}c!{\vrule width 1pt}}{\textbf{General Understanding}}
& \multicolumn{4}{>{\cellcolor[HTML]{E3F2FD}}c}{\textbf{Long-Context Reasoning}} \\
&
& {\small\textbf{CSQA}} & {\small\textbf{StrQA}} & {\small\textbf{BoolQ}} & {\small\textbf{MMLU}} & {\small\textbf{RACE}} & \multicolumn{1}{>{\centering\arraybackslash}m{3.1em}!{\vrule width 1pt}}{\small\textbf{avg.}}
& {\small\textbf{HQA}} & {\small\textbf{NarQA}} & {\small\textbf{LB}} & \multicolumn{1}{>{\centering\arraybackslash}m{3.1em}}{\small\textbf{avg.}} \\
\midrule
\multirow{6}{=}{\centering Reasoning\\Paradigm} & CoT & 263.1 & 251.8 & 314.9 & 401.0 & 612.3 & \multicolumn{1}{>{\centering\arraybackslash}m{3.1em}!{\vrule width 1pt}}{\cellcolor[HTML]{EAECEF}{358.1}} & 322.1 & 729.9 & \cellcolor[HTML]{FCE4D5}{6597.5} & \multicolumn{1}{>{\centering\arraybackslash}m{3.1em}}{\cellcolor[HTML]{EAECEF}{529.6}} \\
 & PS & 490.2 & 472.6 & 686.0 & 660.5 & 638.2 & \multicolumn{1}{>{\centering\arraybackslash}m{3.1em}!{\vrule width 1pt}}{\cellcolor[HTML]{EAECEF}{590.6}} & 507.3 & 1386.6 & 14498.2 & \multicolumn{1}{>{\centering\arraybackslash}m{3.1em}}{\cellcolor[HTML]{EAECEF}{3172.2}} \\
 & SR & 985.7 & 938.4 & 1139.9 & 1361.6 & 702.1 & \multicolumn{1}{>{\centering\arraybackslash}m{3.1em}!{\vrule width 1pt}}{\cellcolor[HTML]{EAECEF}{1058.5}} & 1156.6 & 2363.8 & 21847.2 & \multicolumn{1}{>{\centering\arraybackslash}m{3.1em}}{\cellcolor[HTML]{EAECEF}{5062.3}} \\
\bottomrule\end{tabularx}}%

\vspace{-2mm}
\end{table*}

\begin{table*}[t]
\centering
\vspace{-4mm}
\caption{\textbf{End-to-end latency on Qwen2.5-14B.}}
\vspace{-2mm}
\label{tab:app_lat_qwen}
\footnotesize
{\setlength{\tabcolsep}{1.3pt}%
\setlength{\aboverulesep}{0pt}%
\setlength{\belowrulesep}{0pt}%
\setlength{\extrarowheight}{1.43pt}%
\renewcommand{\arraystretch}{0.823}%
\renewcommand{\tabularxcolumn}[1]{m{#1}}%
\begin{tabularx}{\linewidth}{@{} >{\centering\arraybackslash}p{1.65cm} >{\centering\arraybackslash}m{1.28cm} !{\vrule width 0.4pt} >{\centering\arraybackslash}X >{\centering\arraybackslash}X >{\centering\arraybackslash}X >{\centering\arraybackslash}m{3.1em} >{\centering\arraybackslash}X >{\centering\arraybackslash}X >{\centering\arraybackslash}X >{\centering\arraybackslash}X >{\centering\arraybackslash}X >{\centering\arraybackslash}m{3.1em} @{}}
\toprule
\multirow{2}{=}{\centering\textbf{Category}} & \multirow{2}{=}{\centering\textbf{Method}}
& \multicolumn{4}{>{\cellcolor[HTML]{E3F2FD}}c!{\vrule width 1pt}}{\textbf{Quantitative Reasoning}}
& \multicolumn{6}{>{\cellcolor[HTML]{E3F2FD}}c}{\textbf{Symbolic and Code}} \\
&
& {\small\textbf{GSM}} & {\small\textbf{MATH}} & {\small\textbf{DROP}} & \multicolumn{1}{>{\centering\arraybackslash}m{3.1em}!{\vrule width 1pt}}{\small\textbf{avg.}}
& {\small\textbf{FOL}} & {\small\textbf{PW}} & {\small\textbf{BBH}} & {\small\textbf{HE}} & {\small\textbf{MBPP}} & \multicolumn{1}{>{\centering\arraybackslash}m{3.1em}}{\small\textbf{avg.}} \\
\midrule
\multirow{6}{=}{\centering Reasoning\\Paradigm} & CoT & 25.0 & 47.4 & 26.6 & \multicolumn{1}{>{\centering\arraybackslash}m{3.1em}!{\vrule width 1pt}}{\cellcolor[HTML]{EAECEF}{32.9}} & 39.2 & 25.8 & 23.0 & 60.7 & 56.7 & \multicolumn{1}{>{\centering\arraybackslash}m{3.1em}}{\cellcolor[HTML]{EAECEF}{36.9}} \\
 & PS & 27.9 & 36.7 & 32.2 & \multicolumn{1}{>{\centering\arraybackslash}m{3.1em}!{\vrule width 1pt}}{\cellcolor[HTML]{EAECEF}{31.9}} & 22.4 & 19.6 & 38.0 & 50.1 & 46.1 & \multicolumn{1}{>{\centering\arraybackslash}m{3.1em}}{\cellcolor[HTML]{EAECEF}{29.5}} \\
 & SR & 37.8 & 56.3 & 43.5 & \multicolumn{1}{>{\centering\arraybackslash}m{3.1em}!{\vrule width 1pt}}{\cellcolor[HTML]{EAECEF}{45.4}} & 41.1 & 49.8 & 39.6 & 87.1 & 80.4 & \multicolumn{1}{>{\centering\arraybackslash}m{3.1em}}{\cellcolor[HTML]{EAECEF}{59.6}} \\
\end{tabularx}\par\addvspace{0.35ex}}%
{\setlength{\tabcolsep}{1.3pt}%
\setlength{\aboverulesep}{0pt}%
\setlength{\belowrulesep}{0pt}%
\setlength{\extrarowheight}{1.43pt}%
\renewcommand{\arraystretch}{0.823}%
\renewcommand{\tabularxcolumn}[1]{m{#1}}%
\begin{tabularx}{\linewidth}{@{} >{\centering\arraybackslash}p{1.65cm} >{\centering\arraybackslash}m{1.28cm} !{\vrule width 0.4pt} >{\centering\arraybackslash}X >{\centering\arraybackslash}X >{\centering\arraybackslash}X >{\centering\arraybackslash}X >{\centering\arraybackslash}X >{\centering\arraybackslash}m{3.1em} >{\centering\arraybackslash}X >{\centering\arraybackslash}X >{\centering\arraybackslash}X >{\centering\arraybackslash}m{3.1em} @{}}
\multirow{2}{=}{\centering\textbf{Category}} & \multirow{2}{=}{\centering\textbf{Method}}
& \multicolumn{6}{>{\cellcolor[HTML]{E3F2FD}}c!{\vrule width 1pt}}{\textbf{General Understanding}}
& \multicolumn{4}{>{\cellcolor[HTML]{E3F2FD}}c}{\textbf{Long-Context Reasoning}} \\
&
& {\small\textbf{CSQA}} & {\small\textbf{StrQA}} & {\small\textbf{BoolQ}} & {\small\textbf{MMLU}} & {\small\textbf{RACE}} & \multicolumn{1}{>{\centering\arraybackslash}m{3.1em}!{\vrule width 1pt}}{\small\textbf{avg.}}
& {\small\textbf{HQA}} & {\small\textbf{NarQA}} & {\small\textbf{LB}} & \multicolumn{1}{>{\centering\arraybackslash}m{3.1em}}{\small\textbf{avg.}} \\
\midrule
\multirow{6}{=}{\centering Reasoning\\Paradigm} & CoT & 10.1 & 11.4 & 9.2 & 25.3 & 21.8 & \multicolumn{1}{>{\centering\arraybackslash}m{3.1em}!{\vrule width 1pt}}{\cellcolor[HTML]{EAECEF}{15.7}} & 18.1 & 35.2 & 254.1 & \multicolumn{1}{>{\centering\arraybackslash}m{3.1em}}{\cellcolor[HTML]{EAECEF}{26.7}} \\
 & PS & 11.9 & 13.2 & 16.8 & 25.1 & 23.6 & \multicolumn{1}{>{\centering\arraybackslash}m{3.1em}!{\vrule width 1pt}}{\cellcolor[HTML]{EAECEF}{18.2}} & 20.1 & 38.3 & 262.8 & \multicolumn{1}{>{\centering\arraybackslash}m{3.1em}}{\cellcolor[HTML]{EAECEF}{67.5}} \\
 & SR & 18.9 & 20.6 & 18.9 & 35.4 & 30.2 & \multicolumn{1}{>{\centering\arraybackslash}m{3.1em}!{\vrule width 1pt}}{\cellcolor[HTML]{EAECEF}{25.1}} & 32.0 & 57.2 & 366.3 & \multicolumn{1}{>{\centering\arraybackslash}m{3.1em}}{\cellcolor[HTML]{EAECEF}{97.2}} \\
\bottomrule\end{tabularx}}%

\vspace{-2mm}
\end{table*}

\begin{table*}[t]
\centering
\caption{\textbf{Generated reasoning tokens on Mixtral-8x7B.}}
\vspace{-2mm}
\label{tab:app_tok_mixtral}
\footnotesize
{\setlength{\tabcolsep}{1.3pt}%
\setlength{\aboverulesep}{0pt}%
\setlength{\belowrulesep}{0pt}%
\setlength{\extrarowheight}{1.43pt}%
\renewcommand{\arraystretch}{0.823}%
\renewcommand{\tabularxcolumn}[1]{m{#1}}%
\begin{tabularx}{\linewidth}{@{} >{\centering\arraybackslash}p{1.65cm} >{\centering\arraybackslash}m{1.28cm} !{\vrule width 0.4pt} >{\centering\arraybackslash}X >{\centering\arraybackslash}X >{\centering\arraybackslash}X >{\centering\arraybackslash}m{3.1em} >{\centering\arraybackslash}X >{\centering\arraybackslash}X >{\centering\arraybackslash}X >{\centering\arraybackslash}X >{\centering\arraybackslash}X >{\centering\arraybackslash}m{3.1em} @{}}
\toprule
\multirow{2}{=}{\centering\textbf{Category}} & \multirow{2}{=}{\centering\textbf{Method}}
& \multicolumn{4}{>{\cellcolor[HTML]{E3F2FD}}c!{\vrule width 1pt}}{\textbf{Quantitative Reasoning}}
& \multicolumn{6}{>{\cellcolor[HTML]{E3F2FD}}c}{\textbf{Symbolic and Code}} \\
&
& {\small\textbf{GSM}} & {\small\textbf{MATH}} & {\small\textbf{DROP}} & \multicolumn{1}{>{\centering\arraybackslash}m{3.1em}!{\vrule width 1pt}}{\small\textbf{avg.}}
& {\small\textbf{FOL}} & {\small\textbf{PW}} & {\small\textbf{BBH}} & {\small\textbf{HE}} & {\small\textbf{MBPP}} & \multicolumn{1}{>{\centering\arraybackslash}m{3.1em}}{\small\textbf{avg.}} \\
\midrule
\multirow{6}{=}{\centering Reasoning\\Paradigm}
 & CoT & 348.8 & 439.0 & 534.7 & \multicolumn{1}{>{\centering\arraybackslash}m{3.1em}!{\vrule width 1pt}}{\cellcolor[HTML]{EAECEF}{425.3}} & 394.8 & 353.3 & 471.7 & 617.2 & 574.4 & \multicolumn{1}{>{\centering\arraybackslash}m{3.1em}}{\cellcolor[HTML]{EAECEF}{436.3}} \\
 & PS & 665.9 & 703.1 & 1094.4 & \multicolumn{1}{>{\centering\arraybackslash}m{3.1em}!{\vrule width 1pt}}{\cellcolor[HTML]{EAECEF}{785.4}} & 801.5 & 728.8 & 944.9 & 1000.4 & 868.1 & \multicolumn{1}{>{\centering\arraybackslash}m{3.1em}}{\cellcolor[HTML]{EAECEF}{807.6}} \\
 & SR & 1222.4 & 1336.4 & 1779.2 & \multicolumn{1}{>{\centering\arraybackslash}m{3.1em}!{\vrule width 1pt}}{\cellcolor[HTML]{EAECEF}{1399.6}} & 1403.3 & 1283.2 & 1782.4 & 1970.0 & 1709.8 & \multicolumn{1}{>{\centering\arraybackslash}m{3.1em}}{\cellcolor[HTML]{EAECEF}{1483.3}} \\
\end{tabularx}\par\addvspace{0.35ex}}%
{\setlength{\tabcolsep}{1.3pt}%
\setlength{\aboverulesep}{0pt}%
\setlength{\belowrulesep}{0pt}%
\setlength{\extrarowheight}{1.43pt}%
\renewcommand{\arraystretch}{0.823}%
\renewcommand{\tabularxcolumn}[1]{m{#1}}%
\begin{tabularx}{\linewidth}{@{} >{\centering\arraybackslash}p{1.65cm} >{\centering\arraybackslash}m{1.28cm} !{\vrule width 0.4pt} >{\centering\arraybackslash}X >{\centering\arraybackslash}X >{\centering\arraybackslash}X >{\centering\arraybackslash}X >{\centering\arraybackslash}X >{\centering\arraybackslash}m{3.1em} >{\centering\arraybackslash}X >{\centering\arraybackslash}X >{\centering\arraybackslash}X >{\centering\arraybackslash}m{3.1em} @{}}
\multirow{2}{=}{\centering\textbf{Category}} & \multirow{2}{=}{\centering\textbf{Method}}
& \multicolumn{6}{>{\cellcolor[HTML]{E3F2FD}}c!{\vrule width 1pt}}{\textbf{General Understanding}}
& \multicolumn{4}{>{\cellcolor[HTML]{E3F2FD}}c}{\textbf{Long-Context Reasoning}} \\
&
& {\small\textbf{CSQA}} & {\small\textbf{StrQA}} & {\small\textbf{BoolQ}} & {\small\textbf{MMLU}} & {\small\textbf{RACE}} & \multicolumn{1}{>{\centering\arraybackslash}m{3.1em}!{\vrule width 1pt}}{\small\textbf{avg.}}
& {\small\textbf{HQA}} & {\small\textbf{NarQA}} & {\small\textbf{LB}} & \multicolumn{1}{>{\centering\arraybackslash}m{3.1em}}{\small\textbf{avg.}} \\
\midrule
\multirow{6}{=}{\centering Reasoning\\Paradigm}
 & CoT & 249.9 & 210.3 & 303.7 & 328.0 & 678.5 & \multicolumn{1}{>{\centering\arraybackslash}m{3.1em}!{\vrule width 1pt}}{\cellcolor[HTML]{EAECEF}{336.6}} & 251.4 & 726.3 & 7933.4 & \multicolumn{1}{>{\centering\arraybackslash}m{3.1em}}{\cellcolor[HTML]{EAECEF}{1711.6}} \\
 & PS & 564.3 & 520.0 & 776.7 & 698.3 & 1504.5 & \multicolumn{1}{>{\centering\arraybackslash}m{3.1em}!{\vrule width 1pt}}{\cellcolor[HTML]{EAECEF}{772.5}} & 504.2 & 1532.9 & 16024.9 & \multicolumn{1}{>{\centering\arraybackslash}m{3.1em}}{\cellcolor[HTML]{EAECEF}{3480.7}} \\
 & SR & 1028.7 & 930.4 & 1183.0 & 1336.8 & 2407.1 & \multicolumn{1}{>{\centering\arraybackslash}m{3.1em}!{\vrule width 1pt}}{\cellcolor[HTML]{EAECEF}{1323.7}} & 1021.2 & 2314.3 & 24064.0 & \multicolumn{1}{>{\centering\arraybackslash}m{3.1em}}{\cellcolor[HTML]{EAECEF}{5351.5}} \\
\bottomrule\end{tabularx}}%

\vspace{-2mm}
\end{table*}

\begin{table*}[t]
\centering
\vspace{-4mm}
\caption{\textbf{End-to-end latency on Mixtral-8x7B.}}
\vspace{-2mm}
\label{tab:app_lat_mixtral}
\footnotesize
{\setlength{\tabcolsep}{1.3pt}%
\setlength{\aboverulesep}{0pt}%
\setlength{\belowrulesep}{0pt}%
\setlength{\extrarowheight}{1.43pt}%
\renewcommand{\arraystretch}{0.823}%
\renewcommand{\tabularxcolumn}[1]{m{#1}}%
\begin{tabularx}{\linewidth}{@{} >{\centering\arraybackslash}p{1.65cm} >{\centering\arraybackslash}m{1.28cm} !{\vrule width 0.4pt} >{\centering\arraybackslash}X >{\centering\arraybackslash}X >{\centering\arraybackslash}X >{\centering\arraybackslash}m{3.1em} >{\centering\arraybackslash}X >{\centering\arraybackslash}X >{\centering\arraybackslash}X >{\centering\arraybackslash}X >{\centering\arraybackslash}X >{\centering\arraybackslash}m{3.1em} @{}}
\toprule
\multirow{2}{=}{\centering\textbf{Category}} & \multirow{2}{=}{\centering\textbf{Method}}
& \multicolumn{4}{>{\cellcolor[HTML]{E3F2FD}}c!{\vrule width 1pt}}{\textbf{Quantitative Reasoning}}
& \multicolumn{6}{>{\cellcolor[HTML]{E3F2FD}}c}{\textbf{Symbolic and Code}} \\
&
& {\small\textbf{GSM}} & {\small\textbf{MATH}} & {\small\textbf{DROP}} & \multicolumn{1}{>{\centering\arraybackslash}m{3.1em}!{\vrule width 1pt}}{\small\textbf{avg.}}
& {\small\textbf{FOL}} & {\small\textbf{PW}} & {\small\textbf{BBH}} & {\small\textbf{HE}} & {\small\textbf{MBPP}} & \multicolumn{1}{>{\centering\arraybackslash}m{3.1em}}{\small\textbf{avg.}} \\
\midrule
\multirow{6}{=}{\centering Reasoning\\Paradigm}
 & CoT & \cellcolor[HTML]{FFD4D3}{26.7} & \cellcolor[HTML]{FFD4D3}{40.5} & \cellcolor[HTML]{FFD4D3}{18.0} & \multicolumn{1}{>{\centering\arraybackslash}m{3.1em}!{\vrule width 1pt}}{\cellcolor[HTML]{EAECEF}{\underline{29.1}}} & \cellcolor[HTML]{FCE4D5}{28.6} & \cellcolor[HTML]{FFD4D3}{19.3} & \cellcolor[HTML]{FFD4D3}{27.0} & \cellcolor[HTML]{FCE4D5}{49.7} & \cellcolor[HTML]{FCE4D5}{46.7} & \multicolumn{1}{>{\centering\arraybackslash}m{3.1em}}{\cellcolor[HTML]{EAECEF}{\underline{29.7}}} \\
\end{tabularx}\par\addvspace{0.35ex}}%
{\setlength{\tabcolsep}{1.3pt}%
\setlength{\aboverulesep}{0pt}%
\setlength{\belowrulesep}{0pt}%
\setlength{\extrarowheight}{1.43pt}%
\renewcommand{\arraystretch}{0.823}%
\renewcommand{\tabularxcolumn}[1]{m{#1}}%
\begin{tabularx}{\linewidth}{@{} >{\centering\arraybackslash}p{1.65cm} >{\centering\arraybackslash}m{1.28cm} !{\vrule width 0.4pt} >{\centering\arraybackslash}X >{\centering\arraybackslash}X >{\centering\arraybackslash}X >{\centering\arraybackslash}X >{\centering\arraybackslash}X >{\centering\arraybackslash}m{3.1em} >{\centering\arraybackslash}X >{\centering\arraybackslash}X >{\centering\arraybackslash}X >{\centering\arraybackslash}m{3.1em} @{}}
\multirow{2}{=}{\centering\textbf{Category}} & \multirow{2}{=}{\centering\textbf{Method}}
& \multicolumn{6}{>{\cellcolor[HTML]{E3F2FD}}c!{\vrule width 1pt}}{\textbf{General Understanding}}
& \multicolumn{4}{>{\cellcolor[HTML]{E3F2FD}}c}{\textbf{Long-Context Reasoning}} \\
&
& {\small\textbf{CSQA}} & {\small\textbf{StrQA}} & {\small\textbf{BoolQ}} & {\small\textbf{MMLU}} & {\small\textbf{RACE}} & \multicolumn{1}{>{\centering\arraybackslash}m{3.1em}!{\vrule width 1pt}}{\small\textbf{avg.}}
& {\small\textbf{HQA}} & {\small\textbf{NarQA}} & {\small\textbf{LB}} & \multicolumn{1}{>{\centering\arraybackslash}m{3.1em}}{\small\textbf{avg.}} \\
\midrule
\multirow{6}{=}{\centering Reasoning\\Paradigm}
 & CoT & \cellcolor[HTML]{FFD4D3}{15.6} & \cellcolor[HTML]{FFD4D3}{15.1} & \cellcolor[HTML]{FFD4D3}{11.6} & \cellcolor[HTML]{FFD4D3}{21.3} & \cellcolor[HTML]{FCE4D5}{15.8} & \multicolumn{1}{>{\centering\arraybackslash}m{3.1em}!{\vrule width 1pt}}{\cellcolor[HTML]{EAECEF}{\underline{16.2}}} & \cellcolor[HTML]{FFD4D3}{19.3} & \cellcolor[HTML]{FFD4D3}{15.2} & 15.1 & \multicolumn{1}{>{\centering\arraybackslash}m{3.1em}}{\cellcolor[HTML]{EAECEF}{\underline{17.1}}} \\
 & PS & \cellcolor[HTML]{FCE4D5}{22.2} & 25.6 & 27.2 & \cellcolor[HTML]{FCE4D5}{30.7} & 34.3 & \multicolumn{1}{>{\centering\arraybackslash}m{3.1em}!{\vrule width 1pt}}{\cellcolor[HTML]{EAECEF}{27.8}} & \cellcolor[HTML]{FCE4D5}{22.5} & \cellcolor[HTML]{FCE4D5}{23.3} & 31.3 & \multicolumn{1}{>{\centering\arraybackslash}m{3.1em}}{\cellcolor[HTML]{EAECEF}{24.3}} \\
\bottomrule\end{tabularx}}%

\vspace{-2mm}
\end{table*}

\clearpage
\begin{figure*}[t]
\centering
\begin{subfigure}{0.49\textwidth}
    \centering
    \includegraphics[width=1.0\textwidth]{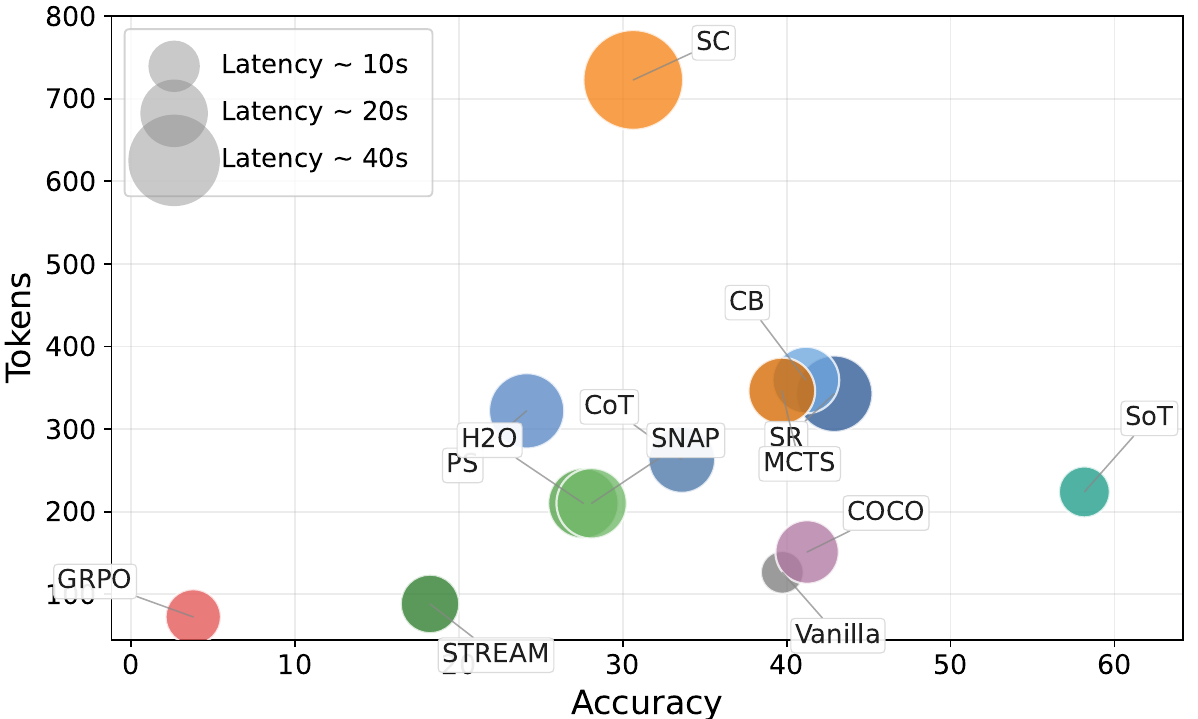}
    \caption{Llama-3.1-8B}
\end{subfigure}
\hfill
\begin{subfigure}{0.49\textwidth}
    \centering
    \includegraphics[width=1.0\textwidth]{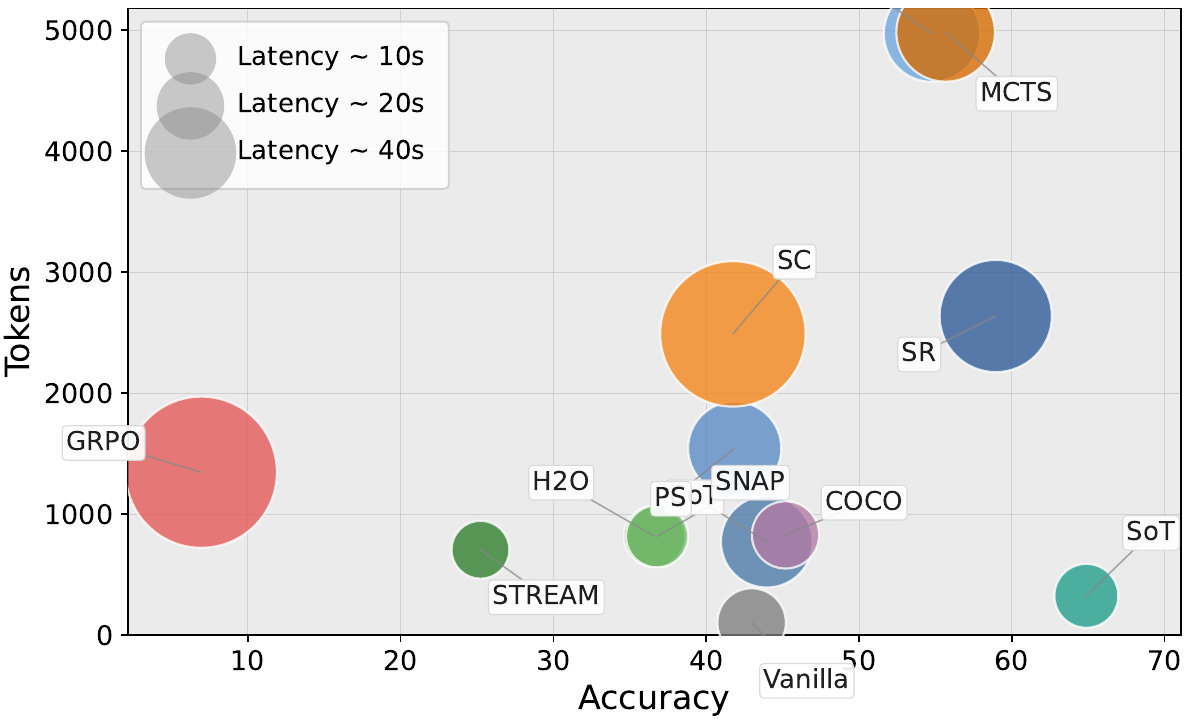}
    \caption{Qwen2.5-14B}
\end{subfigure}

\begin{subfigure}{0.49\textwidth}
    \centering
    \includegraphics[width=1.0\textwidth]{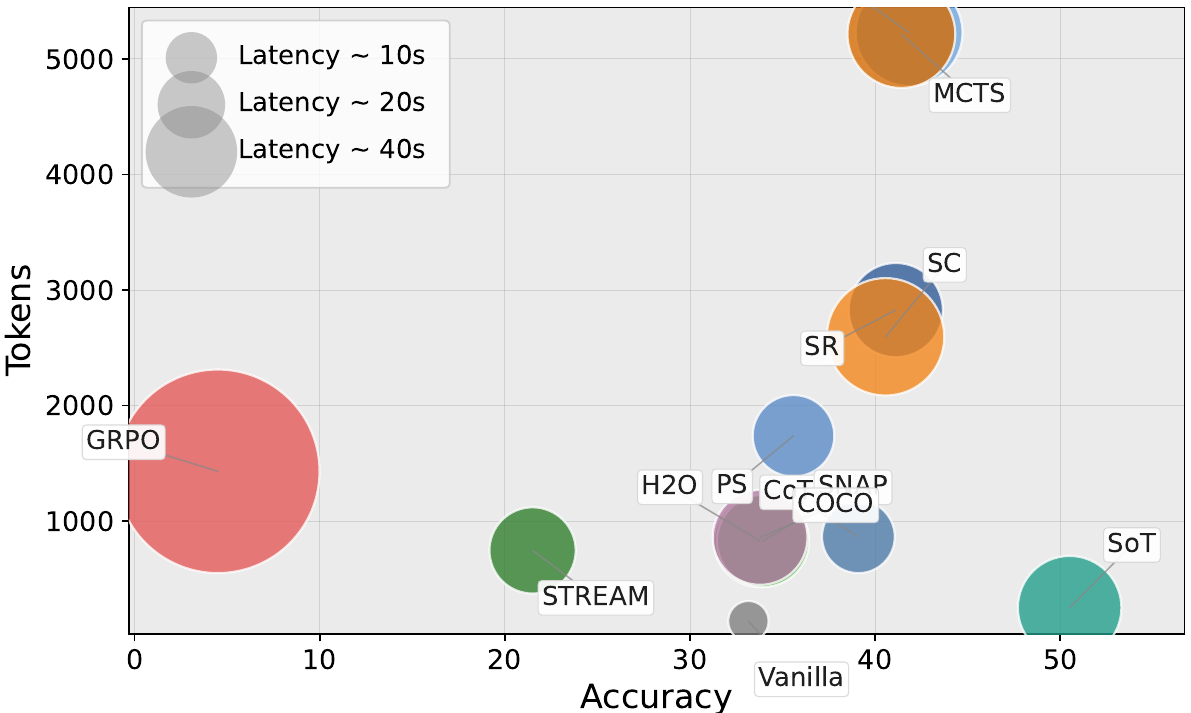}
    \caption{Mixtral-8x7B}
\end{subfigure}
\hfill
\begin{subfigure}{0.49\textwidth}
    \centering
    {\makebox[\linewidth][c]{\includegraphics[width=1.04\textwidth]{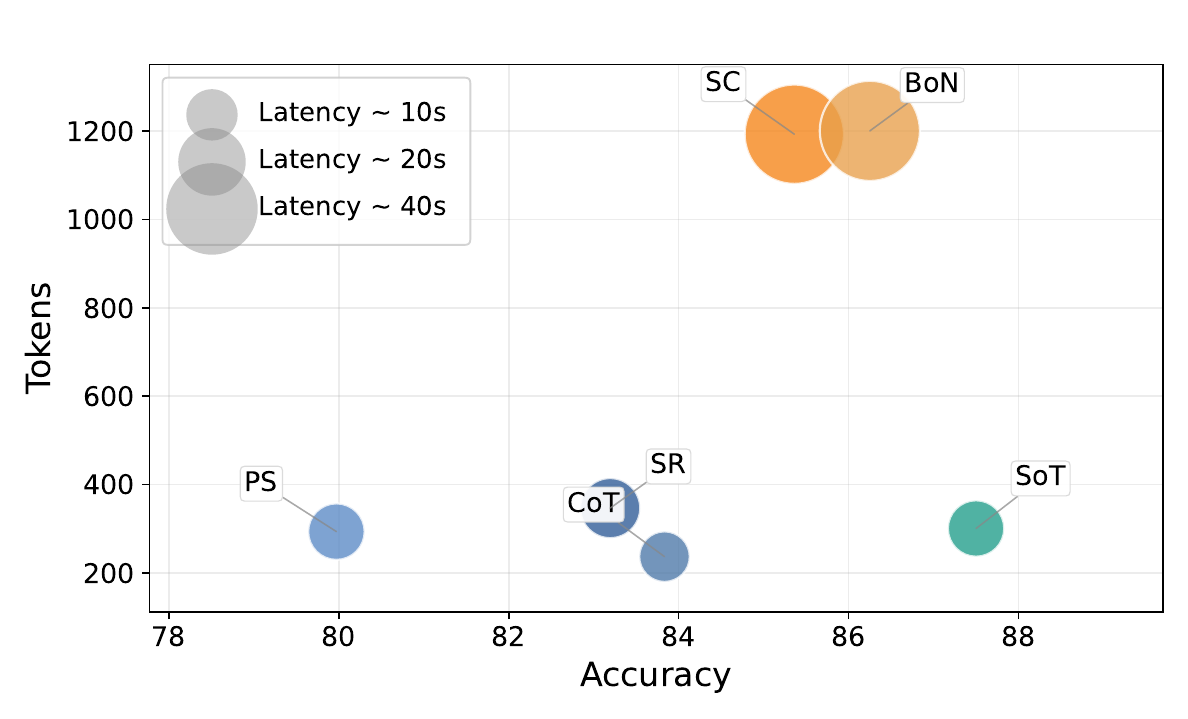}}}
    \caption{{Qwen2.5-VL-7B/32B}}
\end{subfigure}
\caption{\textbf{Extended accuracy--efficiency trade-off plots across backbones.} }
\label{fig:app_tradeoff_all}
\vspace{-3mm}
\end{figure*}

Figure~\ref{fig:app_tradeoff_all} shows a consistent cross-backbone pattern: methods that gain accuracy mainly through broader sampling/search move upward to high-token and high-latency regions, while SoT stays on or near the Pareto frontier.
On text backbones, SoT combines top-tier accuracy with substantially lower compute than search-heavy alternatives, indicating that its gains come from better control allocation rather than budget expansion.
The same trend also appears in multimodal transfer, supporting the view that state-conditioned evidence organization is a mechanism-level efficiency advantage rather than a backbone-specific tuning effect.

\begin{table*}[!t]
\centering
\caption{\textbf{Additional ablation results on Qwen2.5-14B.}}
\label{tab:app_ablation_qwen}
\footnotesize
\setlength{\tabcolsep}{1.75pt}%
\setlength{\aboverulesep}{0pt}%
\setlength{\belowrulesep}{0pt}%
\setlength{\extrarowheight}{0.75pt}%
\renewcommand{\arraystretch}{1.0}%
\begin{tabular}{l%
        @{\kern0.42em}!{\vrule width 0.4pt}@{\kern0.42em}%
        ccc%
        !{\vrule width 1pt}%
        ccc%
        !{\vrule width 1pt}%
        ccc%
        !{\vrule width 1pt}%
        ccc}
\toprule
\multirow{2}{*}{\textbf{Variant}}
& \multicolumn{3}{>{\columncolor[HTML]{E3F2FD}[\tabcolsep][\tabcolsep]}c!{\vrule width 1pt}}{\textbf{QS}}
& \multicolumn{3}{>{\columncolor[HTML]{E3F2FD}[\tabcolsep][\tabcolsep]}c!{\vrule width 1pt}}{\textbf{S\&C}}
& \multicolumn{3}{>{\columncolor[HTML]{E3F2FD}[\tabcolsep][\tabcolsep]}c!{\vrule width 1pt}}{\textbf{GU}}
& \multicolumn{3}{>{\columncolor[HTML]{E3F2FD}[\tabcolsep][\tabcolsep]}c}{\textbf{LCR}} \\
\cmidrule(lr){2-4} \cmidrule(lr){5-7} \cmidrule(lr){8-10} \cmidrule(lr){11-13}
& \textbf{acc.} & \textbf{tok.} & \textbf{lat.}
& \textbf{acc.} & \textbf{tok.} & \textbf{lat.}
& \textbf{acc.} & \textbf{tok.} & \textbf{lat.}
& \textbf{acc.} & \textbf{tok.} & \textbf{lat.} \\
\midrule
Full SoT & 66.1 & 545.8 & 16.8 & 76.8 & 408.4 & 19.0 & 81.3 & 133.2 & 5.8 & 25.2 & 254.0 & 23.4 \\
+ Threshold Tuning & 66.1 & 545.8 & 16.8 & 76.8 & 408.4 & 19.0 & 81.3 & 133.2 & 5.8 & 25.2 & 254.0 & 23.4 \\
\midrule
w/o Evid. Org. \(\mathcal{S}\) & 30.0 & \cellcolor[HTML]{EAECEF}{532.1} & 21.5 & 28.4 & 556.1 & 21.7 & 72.4 & 234.7 & 9.4 & \cellcolor[HTML]{EAECEF}{26.5} & 489.4 & \cellcolor[HTML]{EAECEF}{14.3} \\
w/o Stopping \(\mathcal{T}\) & \cellcolor[HTML]{EAECEF}{76.0} & 1401.9 & 37.4 & 32.4 & 2326.2 & 59.6 & 70.2 & 4175.0 & 100.9 & \cellcolor[HTML]{EAECEF}{31.7} & 2815.8 & 72.8 \\
w/o Geometry \(\delta_t\) & 56.5 & \cellcolor[HTML]{EAECEF}{77.9} & \cellcolor[HTML]{EAECEF}{3.9} & 23.1 & \cellcolor[HTML]{EAECEF}{45.8} & \cellcolor[HTML]{EAECEF}{2.5} & 73.0 & \cellcolor[HTML]{EAECEF}{25.4} & \cellcolor[HTML]{EAECEF}{0.9} & 21.2 & \cellcolor[HTML]{EAECEF}{48.2} & \cellcolor[HTML]{EAECEF}{2.2} \\
w/o Dynamics \((v_t,c_t)\) & \cellcolor[HTML]{EAECEF}{70.0} & 1220.9 & 32.1 & 30.8 & 910.0 & 26.9 & 70.6 & 1170.5 & 31.8 & \cellcolor[HTML]{EAECEF}{31.2} & 869.0 & 27.8 \\
w/o Uncertainty \(H_t\) & 62.5 & \cellcolor[HTML]{EAECEF}{306.0} & \cellcolor[HTML]{EAECEF}{7.6} & 30.6 & \cellcolor[HTML]{EAECEF}{255.6} & \cellcolor[HTML]{EAECEF}{7.6} & 73.0 & \cellcolor[HTML]{EAECEF}{61.3} & \cellcolor[HTML]{EAECEF}{1.9} & \cellcolor[HTML]{EAECEF}{33.6} & \cellcolor[HTML]{EAECEF}{190.2} & \cellcolor[HTML]{EAECEF}{6.6} \\
w/o Sent.-level Units & 62.5 & \cellcolor[HTML]{EAECEF}{306.0} & \cellcolor[HTML]{EAECEF}{7.6} & 30.6 & \cellcolor[HTML]{EAECEF}{255.6} & \cellcolor[HTML]{EAECEF}{7.6} & 73.0 & \cellcolor[HTML]{EAECEF}{61.3} & \cellcolor[HTML]{EAECEF}{1.9} & \cellcolor[HTML]{EAECEF}{33.6} & \cellcolor[HTML]{EAECEF}{190.2} & \cellcolor[HTML]{EAECEF}{6.6} \\
\bottomrule
\end{tabular}
\vspace{-5mm}
\end{table*}

\vspace{-2mm}
\subsection{Additional ablation results}
\label{app:additional_ablation}
\vspace{-1mm}

To test whether the main ablation pattern is specific to Llama-3.1-8B or reflects a more stable mechanism, we report additional ablations on Qwen2.5-14B, Mixtral-8$\times$7B in Table~\ref{tab:app_ablation_qwen} {and Table~\ref{tab:app_ablation_mixtral}}.
The cross-backbone pattern is consistent with the main-text ablation conclusion: removing evidence organization (\(\mathcal{S}\)) causes clear accuracy degradation (e.g., Qwen QS \(66.1\!\rightarrow\!30.0\), S\&C \(76.8\!\rightarrow\!28.4\); Mixtral QS \(46.3\!\rightarrow\!31.3\), LCR \(23.5\!\rightarrow\!3.0\)).
Removing stopping (\(\mathcal{T}\)) can occasionally increase one domain score, but it breaks efficiency by orders of magnitude (e.g., Qwen GU tokens \(133.2\!\rightarrow\!4175.0\)).
Meanwhile, threshold tuning is negligible on text backbones (identical to Full SoT on Qwen/Mixtral), and state-component removals typically trade short outputs for weaker accuracy, supporting the same mechanism-level claim as the main text: SoT gains come from the full closed-loop control, not from isolated calibration or trivial truncation.

\begin{table*}[t]
\centering
\caption{\textbf{Additional ablation results on Mixtral-8$\times$7B.}}
\label{tab:app_ablation_mixtral}
\footnotesize
\setlength{\tabcolsep}{1.75pt}%
\setlength{\aboverulesep}{0pt}%
\setlength{\belowrulesep}{0pt}%
\setlength{\extrarowheight}{0.75pt}%
\renewcommand{\arraystretch}{1.0}%
\begin{tabular}{l%
        @{\kern0.42em}!{\vrule width 0.4pt}@{\kern0.42em}%
        ccc%
        !{\vrule width 1pt}%
        ccc%
        !{\vrule width 1pt}%
        ccc%
        !{\vrule width 1pt}%
        ccc}
\toprule
\multirow{2}{*}{\textbf{Variant}}
& \multicolumn{3}{>{\columncolor[HTML]{E3F2FD}[\tabcolsep][\tabcolsep]}c!{\vrule width 1pt}}{\textbf{QS}}
& \multicolumn{3}{>{\columncolor[HTML]{E3F2FD}[\tabcolsep][\tabcolsep]}c!{\vrule width 1pt}}{\textbf{S\&C}}
& \multicolumn{3}{>{\columncolor[HTML]{E3F2FD}[\tabcolsep][\tabcolsep]}c!{\vrule width 1pt}}{\textbf{GU}}
& \multicolumn{3}{>{\columncolor[HTML]{E3F2FD}[\tabcolsep][\tabcolsep]}c}{\textbf{LCR}} \\
\cmidrule(lr){2-4} \cmidrule(lr){5-7} \cmidrule(lr){8-10} \cmidrule(lr){11-13}
& \textbf{acc.} & \textbf{tok.} & \textbf{lat.}
& \textbf{acc.} & \textbf{tok.} & \textbf{lat.}
& \textbf{acc.} & \textbf{tok.} & \textbf{lat.}
& \textbf{acc.} & \textbf{tok.} & \textbf{lat.} \\
\midrule
Full SoT & 46.3 & 248.2 & 43.8 & 51.7 & 329.8 & 62.6 & 72.5 & 143.6 & 28.0 & 23.5 & 260.6 & 62.8 \\
+ Threshold Tuning & 46.3 & 248.2 & 43.8 & 51.7 & 329.8 & 62.6 & 72.5 & 143.6 & 28.0 & 23.5 & 260.6 & 62.8 \\
\midrule
w/o Evid. Org. \(\mathcal{S}\) & 31.3 & 995.0 & 70.4 & 36.0 & 1002.6 & 65.8 & 63.2 & 836.8 & 51.0 & 3.0 & 921.1 & 90.7 \\
w/o Stopping \(\mathcal{T}\) & 31.3 & 995.0 & 64.0 & 36.0 & 1002.6 & 65.4 & 63.2 & 836.8 & 50.7 & 3.0 & 921.1 & 90.6 \\
w/o Geometry \(\delta_t\) & 34.0 & \cellcolor[HTML]{EAECEF}{58.1} & \cellcolor[HTML]{EAECEF}{11.8} & 30.7 & \cellcolor[HTML]{EAECEF}{56.9} & \cellcolor[HTML]{EAECEF}{13.3} & 57.6 & \cellcolor[HTML]{EAECEF}{50.8} & \cellcolor[HTML]{EAECEF}{7.6} & 3.3 & \cellcolor[HTML]{EAECEF}{54.0} & \cellcolor[HTML]{EAECEF}{13.8} \\
w/o Dynamics \((v_t,c_t)\) & 31.3 & 995.0 & 64.0 & 36.0 & 1002.6 & 65.4 & 63.2 & 836.8 & 50.7 & 3.0 & 921.1 & 90.6 \\
w/o Uncertainty \(H_t\) & 31.3 & 995.0 & 64.0 & 36.0 & 1002.6 & 65.4 & 63.2 & 836.8 & 50.7 & 3.0 & 921.1 & 90.6 \\
w/o Sent.-level Units & 31.3 & 995.0 & 64.0 & 36.0 & 1002.6 & 65.4 & 63.2 & 836.8 & 50.7 & 3.0 & 921.1 & 90.6 \\
\bottomrule
\end{tabular}
\vspace{-2mm}
\end{table*}

\subsection{Limited-access variants and SoT-Judge}

We further expand the exploratory results in Section~\ref{sec:explore} with Tables~\ref{tab:app_boundary_qwen} and \ref{tab:app_boundary_mixtral}.

{On Qwen2.5-14B, Training-free SoT reaches an LCR average of 26.2 versus 25.8 for the Top-3 baselines, while SoT-Embed remains competitive at 23.1. On Mixtral-8x7B, both variants exceed the Top-3 averages in quantitative, symbolic/code, and long-context reasoning, reaching 26.4 and 27.9 versus 23.7 on LCR. The consistent HQA/NarQA gains across both backbones show that trajectory-level state summaries retain useful long-range structure when either controller fitting or internal-state access is removed.}

\begin{table*}[!b]
\centering
\caption{\textbf{Boundary studies of SoT on Qwen2.5-14B (accuracy, \%).}}
\vspace{-2mm}
\label{tab:app_boundary_qwen}
\footnotesize
\setlength{\tabcolsep}{5pt}%
\renewcommand{\arraystretch}{1.05}%
\begingroup
\setlength{\aboverulesep}{0pt}%
\setlength{\belowrulesep}{0pt}%
\setlength{\extrarowheight}{0.95pt}%
\begin{tabular}{l!{\vrule width 0.4pt}*{4}{c}!{\vrule width 1pt}*{6}{c}}
\toprule
& \multicolumn{4}{>{\cellcolor[HTML]{E3F2FD}}c!{\vrule width 1pt}}{\textbf{Quantitative Reasoning}} & \multicolumn{6}{>{\cellcolor[HTML]{E3F2FD}}c}{\textbf{Symbolic and Code}} \\
\cmidrule(lr){2-5}\cmidrule(lr){6-11}
& GSM8K & MATH & DROP & avg. & FOLIO & PW & BBH-T & HE & MBPP & avg. \\
\midrule
Top-3 Baseline avg. & 93.6 & 73.5 & 22.6 & 63.2 & 62.1 & 85.6 & 87.7 & 73.8 & 63.5 & 74.5 \\
SoT-Training-free & 81.8 & 60.8 & 34.6 & 59.1 & 49.5 & 30.2 & 77.0 & 53.0 & 50.4 & 52.0 \\
SoT-Embedding & 82.6 & 58.3 & {34.8} & {58.6} & 51.2 & 59.5 & 71.0 & 48.8 & 49.6 & 56.0 \\
\midrule
\end{tabular}
\par\addvspace{0.35ex}%
\begin{tabular}{l!{\vrule width 0.4pt}*{6}{c}!{\vrule width 1pt}*{4}{c}}
& \multicolumn{6}{>{\cellcolor[HTML]{E3F2FD}}c!{\vrule width 1pt}}{\textbf{General Understanding}} & \multicolumn{4}{>{\cellcolor[HTML]{E3F2FD}}c}{\textbf{Long-Context Reasoning}} \\
\cmidrule(lr){2-7}\cmidrule(lr){8-11}
& CSQA & StrQA & BoolQ & MMLU & RACE & avg. & HQA & NarQA & LB & avg. \\
\midrule
Top-3 Baseline avg. & 72.4 & 70.4 & 84.0 & 69.2 & 64.3 & 72.1 & 19.7 & 29.2 & 28.5 & 25.8 \\
SoT-Training-free & 59.3 & 55.2 & 70.0 & 45.4 & 52.7 & 56.5 & 22.9 & 34.2 & 21.6 & 26.2 \\
SoT-Embedding & 57.0 & 43.0 & 64.8 & 42.8 & 48.7 & 51.3 & {21.2} & {29.8} & 18.2 & {23.1} \\
\bottomrule
\end{tabular}%
\endgroup
\vspace{-1mm}
\end{table*}

\begin{table*}[!b]
\centering
\caption{\textbf{Boundary studies of SoT on Mixtral-8x7B (accuracy, \%).}}
\vspace{-2mm}
\label{tab:app_boundary_mixtral}
\footnotesize
\setlength{\tabcolsep}{5pt}%
\renewcommand{\arraystretch}{1.05}%
\begingroup
\setlength{\aboverulesep}{0pt}%
\setlength{\belowrulesep}{0pt}%
\setlength{\extrarowheight}{0.95pt}%
\begin{tabular}{l!{\vrule width 0.4pt}*{4}{c}!{\vrule width 1pt}*{6}{c}}
\toprule
& \multicolumn{4}{>{\cellcolor[HTML]{E3F2FD}}c!{\vrule width 1pt}}{\textbf{Quantitative Reasoning}} & \multicolumn{6}{>{\cellcolor[HTML]{E3F2FD}}c}{\textbf{Symbolic and Code}} \\
\cmidrule(lr){2-5}\cmidrule(lr){6-11}
& GSM8K & MATH & DROP & avg. & FOLIO & PW & BBH-T & HE & MBPP & avg. \\
\midrule
Top-3 Baseline avg. & 66.6 & 43.1 & 10.2 & 40.0 & 42.7 & 57.4 & 55.3 & 21.7 & 40.4 & 43.5 \\
SoT-Training-free & 75.2 & 52.2 & {34.4} & {53.9} & 31.3 & 50.5 & 63.0 & 56.1 & 49.2 & 50.0 \\
SoT-Embedding & 76.0 & 52.0 & {36.7} & {54.9} & 30.8 & 50.5 & 63.0 & 55.5 & 49.2 & 49.8 \\
\midrule
\end{tabular}
\par\addvspace{0.35ex}%
\begin{tabular}{l!{\vrule width 0.4pt}*{6}{c}!{\vrule width 1pt}*{4}{c}}
& \multicolumn{6}{>{\cellcolor[HTML]{E3F2FD}}c!{\vrule width 1pt}}{\textbf{General Understanding}} & \multicolumn{4}{>{\cellcolor[HTML]{E3F2FD}}c}{\textbf{Long-Context Reasoning}} \\
\cmidrule(lr){2-7}\cmidrule(lr){8-11}
& CSQA & StrQA & BoolQ & MMLU & RACE & avg. & HQA & NarQA & LB & avg. \\
\midrule
Top-3 Baseline avg. & 64.2 & 66.6 & 81.1 & 58.9 & 55.0 & 65.2 & 15.0 & 18.4 & 37.8 & 23.7 \\
SoT-Training-free & 49.5 & 54.5 & 66.5 & 50.8 & 48.3 & 53.9 & {24.3} & {34.3} & 20.5 & {26.4} \\
SoT-Embedding & 49.3 & 44.3 & 52.5 & 51.2 & 47.5 & 49.0 & {24.3} & {34.0} & 25.5 & {27.9} \\
\bottomrule
\end{tabular}%
\endgroup
\vspace{-1mm}
\end{table*}

\clearpage{}
We report full judge behavior in Table~\ref{tab:app_judge_breakdown}.
The same insight as the main section is reinforced: SoT-Judge's largest gains concentrate in trajectory-structure-sensitive regimes (especially LCR, e.g., GPT-5.4 \({81.0}\) vs.\ \(\le {16.7}\) for baselines, Claude 4.7 \({99.0}\) vs.\ \(\le {9.3}\), Qwen-Max \({82.2}\) vs.\ \(\le {14.8}\)), while {trading some GU/S\&C agreement for much stronger QR/LCR agreement}, suggesting that trajectory-state signals are most informative where long-horizon coherence, not local answer matching, is the primary bottleneck.

\begin{table*}[!t]
\centering
\caption{\textbf{Black-box judging with trajectory-based SoT (dataset breakdown, \%).}}
\label{tab:app_judge_breakdown}
\vspace{-2mm}
\footnotesize
\begingroup
\setlength{\tabcolsep}{2.65pt}%
\setlength{\aboverulesep}{0pt}%
\setlength{\belowrulesep}{0pt}%
\setlength{\extrarowheight}{0.88pt}%
\renewcommand{\arraystretch}{0.99}%

\begin{tabular}{ll!{\vrule width 0.4pt} ccc>{\columncolor[HTML]{EAECEF}}c!{\vrule width 1pt}ccccc>{\columncolor[HTML]{EAECEF}}c}
\toprule
\textbf{Model} & \textbf{Method}
& \multicolumn{4}{>{\cellcolor[HTML]{E3F2FD}}c!{\vrule width 1pt}}{\textbf{Quantitative Reasoning}}
& \multicolumn{6}{>{\cellcolor[HTML]{E3F2FD}}c}{\textbf{Symbolic and Code}} \\
\cmidrule(lr){3-6}\cmidrule(lr){7-12}
& & \textbf{GSM8K} & \textbf{MATH} & \textbf{DROP} & \textbf{avg.}
& \textbf{FOLIO} & \textbf{PW} & \textbf{BBH-T} & \textbf{HE} & \textbf{MBPP} & \textbf{avg.} \\
\midrule
\multirow{4}{*}{GPT-5.4}
& Self-Consistency & \underline{92.0} & 78.0 & 5.0 & {58.3} & 69.0 & 77.0 & \underline{86.0} & \underline{87.0} & 77.0 & {79.2} \\
& Self-Verification & \underline{92.0} & 78.0 & 5.0 & {58.3} & 62.0 & 79.0 & 85.0 & \underline{87.0} & \underline{78.0} & {78.2} \\
& LLM-as-a-Judge & \underline{92.0} & \underline{79.0} & 4.0 & {58.3} & \underline{70.0} & 76.0 & \underline{86.0} & 86.0 & \underline{78.0} & {79.2} \\
& \textbf{SoT-Judge} & \textbf{{85.0}} & \underline{\textbf{79.0}} & \underline{\textbf{{94.0}}} & {86.0} & \textbf{{50.0}} & \textbf{{60.0}} & \textbf{{77.0}} & \textbf{86.0} & \textbf{{76.0}} & {69.8} \\
\midrule
\multirow{4}{*}{Claude 4.7}
& Self-Consistency & 90.0 & 78.0 & 1.0 & {56.3} & 65.0 & 73.0 & \underline{58.0} & 94.0 & 86.0 & {75.2} \\
& Self-Verification & \underline{94.1} & \underline{80.0} & 4.0 & {59.4} & \underline{71.4} & \underline{80.0} & 57.1 & \underline{98.4} & \underline{94.4} & {80.3} \\
& LLM-as-a-Judge & 90.0 & 78.0 & 2.0 & {56.7} & 66.0 & 74.0 & 57.0 & 94.0 & 86.0 & {75.4} \\
& \textbf{SoT-Judge} & \textbf{{90.0}} & \textbf{{78.0}} & \underline{\textbf{{91.0}}} & {86.3} & \textbf{{50.0}} & \textbf{78.0} & \textbf{{23.0}} & \textbf{97.0} & \textbf{87.0} & {67.0} \\
\midrule
\multirow{4}{*}{Qwen-Max}
& Self-Consistency & 88.0 & 82.8 & 3.0 & {57.9} & 76.0 & \underline{86.0} & \underline{81.0} & \underline{83.8} & 73.0 & {80.0} \\
& Self-Verification & 46.7 & 35.3 & 6.0 & {29.3} & \underline{80.6} & 82.9 & 74.5 & 81.5 & 70.0 & {77.9} \\
& LLM-as-a-Judge & \underline{92.0} & 82.8 & 3.0 & {59.3} & 74.0 & \underline{86.0} & \underline{81.0} & \underline{83.8} & \underline{74.0} & {79.8} \\
& \textbf{SoT-Judge} & \textbf{{77.0}} & \textbf{{75.9}} & \underline{\textbf{{98.0}}} & {83.6} & \textbf{{68.0}} & \textbf{{64.0}} & \textbf{{74.0}} & \textbf{{75.0}} & \textbf{{67.0}} & {69.6} \\
\midrule
\end{tabular}
\par\addvspace{0.35ex}%
\begin{tabular}{ll!{\vrule width 0.4pt} ccccc>{\columncolor[HTML]{EAECEF}}c!{\vrule width 1pt}ccc>{\columncolor[HTML]{EAECEF}}c}
\textbf{Model} & \textbf{Method}
& \multicolumn{6}{>{\cellcolor[HTML]{E3F2FD}}c!{\vrule width 1pt}}{\textbf{General Understanding}}
& \multicolumn{4}{>{\cellcolor[HTML]{E3F2FD}}c}{\textbf{Long-Context Reasoning}} \\
\cmidrule(lr){3-8}\cmidrule(lr){9-12}
& & \textbf{CSQA} & \textbf{StrQA} & \textbf{BoolQ} & \textbf{MMLU} & \textbf{RACE} & \textbf{avg.}
& \textbf{HQA} & \textbf{NarQA} & \textbf{LB} & \textbf{avg.} \\
\midrule
\multirow{4}{*}{GPT-5.4}
& Self-Consistency & 85.0 & 82.0 & \underline{94.0} & 76.0 & 70.0 & {81.4} & 7.0 & 9.0 & 34.0 & {16.7} \\
& Self-Verification & 84.0 & 81.0 & 93.0 & 75.0 & 70.0 & {80.6} & 7.0 & 9.0 & 32.0 & {16.0} \\
& LLM-as-a-Judge & \underline{86.0} & \underline{85.0} & \underline{94.0} & 75.0 & 70.0 & {82.0} & 7.0 & 9.0 & 33.0 & {16.3} \\
& \textbf{SoT-Judge} & \textbf{85.0} & \textbf{{54.0}} & \textbf{{67.0}} & \textbf{{74.0}} & \textbf{{70.0}} & {70.0} & \underline{\textbf{{98.0}}} & \underline{\textbf{87.0}} & \underline{\textbf{{58.0}}} & {81.0} \\
\midrule
\multirow{4}{*}{Claude 4.7}
& Self-Consistency & \underline{83.0} & 86.0 & \underline{95.0} & 76.0 & 78.0 & {83.6} & 11.0 & 1.0 & 3.0 & {5.0} \\
& Self-Verification & 46.5 & 87.2 & 94.0 & 23.3 & 62.7 & {62.7} & 23.0 & 1.0 & 4.0 & {9.3} \\
& LLM-as-a-Judge & 82.0 & 85.0 & \underline{95.0} & \underline{77.0} & \underline{79.0} & {83.6} & 11.0 & 1.0 & 4.0 & {5.3} \\
& \textbf{SoT-Judge} & \textbf{81.0} & \textbf{{49.0}} & \textbf{{53.0}} & \textbf{{67.0}} & \textbf{{74.0}} & {64.8} & \underline{\textbf{{100.0}}} & \underline{\textbf{{99.0}}} & \underline{\textbf{98.0}} & {99.0} \\
\midrule
\multirow{4}{*}{Qwen-Max}
& Self-Consistency & 82.0 & 82.0 & 91.0 & 77.0 & \underline{88.0} & {84.0} & 11.0 & 1.0 & 31.5 & {14.5} \\
& Self-Verification & \underline{86.7} & \underline{86.2} & \underline{93.8} & \underline{80.6} & 85.0 & {86.5} & 1.9 & 1.4 & 31.5 & {11.6} \\
& LLM-as-a-Judge & 84.0 & 82.0 & 93.0 & 77.0 & \underline{88.0} & {84.8} & 12.0 & 1.0 & 31.5 & {14.8} \\
& \textbf{SoT-Judge} & \textbf{{80.0}} & \textbf{{74.0}} & \textbf{{75.0}} & \textbf{{72.0}} & \textbf{{86.0}} & {77.4} & \underline{\textbf{{95.0}}} & \underline{\textbf{{100.0}}} & \underline{\textbf{{51.7}}} & {82.2} \\
\bottomrule
\end{tabular}%
\endgroup
\vspace{-2mm}
\end{table*}

\subsection{Sensitivity to trajectory quality}
\label{app:traj_quality}
We test whether SoT remains informative when the offline trajectory pool is collected or curated differently, with Llama-3.1-8B results in Table~\ref{tab:app_traj_quality} (default HQ pool vs.\ weaker variants and training-free SoT).
The table shows a consistent split: weaker pools largely leave QS/S\&C accuracy near plateau but disproportionately hurt GU/LCR while greatly inflating tokens and latency relative to HQ; training-free SoT sits in the same heavy-trajectory regime with domain-mixed accuracy shifts.
Overall, curation affects whether supervision yields a cheap, reliable trajectory-level control signal more than it acts as a uniform rescaling of backbone capability.

\begin{table*}[!b]
\centering
\caption{\textbf{Sensitivity to trajectory-pool quality on Llama-3.1-8B}.}
\label{tab:app_traj_quality}
\vspace{-2mm}
\footnotesize
\begingroup
\setlength{\tabcolsep}{4.2pt}%
\setlength{\aboverulesep}{0pt}%
\setlength{\belowrulesep}{0pt}%
\setlength{\extrarowheight}{0.75pt}%
\renewcommand{\arraystretch}{1.0}%
\begin{tabular}{l!{\vrule width 0.4pt}ccc!{\vrule width 1pt}ccc!{\vrule width 1pt}ccc!{\vrule width 1pt}ccc}
\toprule
\multirow{2}{*}{\textbf{Trajectory pool}}
& \multicolumn{3}{>{\cellcolor[HTML]{E3F2FD}}c!{\vrule width 1pt}}{\textbf{QS}}
& \multicolumn{3}{>{\cellcolor[HTML]{E3F2FD}}c!{\vrule width 1pt}}{\textbf{S\&C}}
& \multicolumn{3}{>{\cellcolor[HTML]{E3F2FD}}c!{\vrule width 1pt}}{\textbf{GU}}
& \multicolumn{3}{>{\cellcolor[HTML]{E3F2FD}}c}{\textbf{LCR}} \\
\cmidrule(lr){2-4} \cmidrule(lr){5-7} \cmidrule(lr){8-10} \cmidrule(lr){11-13}
& \textbf{acc.} & \textbf{tok.} & \textbf{lat.}
& \textbf{acc.} & \textbf{tok.} & \textbf{lat.}
& \textbf{acc.} & \textbf{tok.} & \textbf{lat.}
& \textbf{acc.} & \textbf{tok.} & \textbf{lat.} \\
\midrule
Full HQ pool & 70.8 & 247.9 & 6.3 & 57.1 & 324.2 & 11.1 & 75.5 & 58.0 & 1.6 & 37.3 & 121.1 & 97.5 \\
No filter & 71.3 & 931.3 & 30.0 & 57.7 & 1080.8 & 84.1 & 49.2 & 886.0 & 29.5 & 21.5 & 900.9 & 91.5 \\
Weak filter & 71.3 & 931.3 & 24.3 & 57.7 & 1080.8 & 34.5 & 49.2 & 886.0 & 17.7 & 21.3 & 900.9 & 90.7 \\
No pipeline & 71.3 & 931.3 & 26.1 & 57.7 & 1080.8 & 53.4 & 49.2 & 886.0 & 49.3 & 21.5 & 900.9 & 89.3 \\
Training-free & 72.3 & 917.4 & 26.0 & 58.3 & 1078.9 & 35.0 & 53.8 & 906.6 & 19.7 & 28.1 & 872.9 & 95.8 \\
\bottomrule
\end{tabular}%
\endgroup
\vspace{-2mm}
\end{table*}

\clearpage
\vspace{-3mm}
\section{Case Studies}
\label{app:case_studies}
\vspace{-2mm}

\subsection{Case 1: Quantitative Reasoning (QS)}
\begin{figure}[H]
\centering
\includegraphics[width=0.97\textwidth]{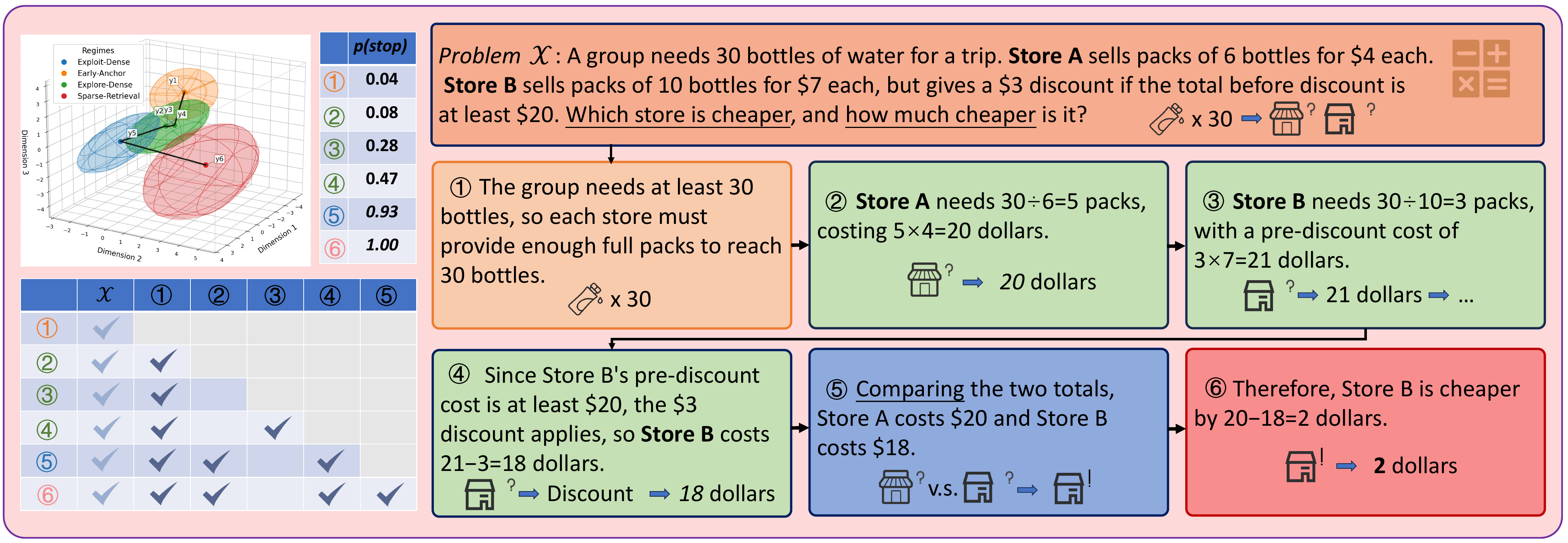}
\vspace{-2mm}
\caption{Case study on Quantitative Reasoning (QS).}
\label{fig:app_case_arithmetic}
\end{figure}

\vspace{-1mm}
Figure~\ref{fig:app_case_arithmetic} follows a staged numeric workflow (per-store accounting, then comparison).
Intermediate steps reuse only the quantities that remain algebraically active in the next operation, and \(p(\mathrm{stop})\) ramps once the competing subtotals are both fixed: the mechanism reads as \emph{derivation support} that is pruned as subgoals close, not as uniform context replay.
\vspace{-3mm}

\subsection{Case 2: General Understanding (GU)}
\begin{figure}[H]
\centering
\includegraphics[width=0.97\textwidth]{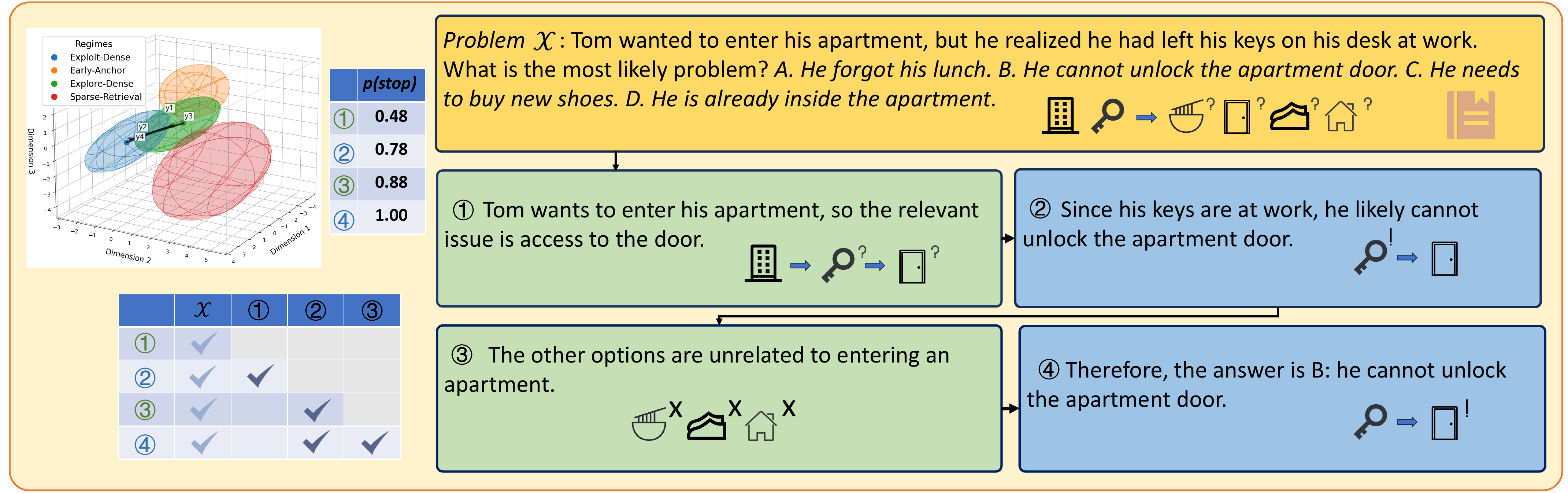}
\vspace{-2mm}
\caption{Case study on General Understanding (GU).}
\label{fig:app_case_cki}
\end{figure}

\vspace{-1mm}
Figure~\ref{fig:app_case_cki} is a commonsense elimination problem: late steps depend on the prompt and the locally decisive intermediates while dropping earlier scaffolding—a pattern that matters when verbal distractors would otherwise pollute a full-history transcript.
\vspace{-3mm}

\subsection{Case 3: Symbolic and Code (S\&C)}
\begin{figure}[H]
\centering
\includegraphics[width=0.97\textwidth]{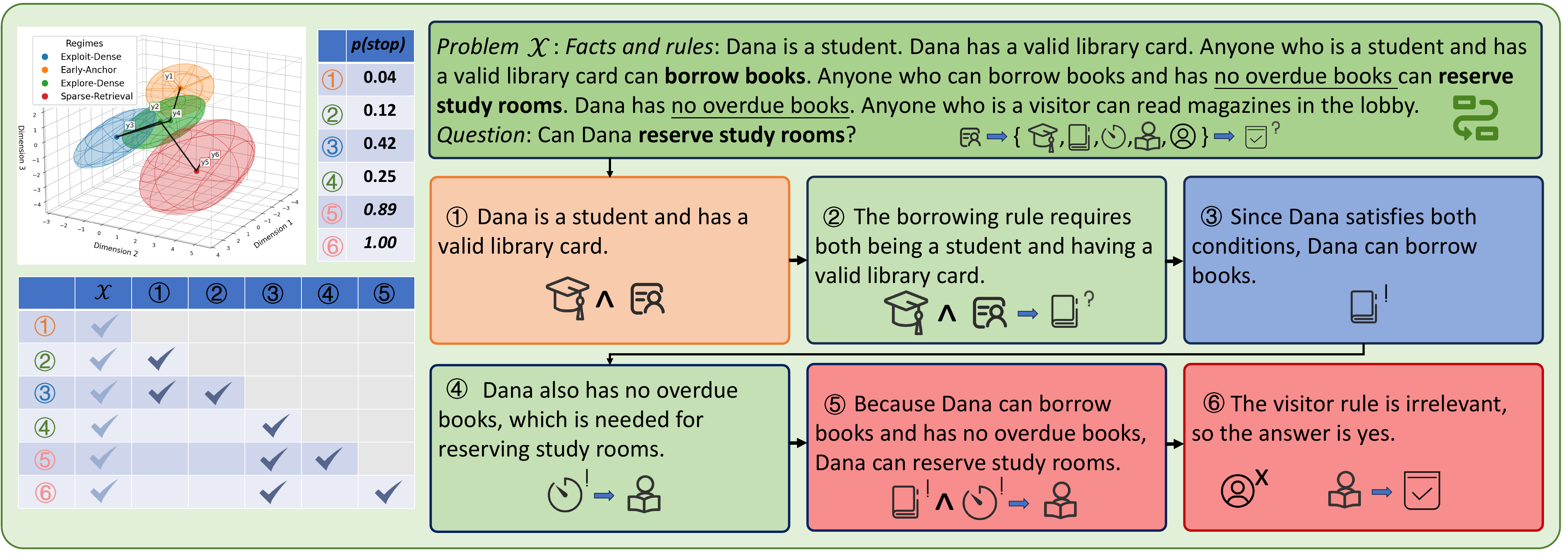}
\vspace{-2mm}
\caption{Case study on Symbolic and Code (S\&C).}
\label{fig:app_case_formal}
\end{figure}

\vspace{-1mm}
Figure~\ref{fig:app_case_formal} chains explicit rules forward but the evidence matrix is non-monotone—later steps re-anchor on conclusions and skip obsolete rule instantiations—so correctness is tied to \emph{structural} carry (what still entails the next judgment), not to preserving the full forward chain verbatim.
\vspace{-3mm}

\subsection{Case 4: Long-Context Reasoning (LCR)}
\begin{figure}[H]
\centering
\includegraphics[width=0.98\textwidth]{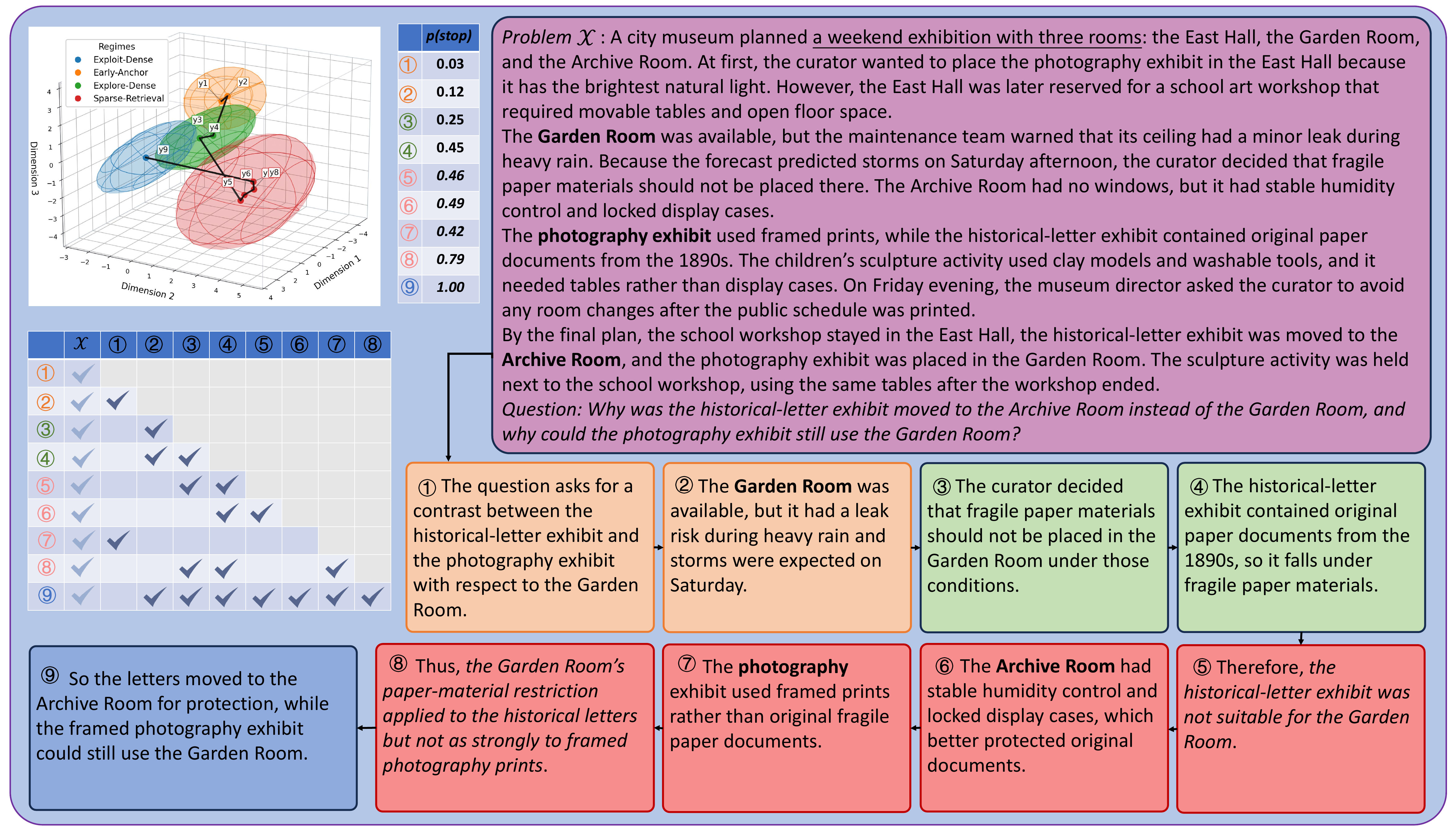}
\vspace{-2mm}
\caption{Case study on Long-Context Reasoning (LCR).}
\label{fig:app_case_longcontext}
\end{figure}

\vspace{-1mm}
Figure~\ref{fig:app_case_longcontext} integrates constraints spread across a long narrative; the trajectory lingers in exploration-like regimes while facts are harvested, then compresses into a sparse-retrieval phase that fuses only the clauses still relevant to the contrastive \emph{why} question—i.e., compositional support rather than verbatim long-context carry.
\vspace{-3mm}

\subsection{Case 5: Visual-language reasoning}
\begin{figure}[H]
\centering
\includegraphics[width=0.98\textwidth]{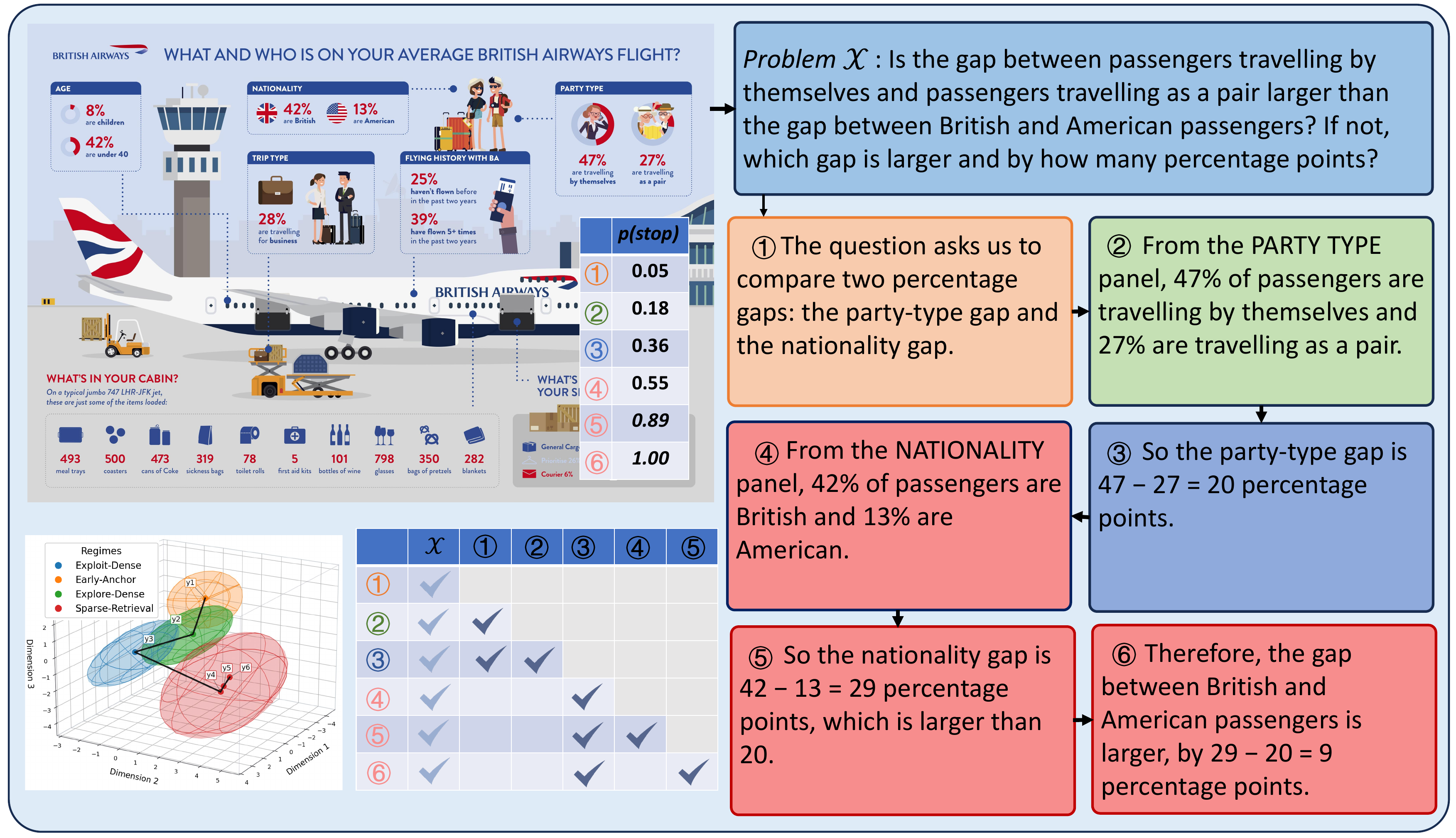}
\vspace{-2mm}
\caption{Case study on visual-language reasoning.}
\label{fig:app_case_vlm}
\end{figure}

\vspace{-1mm}
Figure~\ref{fig:app_case_vlm} (Qwen3-VL-8B) shows the same interface on chart QA: panel-wise extraction, numeric gap reasoning, rising \(p(\mathrm{stop})\) through the computation, and sparse dependencies—evidence that SoT remains a trajectory-level organizer when evidence is grounded in pixels rather than text alone.
\vspace{-3mm}

\clearpage
\AppBlock{Part E: Related Work and Discussion}
\vspace{-2mm}

\section{Related Work}
\label{app:related_work}
\vspace{-1mm}

This section expands the compressed related-work discussion from the main paper.
SoT sits at the intersection of four research directions, but is not reducible to any one of them.
\vspace{-1mm}

\subsection{Externally scripted reasoning}
\vspace{-1mm}
One major line of work improves reasoning by imposing explicit inference-time structure:
step-by-step prompting~\citep{wei2022chain},
self-consistency~\citep{wang2023selfconsistency},
plan-and-solve decomposition~\citep{wang2023plan},
self-refinement~\citep{madaan2023self},
and tree-style search~\citep{yao2023tree}.
These methods substantially advance empirical reasoning performance, but they share a common modeling assumption:
reasoning improves when we prescribe a better \emph{external} program for how the model should think.

SoT differs at the level of problem formulation.
It does not ask for a better externally specified chain.
It asks whether the model's current endogenous regime can decide what evidence should remain active for the next decision.
In this sense, SoT shifts the focus from \emph{external reasoning choreography} to \emph{endogenous reasoning control}.

\vspace{-1mm}
\subsection{Internal reasoning trajectories}
\vspace{-1mm}
Another relevant line studies hidden-state geometry, representation dynamics, and correctness-related trajectory structure in language models~\citep{ballon2026probing,damirchi2026truth,park2025geometry,park2024linear,sun2026llm}.
Related work also shows that internal signals can support monitoring and early-exit decisions~\citep{feng2025monitoring,hosseini2026early,tikhonov2026confidence}.
These works motivate the view that reasoning-relevant structure exists inside the model.

However, most of this literature is primarily descriptive:
it interprets hidden trajectories, monitors confidence, or probes latent variables.
SoT draws on similar geometric-dynamic structure but uses it differently.
Rather than analyzing the state after the fact, SoT engages it online as the control variable that determines evidence organization and continuation.

\vspace{-1mm}
\subsection{Context augmentation and compression}
\vspace{-1mm}
Retrieval-augmented and context-management methods also intersect with SoT.
Examples include retrieval-interleaved reasoning~\citep{trivedi2023interleaving,wang2024rat}, context compression~\citep{jiang2024longllmlingua,li2023compressing}, and KV-cache or memory-selection methods such as H\(_2\)O, SnapKV, and StreamingLLM~\citep{li2024snapkv,xiao2023streamingllm,zhang2023h2o}.
These methods are close to SoT in execution form because they do not always carry the entire history forward.

The key distinction lies in \emph{what drives selection}.
Most prior methods rely on semantic relevance, recency, generic compression objectives, or infrastructure efficiency.
SoT instead conditions historical activation on the model's current reasoning regime.
This turns memory selection from a general context-management tool into a component of reasoning control itself.

\vspace{-1mm}
\subsection{Latent reasoning and test-time control}
\vspace{-1mm}
Recent work also explores reasoning in latent space or with alternative test-time control ~\citep{hao2025training,hassid2025dont,wang2025system}.
Some methods push more reasoning inside hidden computation; others show that shorter or differently allocated reasoning traces can be beneficial.
These directions are highly relevant because they challenge the naive idea that better reasoning must always mean longer explicit token chains.

SoT is complementary to this line.
It does not replace explicit reasoning with a fully latent process, nor does it merely shorten chains.
Instead, it treats explicit sentence-level reasoning as a trajectory whose supporting context should be routed by a compact endogenous controller.
This makes SoT a bridge between explicit-token and latent-control views of reasoning.

\vspace{-1mm}
\subsection{Positioning summary}
\vspace{-1mm}
Overall, SoT should not be read as another CoT variant, another search baseline, or another memory-pruning heuristic.
Its central contribution is to formulate reasoning-time control as a closed loop in which endogenous state governs sparse evidence activation and stopping.
That is a different object of study from externally scripted reasoning, descriptive hidden-state analysis, or generic context compression.

\section{Limitations and Future Work}
\label{app:limitations_future}

This paper focuses on a concrete instantiation of SoT across multiple text backbones and {two scales of one multimodal backbone family}; several scope boundaries naturally remain.
From a broader-impact perspective, improving reasoning under fixed inference budgets could make capable models more usable in cost- or latency-sensitive settings (e.g., education, healthcare-related assistance, research tooling), where efficiency and reliability both matter.
As with any method that can improve fluent reasoning, deployment should follow domain-appropriate safeguards (e.g., policy constraints, human oversight in high-stakes use, and monitoring for misuse).

\subsection{Current limitations}
\vspace{-1mm}
\begin{itemize}[leftmargin=1.6em,itemsep=0pt,topsep=0.1em,parsep=0pt,partopsep=0pt]
    \item \textbf{Performance depends on the state interface.}
    SoT routes memory and stopping through the extracted endogenous state; if that interface is a poor summary of the regime relevant to the next decision, control quality can degrade.

    \item \textbf{The current state is intentionally compact.}
    Our implemented controller uses \(m_t\in\mathbb{R}^{4}\) for clarity and portability, but this choice may under-specify unusually long, highly branching, or richly grounded trajectories where additional structure could help.

    \item \textbf{Mechanistic evidence is supportive but not uniquely identifying.}
    Our ablations and interventions are consistent with state-conditioned routing playing a central role, yet---as in most systems that couple representation, memory, and generation---alternative narratives cannot be ruled out in every corner case without heavier instrumentation.

    \item \textbf{Empirical coverage matches the paper's claims, not every deployment regime.}
    We report results across a broad multi-task suite, but stronger proprietary models, different tool-use/agent stacks, and safety-critical deployments are outside our evaluated scope.

    \item \textbf{Limited-access variants are informative but not exhaustive.}
    Training-free SoT, embedding-based features, and \textsc{SoT-Judge} probe the same interface under restricted supervision or observability; they are evaluated at a smaller footprint than the full supervised controller and are best read as exploratory complements.
\end{itemize}

\subsection{Future work}
\vspace{-1mm}
\begin{itemize}[leftmargin=1.6em,itemsep=0pt,topsep=0.1em,parsep=0pt,partopsep=0pt]
    \item \textbf{Richer but still compact states.}
    A natural direction is to add state dimensions or structured summaries only when they measurably improve control (e.g., branching, retrieval confidence, multimodal grounding), without bloating the feedback loop.

    \item \textbf{Tighter identification of what the controller uses.}
    Additional counterfactual replay, routing interventions, and closed-loop stress tests could further clarify which components of the interface matter most across task families.

    \item \textbf{Adaptive memory units.}
    Sentence-level units are easy to audit, but variable segmentation or hybrid symbolic/latent memories may better match heterogeneous reasoning traces.

    \item \textbf{Unified white-box and black-box trajectory control.}
    SoT-Embed and \textsc{SoT-Judge} suggest a continuum between internals-aware controllers and trajectory-only signals; a more systematic comparison of that continuum is an interesting next step.

    \item \textbf{Broader multimodal and interactive settings.}
    Environments with dynamic evidence---long multimodal contexts, tool use, and interactive decisions---are a natural fit for state-conditioned organization, but will require task suites and training protocols beyond this paper.
\end{itemize}

\paragraph{Final perspective.}
We intend SoT as a precise formulation of reasoning-time control---not a claim that any single architecture is final.
The underlying point is empirical and conceptual: reasoning benefits from treating context activation and continuation as a feedback process coupled to the model's evolving internal regime, rather than only as prompt programming over a growing transcript.